\documentclass[preprint,authoryear,3p]{elsarticle}

\usepackage{booktabs}

\usepackage{amssymb,amsthm,amsmath}

\usepackage{graphicx}
\usepackage{subfig}
\usepackage{float}
\usepackage{multirow}
\usepackage{longtable,lscape}
\usepackage{cases}
\usepackage{url}
\usepackage{caption}
\usepackage{color}
\usepackage{amsthm}
\usepackage[pagewise]{lineno}

\usepackage{fancyhdr}
\usepackage{nomencl}
\makenomenclature

\usepackage{etoolbox}
\renewcommand\nomgroup[1]{%
 \item[\bfseries
  \ifstrequal{#1}{A}{States and Actions}{%
  \ifstrequal{#1}{B}{Car-following Models}{%
  \ifstrequal{#1}{C}{Other Notations}{}}}%
]}

\usepackage{array}
\usepackage{multirow}
\usepackage{multicol}
\usepackage{makecell}
\usepackage{comment}
\usepackage{graphicx}
\usepackage{soul}

\usepackage{booktabs,caption}
\usepackage[flushleft]{threeparttable}
\usepackage{arydshln}
\usepackage[ruled,vlined, linesnumbered]{algorithm2e}

\usepackage{enumerate}

\usepackage{makecell}
\setcellgapes{4pt}
\usepackage[export]{adjustbox}
\definecolor{mygray}{gray}{0.6}

\usepackage{fullpage}

\theoremstyle {plain}

\newcounter{x}

\theoremstyle{definition}
\newtheorem{defn}{Definition}[section]

\theoremstyle{remark}

\newtheorem*{note}{Note}

\usepackage{titlesec}
\titleformat{\paragraph}
{\normalfont\normalsize\bfseries}{\theparagraph}{1em}{}
\titlespacing*{\paragraph}
{0pt}{3.25ex plus 1ex minus .2ex}{1.5ex plus .2ex}

\usepackage{color}

\newcommand{\tcck}[1]{{\textcolor{black}{#1}}}

\makeatletter
\def\ps@pprintTitle{%
	\let\@oddhead\@empty
	\let\@evenhead\@empty
	\def\@oddfoot{\reset@font\hfil\thepage\hfil}
	\let\@evenfoot\@oddfoot
}
\makeatother

\numberwithin{equation}{section}

\journal{Transportation Research Part C}

\begin{document}

\begin{frontmatter} 
		
		
		
\title{A Physics-Informed Deep Learning Paradigm for Car-Following Models}
		
\date{\today}
		
		
\author[cu]{Zhaobin Mo}

\author[cu]{Rongye Shi}

\author[cu,dsi]{Xuan Di\corref{cor}}
\ead{sharon.di@columbia.edu}

\cortext[cor]{Corresponding author. Tel.: +1 212 853 0435;}

\address[cu]{Department of Civil Engineering and Engineering Mechanics, Columbia University}
\address[dsi]{Data Science Institute, Columbia University}

\begin{abstract}

Car-following behavior has been extensively studied using physics-based models, such as Intelligent Driving Model (IDM). These models successfully interpret traffic phenomena observed in the real world but may not fully capture the complex cognitive process of driving. Deep learning models, on the other hand, have demonstrated their power in capturing observed traffic phenomena but require a large amount of driving data to train. This paper aims to develop a family of neural network based car-following models that are informed by physics-based models, which leverage the advantage of both physics-based (being data-efficient and interpretable) and deep learning based (being generalizable) models. 
We design physics-informed deep learning car-following model (PIDL-CF) architectures encoded with 4 popular physics-based models - 
\tcck{
the IDM, the Optimal Velocity Model, the Gazis-Herman-Rothery model, and the Full Velocity Difference Model.}
Acceleration is predicted for 4 traffic regimes: acceleration, deceleration, cruising, and emergency braking. 
\tcck{The generalization of PIDL method is further validated using two representative neural network models: the artificial neural networks (ANN) and the long short-term memory (LSTM) model.}
Two types of PIDL-CF problems are studied, one to predict acceleration only and the other to jointly predict acceleration and discover model parameters. 
We also demonstrate the superior performance of PIDL with the Next Generation SIMulation (NGSIM) dataset over baselines, especially when the training data is sparse. 
The results demonstrate the superior performance of neural networks informed by physics over those without.


\end{abstract}
		
\end{frontmatter}

\let\thefootnote\relax\footnotetext{Accepted by Transportation Research Part C (\url{https://doi.org/10.1016/j.trc.2021.103240})}

\section{Motivation}
\label{sec:intro}

Car following (CF) behavioral modeling, dated back to 1930s since Greenshields' pioneering paper \citep{greenshields1935study}, 
has long been studied by the transportation community to reveal how human drivers react to ambient traffic environment. 
Recent years have witnessed a growing interest of modeling CF behavior using deep learning (DL) methods,  
thanks to big data generated from a variety of sensors \citep{di2021survey}.  
Understanding how people drive is not only crucial to develop human-like autonomous driving policies, but also to enhance the inference ability of autonomous vehicles in the transition period when they have to drive alongside humans.

We categorize CF modeling methods into two categories: physics-based and data-driven. 
\textbf{Physics-based} (also known as model-based or theory-based \citep{zhou2019longitudinal}) CF models refer to
pre-designated mathematical functional forms with a limited number of parameters. 
Physics-based driving models 
assume each driver-vehicle unit behaves like an automated particle or automaton, 
within which human cognitive process and the machine's mechanical dynamics are highly simplified. 
In microscopic models, cars are assumed to select their driving velocity and acceleration dynamically 
based on the following distance from their immediate leader, speed difference, and other features. 
The mathematical tool is ordinary differential equation (ODE).  
The widely used car following models (CFMs) include intelligent driver model (IDM)~\citep{Treiber-2000} and optimal velocity model (OVM)~\citep{Bando-1995}. 

\textbf{Data-driven} 
CF models refer to cutting-edge models that mimic human intelligence, leveraging deep neural networks, reinforcement learning, imitation learning, and other advanced machine learning methods. 
There is no predesignated mathematical form and model training purely relies on observations. 
A growing number of data-driven models employ (deep) neural networks (NN). 
An artificial neural networks (ANN) model was first applied in \citep{Panwai2007} to model car-following behaviors, which showed that ANN outperformed physics-based models like Gipps in terms of overall prediction accuracy and robustness. 
\tcck{
Data-driven models are also found to be able to better capture the asymmetric car-following behaviors, which can not be well solved by pure physics-based and data-driven methods. 
\tcck{Recurent Neural Networks (RNN) and Long Short-Term Memory (LSTM) model}  \citep{Zhou2017g,huang2018car,gu2020lstm,shou2020long} are other paradigms of NN, which leverage historic data and are able to learn the asymmetric behaviors arising from driver's different driving patterns in free-flow and stop-and-go regimes. }  
Neural networks exploit no explicit traffic models nor make pre-assumptions, and thus can be treated as an uninterpretable black box. 
They also contain a large number of parameters, usually over hundreds or thousands. 
Such a large parameter space is a two-sided sword. On one hand, it helps learn and approximate a more realistic driving pattern. On the other hand, it demands large training data and incurs high computational cost. 

\tcck{
Both physics-based and data-driven models have their pros and cons. 
The physics-based models may not be generalizable, because 
the predefined motion heuristics usually make strong assumptions about driving behaviors with a small set of parameters, 
which may not be able to capture human's strategic planning behaviors and may not generalize well to diverse driving scenarios in a highly interactive environment. 
Data-driven models are data-hungry, 
meaning that they require a large amount of driving data to calibrate.  
In addition, lack of interpretability is a key shortcoming of these models. In other words, the trained neural networks can not offer insights into how driving behavior evolves as traffic environment changes. 
Moreover, compared to physics-based models, 
the DL method is more vulnerable to data noise and unseen data. 
Driving data, usually measured by GPS sensors, is generally sparse because of high installation and data acquisition cost. Thus a paradigm that is data-efficient and highly accuracy is preferred, especially in the advent of the era of autonomous driving. 
}
This paper aims to develop a family of NN based CFMs that are informed by physics-based models, which leverage the advantage of both physics-based (which is data-efficient and interpretable) and deep learning based (which is generalizable) models.

The rest of the paper is organized as follows: 
Section \ref{sec:related_works} introduces the related works and preliminaries of CFMs and PIDL. 
Section \ref{sec:methodology} introduces two types of problems of PIDL for car-following modeling (PIDL-CF): one for acceleration prediction only and one for joint prediction and parameter discovery, within which each component of a PIDL-CF is introduced, including architecture design, collocation point selection, loss function design, and training algorithms. 
Section \ref{sec:numerical} demonstrates the robustness of PIDL-CF by encoding two CFMs, namely IDM and OVM, 
and Section \ref{sec:real_exp} evaluates the PIDL on the NGSIM data. 
Section \ref{sec:conclusion} concludes our work and projects future research questions in this promising arena. 

\section{Related Work}
\label{sec:related_works}

Prior to delving into literature review, we will first define notations that will be used for the rest of the paper. 

\nomenclature[A, 01]{$h$}{spacing.}
\nomenclature[A, 02]{$\Delta v$}{velocity difference between the subject vehicle and its leader.}
\nomenclature[A, 03]{$v$}{velocity of the subject vehicle.}
\nomenclature[A, 04]{$\mathbf{s}$}{state $\mathbf{s}=(h, \Delta v, v)$, which is a vector of spacing, velocity difference and velocity.}
\nomenclature[A, 04.1]{$\cal S$}{state space.}
\nomenclature[A, 05]{$a$}{action, which is the longitudinal acceleration.}
\nomenclature[A, 05.5]{$\cal A$}{action space.}
\nomenclature[A, 06.1]{$\cal O$}{observed state space. }
\nomenclature[A, 06.3]{$\cal C$}{collocation state space.}
\nomenclature[A, 08]{$()^{(i)}$}{the $i$th state or action after shuffling.}
\nomenclature[A, 08.1]{$\hat{()}$}{observed state or action. }
\nomenclature[A, 09.4]{$N$}{number of elements in a set of states.}
\nomenclature[A, 10]{$N_O$}{number of observed states. }
\nomenclature[A, 11]{$N_C$}{number of collocation states.}

\nomenclature[B,01]{$\theta$}{parameters of the physics-uninformed neural network.}
\nomenclature[B,02]{$\lambda$}{parameters of the physics-based computational graph, $\lambda \subseteq \Lambda$. }
\nomenclature[B,02.5]{$\Lambda$}{domain of $\lambda$.}
\nomenclature[B,04]{$f_\theta$}{function parameterized by $\theta$ that maps from states to actions.}
\nomenclature[B,05]{$f_\lambda$}{function parameterized by $\lambda$ that maps from states to actions.}
\nomenclature[B,06]{$\cal N_\theta$}{physics-uninformed neural network.}
\nomenclature[B,07]{$\cal N_\lambda$}{physics-based computational graph.}
\nomenclature[B,08]{$\alpha$}{weight of the physics-uninformed neural network that contributes to the loss function.}
\printnomenclature

\subsection{Preliminaries of CFM}

Denote ${\cal S}\subseteq \mathbb{R}_+^m$ and ${\cal A}\subseteq \mathbb{R}_+^n$ as state and action spaces, respectively, where $m\in \mathbb{N}_+$ and $n\in \mathbb{N}_+$ are the dimensions of state and action vectors, respectively.
A car-following model (CFM) is a mathematical mapping parameterized by $\theta$, denoted as $f_{\theta}(\cdot|\theta)$, from states $\mathbf{s}\in {\cal S}$ (i.e., observations of the traffic environment) to actions $a\in {\cal A}$ (i.e., longitudinal accelerations): 
\begin{equation}\label{eqn: generic_CFM}
f_{\theta}: \mathbf{s} \longrightarrow a. 
\end{equation}

The input of the CFM includes all possible signals $\mathbf{s}(t)$ at time $t$ from neighboring vehicles in the traffic environment
, and its output is the acceleration $a({t+\Delta t})$ at the next time step. 
The key question of CFMs is to find an optimal set of parameters $\theta^*$ that best fit real-world data. 
Modeling the driving policy mapping is categorized into physics-based and DL-based. 
Physics-based mapping can be characterized by mathematical formulas, 
while DL-based mapping is usually represented by deep neural networks. 
Table~\ref{tab:map} summarizes relevant models in these two categories, along with a third category on the hybrid of these two. 

\begin{table}[H]\centering
\begin{threeparttable}
	\centering
	\caption{A summary of car-following models \label{tab:map} }
	\begin{tabular}{|m{0.38 cm}<{\centering} |m{2.2 cm}<{\centering} ||p{1.7 cm}<{\centering} |p{1 cm}<{\centering} ||p{4 cm}<{\centering} ||p{4 cm}<{\centering} |}
		\hline
		\multicolumn{2}{|c||}{Models} & Input & Output & Physics Model Parameters & Reference \\ \hline
		\multirow{6}{*}{\rotatebox[origin=c]{90}{\textcolor{black}{Physics-Based}}}
		
		& IDM & $h, \Delta v, v$ &  $a$  & desired velocity; desired time headway; minimum spacing; maximum acceleration; comfortable deceleration  &  \cite{treiber2000congested, ossen2011heterogeneity, rahman2017evaluation, huang2018experimental,sharma2019more}
		\\ \cline{2-6}
		
		& OVM & $h,  v$ &  $a$  & maximum velocity; safe distance   & \cite{Bando-1995,batista2010optimal, jin2014dynamics,lazar2016review}
		\\ \cline{2-6}
		
		
		& Gazis-Herman-Rothery (GHR) & $h,\Delta v ,v$ & $a$ & \tcck{-}  & \cite{gazis1961nonlinear, zhu2020impact} 
		\\ \cline{2-6}
		
		& Gipps & $h,\Delta v ,v$  & $v$ & desired deceleration; desired acceleration; desired speed, etc  & \cite{gipps1981behavioural, wilson2001analysis} 
		\\ \cline{2-6}
		
		& Helly & $h, \Delta v, v$ & $a$ & safe distance  & \cite{helly1959simulation} 
		\\ \cline{2-6}
		

		& Full velocity difference Model (FVDM) & $h, \Delta v, v$ &  $a$  & maximum velocity; safe distance  & \cite{jiang2001full,jin2010non} \\ \cline{2-6}
		
		
		
		& Wiedemann     /Fritzsche model & $h, \Delta v, v$ &  $a$  & stand still spacing and acceleration; desired time headway; additional safety distance, etc   & \cite{ wiedemann1992microscopic,fritzsche1994model, durrani2016calibrating,wang2017comparison}
		\\ \cline{2-6}

		& Optimal control & $h, \Delta v, v,a$ &  $a$   & desired time headway; minimum spacing; safe distance  & \cite{zhou-2017,zhou2019robust,zhou2019distributed}
		\\ \hline\hline

		
		\multirow{4}{*}{\rotatebox[origin=c]{90}{Data-Driven}}  
		& SVM & $h, \Delta v, v$ & $v$ & - & \cite{wei2013analysis} \\ \cline{2-6}	
		
		& ANN & $h, v$ & $a $ & - & \cite{Panwai2007} \\ \cline{2-6}	
		
		& LSTM  /RNN & A time series of $(h, \Delta v, v)$, images & $a / v$  & - & \cite{Zhou2017g,huang2018car,gu2020lstm} \\ \cline{2-6} 
		
		& KNN & $h, \Delta v, v$ & $v$  & -  & \cite{he2015simple} \\ \cline{2-6}	
		
		
		 & DDPG & $h,\Delta v,v$ & $a$   & - &\cite{NN_DRL_zhu2018human}  \\\cline{2-6}

		& GAIL & $h, \Delta v, v$ & $a$  & -& \cite{kuefler2017imitating, zhou2020modeling} \\\hline\hline
		
		\multirow{3}{*}{\rotatebox[origin=c]{90}{Hybrid}}  
		& Gipps  + ANN & $h, \Delta v, v$ & $a$ & maximum velocity  & \cite{yang2018novel}\\ \cline{2-6}

		& \textcolor{black}{Physics Regularized Gaussian Process} & $x,v,a$ & $a$ & Parameters of GHR, Gipps, and other CFMs  & \cite{yuan2020modeling}\\ \cline{2-6}
		
		& PIDL & $h, \Delta v, v$ & $a$ & Parameters of the IDM, OVM, GHR, and FVDM  & this paper\\ \hline
		
		\end{tabular}
		
		\begin{tablenotes}
      \footnotesize  
      \item Note. $x$, $h$, $v$, $a$ are the position, spacing, velocity, and acceleration of the subject vehicle, respectively. $\Delta v$ is the velocity difference between the subject vehicle and its leader. 
      (In the ``Physics Model Parameters'', only parameters related to a physical meaning are listed. Parameters like constant coefficients sensitivity parameters are not listed.) 
    \end{tablenotes}
    \end{threeparttable}
    
\end{table}

\subsubsection{Physics-based CFM}

Calibration of a physics-based CFM can be formulated as an optimization problem below: 

\begin{equation}\label{equ:generic_CFM_optimization}
\begin{split}
    \min_\theta \quad & \: \sum_{t=\Delta t}^{T} (a(t) - \hat{a}(t))^2 \\
    \textrm{s.t.} \quad & a(t+\Delta t) = f_{\theta}(\hat{\mathbf{s}}(t) | \theta), \quad t=0, \Delta t,\ldots,T,
\end{split}
\end{equation}
where, \\
$\hat{a}(t)$: ground-truth acceleration; \\
$a(t)$: estimated acceleration from a predefined mapping;\\
$\Delta t$: prediction horizon;\\
$T$: time horizon of one trajectory;\\ 
$f_{\theta}$: the mapping function, which can be any physics-based CFM, like the Intelligent Driving Model (IDM) or the Optimal Velocity Model (OVM), which are listed in Table~\ref{tab:map}. 

\subsubsection{DL-based CFM}\label{sec:method:nn}

\begin{figure}[h]
    \centering
    \includegraphics[width=0.4\columnwidth]{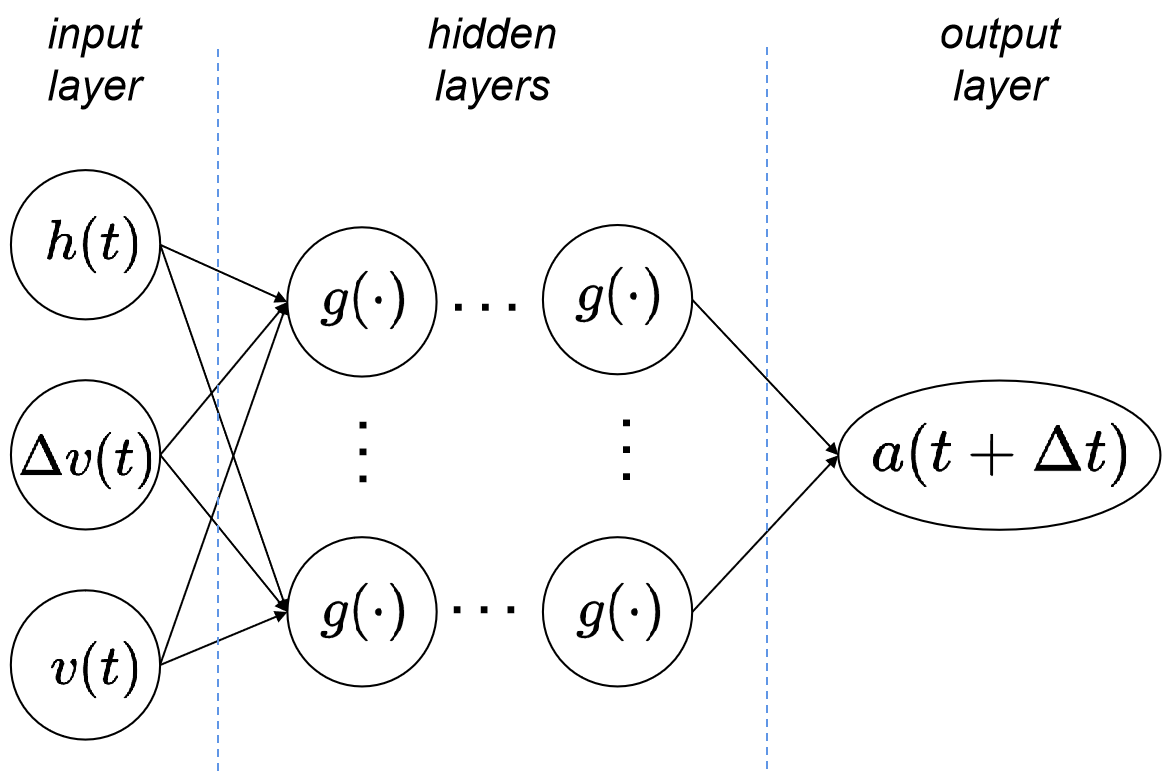}
    \caption{Structure of the DL-based CFM.}
    \label{fig:fnn}
\end{figure}

The artificial neural network (ANN) is a widely used function approximator of car-following behavior. It takes the states as input and the acceleration as output. The general architecture of an ANN is shown in Fig.~\ref{fig:fnn}. 
In each layer, each node linearly combines the outputs from the last layer, and passes it to the next layer after a non-linear activation function $g(\cdot)$ is applied. 
For example, if the $j$ th node in the $i$ th layer is connecting with all the nodes in the $i-1$ th layer, its output is

\begin{equation}
    O^{(i)}_j = g(I^{(i)}_j) = g( \sum_k w^{(i-1)}_{k,j} \cdot O^{(i-1)}_{k} + b^{(i)}_j),
\end{equation}
where, \\
$I^{(i)}_j$: input of the $j$th node in the $i$th layer. For example, $I^{(1)}_1$ denotes the spacing $h(t)$.\\
$O^{(i)}_j$: output of the $j$th node in the $i$th layer;\\
$O_k^{(i-1)}$: output of the $k$th node in the $i-1$th layer in a neural network. \\
$b^{(i)}_j$: bias of the $j$th node in the $i$th layer; \\
$w^{(i-1)}_{k,j}$: weights stored in the link between the $k$ th node in the $i-1$ th layer and the $j$ th node in the $i$ th layer \\
$g(\cdot)$: activation function.\\

\subsection{Literature on PIDL}

This paper aims to integrate both a mathematical formula and a neural network into one framework to achieve a better prediction in car-following behavior, which is physics-informed deep learning (PIDL). 
\tcck{
PIDL has become a mainstream research topic and its notion bears variants like ``theory-guided data science" \citep{Karpatne-2017}, ``model-informed machine learning" \citep{yu2021model}, and ``physics-informed deep learning" \citep{raissi2019physics}.
Its popularity lies in that it leverages the strengths of both physics-based and data-driven models. 
More importantly, it can overcome the challenges of training data-hungry DL models particularly arising from
limited data, 
high cost of data acquisition, 
imperfect data (such as missing data, outliers, complex noise processes),
and large parameter spaces of neural networks.  
Physics are usually encoded as \emph{governing equations}, \emph{physical constrains}, or \emph{regularity terms} in deep neural networks' loss functions \citep{alber2019integrating}. 
}
In the pioneering work~\citep{Raissi-2018a, Raissi-2018b}, PIDL was proposed as an alternative solver of partial differential equations (PDE). 
Since its inception, PIDL has become an increasingly popular tool for data-driven solution or discovery of nonlinear dynamical systems 
\tcck{
(including identification of a functional form of a physics-based model and calibration of associated parameters defined in a physics-based model) 
}
in various engineering areas \citep{Yang-2019,Maziar-2019,Fang-2020,Maziar-2020,rai2020driven}. 
There is also a gradually rising trend to encode ordinary differential equations (ODE) into neural networks for approximation of ODE solutions \citep{di2020finding} or identification of dynamical system models \citep{roehrl2020modeling}. 

While PIDL has increasingly demonstrated its predictive power in various fields, transportation modeling is lagging behind in combining both models and data aspects. 
In transportation, 
\cite{shi2020aaai,shi2020pidl} encoded 
traffic flow models into neural networks for traffic state estimation. 
The application of PIDL to data-driven solution of car-following dynamical equations or system identification of such behavior, however, is a largely unexploited area.
A related study~\citep{Wu-2018} used neural networks of different structures to capture the general behavior of a car-following model and predict the acceleration of a vehicle using velocity and distance information. \textcolor{black}{This hybrid method leverages the prior knowledge and outperforms the NN-based algorithm.}  
\cite{yang2018novel} combined the Gipps model with DL-based models (back-propagation neural networks and random forest, respectively) and optimal weights of these two models \tcck{are selected} for collision avoidance of predicted trajectories. \textcolor{black}{The combined model can reconstruct trajectories that balance authenticity and safety.} \tcck{In this combined model, Gipps model is not encoded into the DL-based model, and they are independently trained instead.}
\tcck{Apart from NN-based model}, \cite{yuan2020modeling} introduced a physics regularized Gaussian process model by incorporating traditional CFMs \tcck{into} a Gaussian Process model to predict driver accelerations. Their experiments show that the developed model outperforms the pure Gaussian process model in terms of both prediction accuracy and robustness. \tcck{
In this study, we propose two ways of hybridizing the physics into the neural networks, in which the physics is encoded into the NN structure to regularize the training of the NN.}  

\subsection{Contributions of this paper}

Human driving exhibits highly unstable and nonlinear behaviors \citep{huang2019stable,huang2020stable} and accordingly, physics-based models alone may not suffice to reveal highly nonlinear nature of driving behavior, leading to high bias. 
The generic architecture of a PIDL consists of two computational graphs: 
one neural network (i.e., the data-driven component) for predicting the unknown solution, 
while the other (i.e., the physics-driven component), in which physics 
are encoded into NN, for evaluating whether the prediction aligns with the given physics. 
The physics encoded computational graph can be treated as a regularization term of the other deep neural network to prevent overfitting, i.e., high-variance. 
In summary, the hybrid of both components overcomes high-bias and high-variance induced by the individual ones, rendering it possible to leverage the advantage of both the physics-based and data-driven methods in terms of model accuracy and data efficiency. 

\tcck{
The PIDL paradigm opens up a promising research direction in driving behavior modeling.
However, the design of how physics can be encoded remains an open question that varies domain by domain. 
This paper establishes the PIDL paradigm using two-computational graphs for the customization of PIDL to the identification of car-following behavior. 
We will explore how existing physics-based CFMs can be leveraged in deep neural networks for improved car-following behavior training and prediction. 
}
Specifically, the main contributions of this paper include:

\begin{itemize}

    \item Design PIDL architectures encoded with \tcck{2} popular physics-based models - IDM and OVM, on which 
    acceleration is predicted for 4 traffic regimes: acceleration, deceleration, cruising, and emergency braking; We also include other physics-based models - \tcck{GHR and FVDM, to investigate how different physics components affect the performance of PIDL.} 

\item 
\tcck{
Identify parameters used in CFMs by solving the PIDL problem of joint state prediction and model parameter discovery. 
Most existing PIDL studies assume physics-based models are calibrated prior to the training of neural networks, which may lead to larger errors if physics are not calibrated accurately. Thus we introduce the joint estimation problem and show that joint estimation leads to a higher estimation accuracy than the prediction only one.  
}

\item \tcck{Demonstrate the generalization of PIDL methods by adopting two different NN structures: 
ANN and LSTM. The results show that PIDL holds the potential to enhancing the training efficiency and prediction accuracy of existing deep learning methods by incorporating prior physical knowledge into the training process.  }

\item Demonstrate the superiority of PIDL over pure physics-based and data-driven models using a comprehensive set of numerical experiments and real-world data (i.e., Next Generation SIMulation (NGSIM)). 
\tcck{This hybrid paradigm outperforms its two individual components across various training data, especially when measurements are sparse.}

\end{itemize}

\section{PIDL for Identification of Car-Following Behavior}\label{sec:methodology}

In this section, we provide a generic problem statement of PIDL for CFM, which includes two types of problems: one is acceleration-prediction only 
and the other is joint acceleration-prediction and model-parameter-discover. 

\subsection{Problem statement}

Subsequently we denote the observed quantities with a hat, such as $\hat{s}, \hat{a}$, indicating that their values are observed. 

Define two disjoint sets $\cal O$ and $\cal C$ where $\cal{O}\cap \cal C =\emptyset, \cal O\cup \cal C\subseteq S$. 
$\cal O$ represents the observed state space and $\cal C$ the (unobserved) collocation state space. 
They both consist of a set of 3-dimensional state vectors ${\cal O}=\{\hat{\mathbf{s}}^{(i)}=(\hat{h}^{(i)},\Delta \hat{v}^{(i)}, \hat{v}^{(i)}), i=1,...,N_o\}, {\cal C}=\{\mathbf{s}^{(j)}=(h^{(j)},\Delta v^{(j)}, v^{(j)}), j=1,...,N_c\}$,  
where $N_o$ and $N_c$ are the size of observed states and collocation states, respectively. 
Denote $\cal{A}$ as the observed action set, which contains scalar accelerations ${\cal A} =\{\hat{a}^{(i)}, i=1,..,N_o\}$, where $N_o$ is the size of observed states. Note that the observed action set shares the same size as the observed state set, because it is a one-to-one mapping. 
A pair of observed state and action indexed by $i\in \{1,...,N_o\}$ constitute one observation data, denoted as $\{\hat{\mathbf{s}}^{(i)}, \hat{a}^{(i)}\}\in \cal{O}\times \cal{A}$. 

\begin{note}
The existing literature uses various variables as input, but in this paper we only use three variables, including the spacing $\hat{h}(t)$, velocity difference $\Delta \hat{v}(t)$ and the velocity $\hat{v}(t)$ of the subject vehicle at current time step. 
The PIDL framework can be easily generalized to include higher dimensional input vectors. 
\end{note}

Define a physics-uninformed neural network (PUNN) ${\cal N}_{\theta}$ parameterized by a hyperparameter vector ${\theta}$ that defines a function approximator, $f_{\theta}(\hat{\mathbf{s}}| \theta)$, which maps observed states to actions, which is, ${\cal N}_{\theta} \triangleq f_{\theta}(\hat{\mathbf{s}} | \theta): \hat{\mathbf{s}}\in {\cal O} \longrightarrow a$.
Define a physics-based computational graph ${\cal N}_{\lambda}$ parameterized by a physics-based parameter vector $\lambda$ that characterizes a physics-based mapping, $f_{\lambda}(\mathbf{s} | \lambda)$, which maps collocation states to actions, which is, ${\cal N}_{\lambda} \triangleq f_{\lambda}(\mathbf{s} | \lambda): \mathbf{s}\in {\cal C} \longrightarrow a$. 


With all the notations, we are ready to present the PIDL-CF framework in the context of car-following.

\begin{defn}
\textbf{Physics-informed deep learning for modeling car-following (PIDL-CF).} 
Assume we have the observed states ${\cal O}$, the observed actions ${\cal A}$, and the collocation points ${\cal C}$ defined below
\vspace{-0.1in}
\begin{equation}
\label{equ-3-x1}
\begin{split}
\left\{ {\begin{array}{*{20}l}
    \  {\cal O}=\{\mathbf{\hat{s}}^{(i)}=(\hat{h}^{(i)},\Delta \hat{v}^{(i)}, \hat{v}^{(i)}), i=1,..,N_o\}\subseteq {\cal S}, \ \  \\
   \ {\cal A}=\{\hat{a}^{(i)}, i=1,..,N_o\}, \  \  \\
   \ {\cal C}=\{\mathbf{s}^{(j)}=(h^{(j)},\Delta v^{(j)}, v^{(j)}), j=1,...,N_c\} \subseteq {\cal S}. 
\end{array}} \right.\
\end{split}, 
\end{equation}
with the design of two neural networks: 
(1) a PUNN, denoted as $f_{\theta}(\hat{\mathbf{s}}| \theta)$, for car-following mapping approximation $a^{(i)}$, 
and (2) a physics-based computational graph, denoted as $f_{\lambda}(\mathbf{s} | \theta)$, for computing physics-based acceleration of $a_{phy}^{(j)}$. 

A general PIDL-CF model, which denoted as $f_{\theta,\lambda}(\hat{\mathbf{s}}|\theta,\lambda)$, is to train an optimal parameter set $\theta^*$ for PUNN and an optimal parameter set $\lambda^*$ for the physics. 
The PUNN parameterized by the solution $\theta^{*}$ can then be used to predict accelerations $\hat{a}_{new}$ on a new set of observed states ${\cal O}_{new}\subseteq {\cal S}$, and $\lambda^{*}$ is the most likely model parameters that describe the intrinsic physics of observed data. 
\end{defn}


\begin{defn}
\textbf{Two problems for PIDL-CF.} 
\begin{enumerate}
    \item 
\textbf{Acceleration-prediction-only problem} (``prediction-only" for short): 
The prediction-only problem aims to train a PIDL model, encoded with a known physics-based model, that predicts its acceleration. It can be formulated as an optimization problem:
\begin{equation}\label{equ:generic_CFM_optimization_PIDL_preOnly}
 \begin{split}
     \min_\theta \quad &  \sum_{i=1}^{N} (a^{(i)} - \hat{a}^{(i)})^2 \\
     \textrm{s.t.}  \quad & a^{(i)} = f_{\theta}(\mathbf{\hat{s}}^{(i)} | \theta,\hat{\lambda}), \quad i = 1,\ldots, N,
 \end{split}
 \end{equation}
where $\hat{a}_i$ denotes the actual observed acceleration 
and $\hat{\lambda}$ is the pre-calibrated model parameter set. 


\item 
\textbf{Joint acceleration-prediction and model-parameter-discovery problem} (``joint estimation" for short): 
The joint prediction-and-calibration problem not only predicts accelerations, but also estimates model parameters $\lambda$ used in the physics-based models. It can be formulated as an optimization problem:

\begin{equation}\label{equ:generic_CFM_optimization_PIDL_jointEst}
 \begin{split}
     \min_{\theta,\lambda} \quad &  \sum_{i=1}^{N} (a^{(i)} - \hat{a}^{(i)})^2 \\
     \textrm{s.t.} \quad & a^{(i)} = f_{\theta}(\mathbf{\hat{s}}^{(i)} | \theta,\lambda), \quad i = 1,\ldots, N, \\
     & \lambda \subseteq \Lambda.
 \end{split}
 \end{equation}
     where $\Lambda$ is the feasible domain of the parameters of the \tcck{physics} $\lambda$, representing the physical range of each physics parameter.
\end{enumerate}
\end{defn}

In the rest of the section, we will elaborate on the two types of PIDL-CF problems and detail each component one by one, including the design of neural network architectures, selection of data, loss functions, and the development of training algorithms.

\subsection{PIDL-CF for the prediction-only problem}

In the prediction-only problem, the parameters appearing in the pre-selected physics-based CFM are calibrated beforehand using the available training data. 
In other words, model parameter calibration 
needs to be done before a PIDL model is trained. 

\subsubsection{PIDL-CF architecture}

Physics-informed deep learning, apparently, should consist of a physics component and a deep learning component in the form of neural networks. 
That being said, 
a generic PIDL-CF model is comprised of two elements: a physics-uninformed neural network (PUNN) and a \tcck{physics-based computational graph}. 
PUNN is a deep neural network,  
while the \tcck{physics-based computational graph} is encoded with a physics-based CFM (e.g., IDM or OVM). 
Both components share the same structure of input and output, both taking a state vector as input and output an acceleration quantity.

\begin{figure}[h!]
    \centering
    \includegraphics[width=0.4\columnwidth]{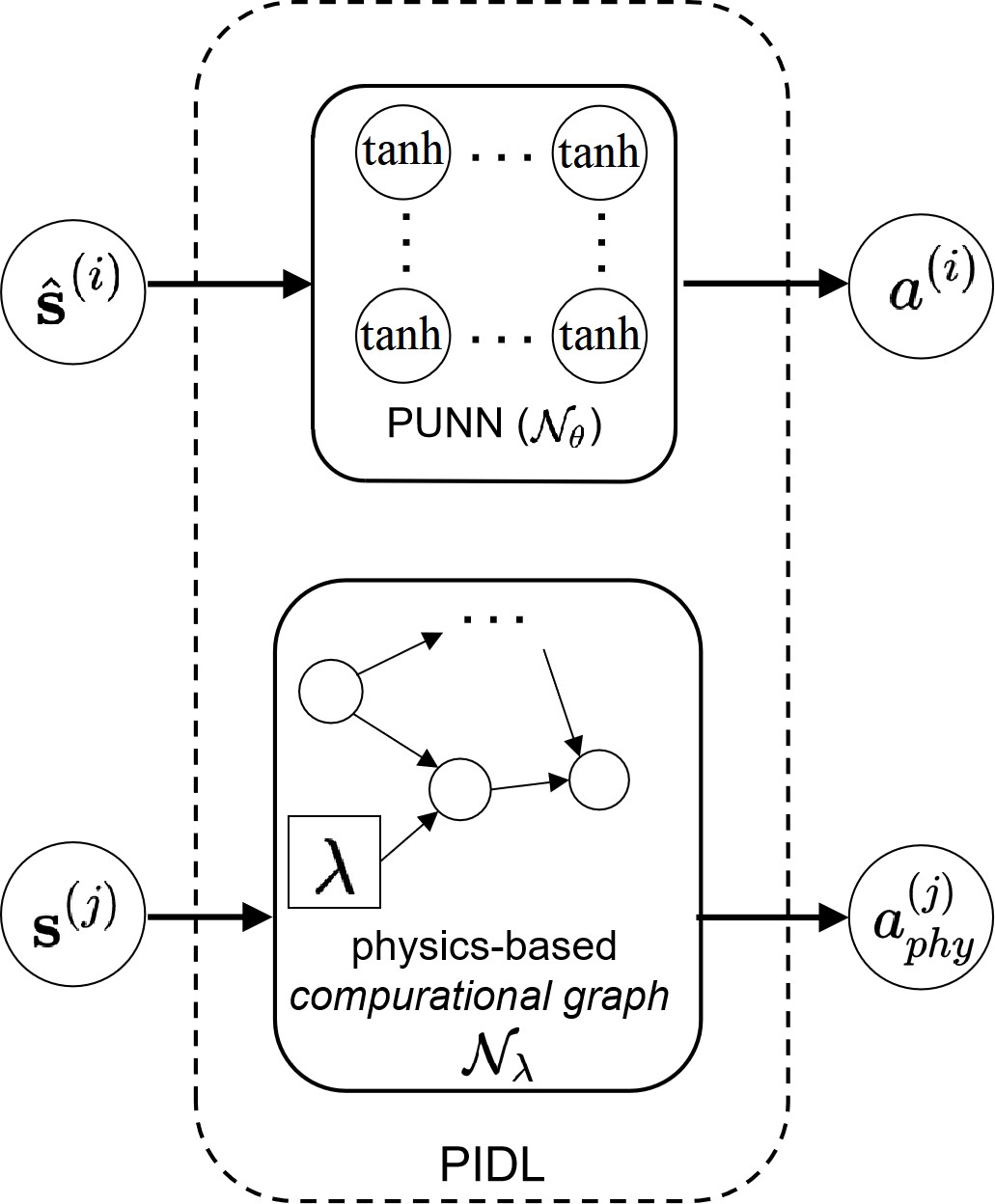}
    \caption{Architecture of PIDL.}
    \label{fig:struct_pidl}
\end{figure}
Fig.~\ref{fig:struct_pidl} illustrates the structure of a PIDL-CF model. The upper part is the PUNN, in which each node represents a neuron associated with an activation function and each edge is attached with a parameter that needs to be trained using observations. The bottom part is the \tcck{physics}. Each circle node represents a variable or a derived intermediate variable from the physics-based model, each rectangle represents a model parameter that can be either pre-calibrated or jointly trained, and each edge is encoded with a known operator.

In this paper, we adopt ANNs for PUNN, but note that other types of neural networks, like recurrent neural networks, can also be adopted. We choose a hyperbolic tangent ($tanh$) function as the activation function because it outperforms other functional forms when ANNs are used for supervised learning tasks \citep{karlik2011performance}, such as prediction of acceleration in our context.

\subsubsection{Observations and collocation states}

In this subsection we will introduce the selection of the observed data and collocation states, and how to split them into training, validation and test data for training a PIDL-CF model. 
\textcolor{black}{Those data would be used to calculate the loss function and to train the PIDL-CF model, which will be detailed in Sec.~\ref{prediction-only-alg}}.

Observed data is a set of state-action pairs: $\{\hat{\mathbf{s}}^{(i)}, \hat{a}^{(i)}\}\in \cal{O}\times \cal{A}$, where the state is a vector of spacing, velocity difference, and velocity, while the action is acceleration. All these values are extracted from observed trajectory data, which is a time-series data containing positions of a leading vehicle and a following vehicle. From this trajectory profile, velocity and acceleration can be computed by taking the first and the second derivatives of its position with respect to time, respectively. Then the state at each time step is paired with the acceleration at the next time step to form an observation data: $\{\hat{\mathbf{s}}^{(i)}, \hat{a}^{(i)}| i=1,...,N_O\} = \{\hat{\mathbf{s}}(t), \hat{a}(t+\Delta t)| t=0,\Delta t,...,T\}$. \tcck{After pairing, observation data is shuffled to break the time-dependency among each other}.


\begin{figure}[b!]
    \centering
    \includegraphics[width=1\columnwidth]{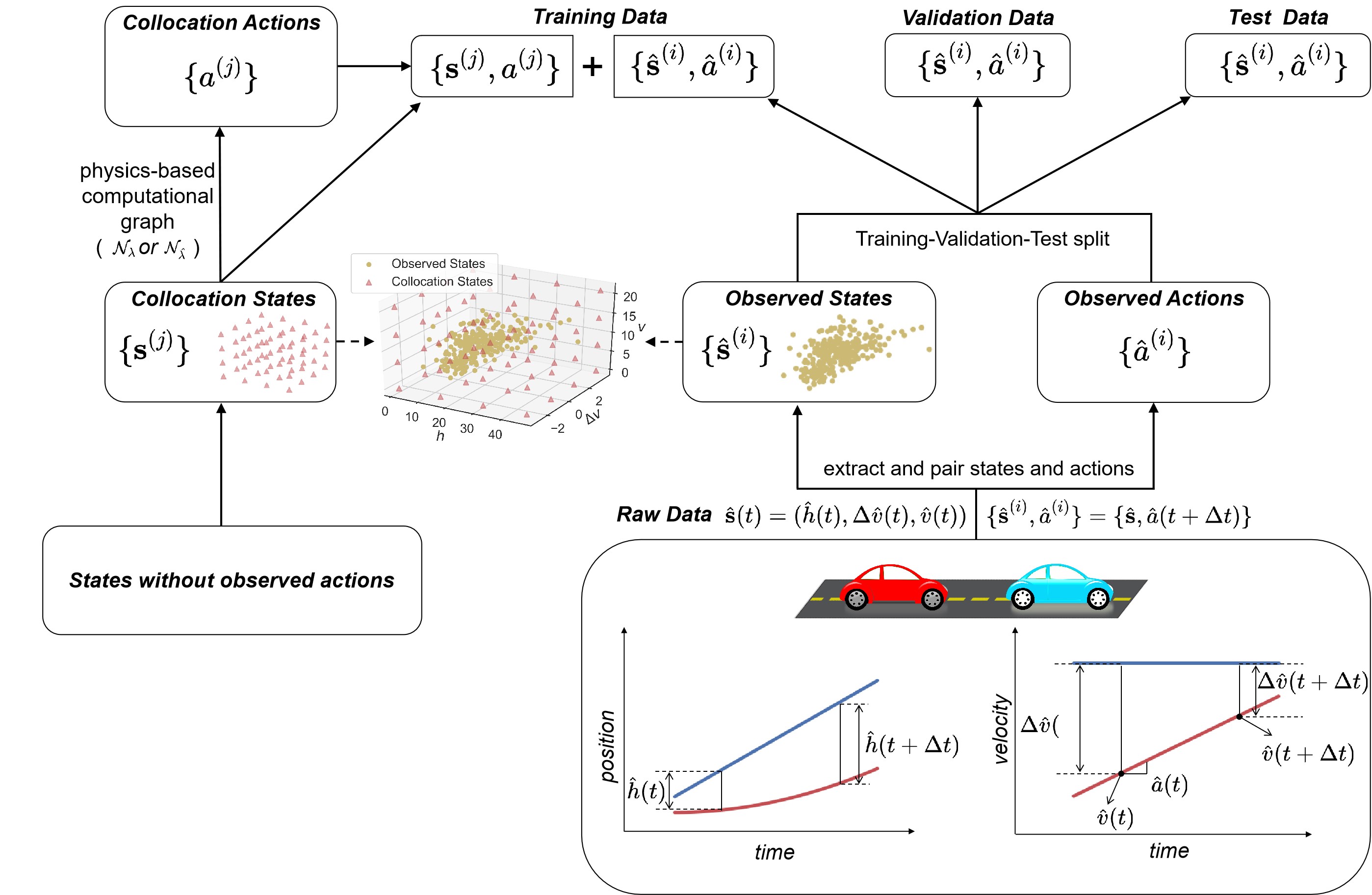}
    \caption{Relationship between raw trajectory data, observed data, collocation data, and training-validation-test split.}
    \label{fig:data_stuff}
\end{figure}



Other than these standard datasets,  we need to define a new set of collocation points that are particularly unique to the PIDL framework. 
The purpose is to compute one's acceleration from physics-based models over these pre-defined collocation points. 
The common practice is to select collocation states based on physics knowledge. 
For example, when the PIDL model was applied to discover the data-driven solution of a PDE whose inputs are a 2-dimensional time-location vectors over the temporal-spatial domain \citep{Raissi-2018a}, the collocation points are selected as points on the temporal-spatial mesh grid. 
The size of the collocation points usually depends on the granularity of the temporal-spatial grid as well as the size of the observed data \citep{Raissi-2018a, Raissi-2018b,raissi2019physics}. 
In the CF context, a collocation point is the state whose action is not observed, which is denoted as a ``collocation state'' in the rest of this paper. Because no observed action is paired, the selection of collocation states is not constrained by any actually observed data. In other words, a collocation state does not necessarily need to be observed in any raw trajectory data. 
Since the target value is not observed corresponding to a specific collocation state, 
the acceleration value of one collocation state is computed using a physics-based CFM calibrated with the observed data. 
We denote the collocation pair as $\{\mathbf{s}^{(j)}, a^{(j)}_{phy}|j=1,...,N_C\}.$ 

Observed data and collocation data need to be further grouped into training, validation, and test data for the PIDL-CF training. 
The observed data is shuffled and split into three subsets following a certain training-validation-test split ratio. 
The collocation data are all counted towards the training data. 
In summary, the total training dataset is composed of one subset of the observed data and all collocation data. 
The other two subsets of the observed data are used for the purpose of validation and test, respectively.

Fig.~\ref{fig:data_stuff} illustrates the relationship among all the datasets mentioned above.
Let us start from the bottom-right box enclosing a pair of leader-follower vehicles. Suppose the red car is following the blue car and both their position and velocities are observed. We can then extract the state and the action, and then pair them to constitute one observed data, shown in the two middle boxes. A training-validation-test split is then applied to further divide this observed data pair into three subsets, shown in the three boxes on the top. 
Now we will inspect the bottom-left box enclosing a set of collocation states. The collocation states can be pre-selected uniformly on a mesh grid over the state space. 
These collocation states, along with their computed acceleration values from some physics-based CFM \textcolor{black}{that has been calibrated using the observed data}, constitute the other part of the training data.


\subsubsection{Loss function}

With observed and collocation data defined, we are ready to define the loss function associated with PUNN and the \tcck{physics}. 
The loss function is composed of two components, 
one to minimize the discrepancy between the observed acceleration and that predicted from PUNN (i.e., data discrepancy), 
and the other to minimize the discrepancy between physics-based acceleration prediction and that computed from PUNN (i.e., physics discrepancy). 
Specifically, the data discrepancy is evaluated as the Mean Square Error (MSE) between the acceleration approximated by PUNN on the observed data and its ground-truth value. The physics discrepancy is computed using the MSE between the accelerations approximated by PUNN and the \tcck{physics}. 
Accordingly, we propose a loss function as a weighted sum of these two MSEs as follows:

\begin{align}\label{eqn:loss}
\begin{split}
\operatorname{Loss}_{\theta} &=   \alpha \cdot M S E_{O}+ \left(1-\alpha\right) \cdot M S E_{C} \\
&=   \alpha \cdot \frac{1}{N_{O}} \sum_{i=1}^{N_{O}}\left|a^{(i)} - \hat{a}^{(i)} \right|^{2} 
+(1-\alpha)\cdot \frac{1}{N_{C}} \sum_{j=1}^{N_{C}}\left| a^{(j)} - a_{phy}^{(j)}\right|^{2} \\
& = \alpha \cdot  \underbrace{\frac{1}{N_O} \sum_{i=1}^{N_O} \left| f_{\theta}(\hat{\mathbf{s}}^{(i)}|\theta)-\hat{a}^{(i)} \right|^2}_{\text {data discrepancy}} 
+
(1-\alpha) \underbrace{ \frac{1}{N_C} \sum_{j=1}^{N_C} \left|f_{\theta}(\mathbf{s}^{(j)}|\theta) - f_{\hat{\lambda}}(\mathbf{s}^{(j)}|\hat{\lambda}) \right|^2}_{\text {physics discrepancy}} 
\end{split}
\end{align}
\\
where,\\
$\alpha$: the weight of the loss function that balances the contributions made by the data discrepancy and the physics discrepancy;\\
$\hat{\lambda}$: parameters of the \tcck{physics} that are calibrated beforehand;\\
$\theta$: PUNN's parameters;\\
$f_{\theta}(\cdot)$: PUNN parameterized by $\theta$;\\
$f_{\hat{\lambda}}(\cdot)$: \tcck{physics} parameterized by $\hat{\lambda}$, which is calibrated prior to training;\\
$N_O$: the number of observed data; \\
$N_C$: the number of collocation states; \\
$\hat{\mathbf{s}}^{(i)}$: the $i$th observed feature; \\
$\mathbf{s}^{(j)}$: the $j$th collocation feature; \\
$\hat{a}^{(i)}$: the $i$th observed acceleration;\\ 
$a^{(i)}$: the $i$th acceleration predicted by PUNN;\\
$a^{(j)}$: the $j$th acceleration predicted by PUNN;\\
$a_{phy}^{(j)}$: the $j$th acceleration predicted by the \tcck{physics}.

\tcck{Note that in the Appendix. A we show that the PIDL-CFM framework is agnostic of the training metrics. MSE is selected as it is commonly used in modeling CF behaviors.}

\subsubsection{Training algorithm} \label{prediction-only-alg}
Training PIDL-CF in the prediction-only problem consists of two subsequent processes, the calibration of the \tcck{physics} and the training of PUNN.

\paragraph*{Calibration of the \tcck{Physics}}
 \tcck{In the prediction-only problem,} \textcolor{black}{physics-based CFM is calibrated using the observed training data}. 
 Calibration of \tcck{physics} is essentially the parameter estimation of a traditional CFM using the observed data, which has been extensively studied in the transportation literature. In accordance with the generic CF problem statement described in Eq.~\ref{eqn: generic_CFM}, we can write the calibration of \tcck{physics} as an optimization problem as follows:
\begin{equation}\label{equ:generic_CFM_optimization_PIDL_calibration}
 \begin{split}
     \min_\lambda \quad &  Obj = \frac{1}{N_O}\sum_{i=1}^{N_O} \left|a_{phy}^{(i)} - \hat{a}^{(i)}\right|^2\\
     \textrm{s.t.}  \quad & a_{phy}^{(i)} = f_{\lambda}(\hat{\mathbf{s}}^{(i)} |\lambda), \quad i = 1,\ldots, N_O,\\
     & \lambda \subseteq \Lambda,
 \end{split}
 \end{equation}
The objective function $Obj$ measures the discrepancy of the acceleration estimated by the \tcck{physics-based CFM} from its ground-truth value in an MSE form. \tcck{To distinguish with the ``loss function" that will be introduced in the training of PUNN (prediction-only problem) and the joint training of the PUNN and physics (joint-estimation problem), in this paper ``objective function'' is used for the calibration of physics.}

The calibration algorithms can be divided into two categories: gradient-based and gradient-free. Gradient-based algorithms 
include Least Square (LS) estimation \citep{ossen2005car}, maximum likelihood estimation \citep{treiber2013microscopic}, and sequential quadratic programming \citep{he2019weight}. 
Gradient-free algorithms use heuristic methods that do not require any knowledge of gradient. 
The mostly used gradient-free algorithm for solving CF problems is Genetic Algorithm (GA) \citep{kumar2010genetic}. 
In this paper, we will use both types of algorithms for different evaluation methods, which will be discussed in Section~\ref{sec:PIDL_for_ngsim}.

\paragraph*{Training of PUNN}
Once the physics-based model parameter $\hat{\lambda}$ is obtained, we are ready to train the PUNN's parameters $\theta$ using the loss function Eq.~\ref{eqn:loss}. 
We use an Xavier uniform initializer \citep{sirignano2019scaling_Xavier} to initialize the weights and biases of the PUNN. 
Then we use a popular stochastic gradient decent algorithm, the Adam optimizer \citep{NN_adam_nageshrao2019autonomous}, to update the parameters of PUNN, $\theta$. Gradients of the loss function with respect to 
the weights and biases are calculated using the auto-differential function in 
Tensorlow \citep{abadi2016tensorflow}. 
To avoid overfitting, we adopt an early-stopping criteria on the validation dataset: 
the training process is terminated when the minimum validation error $min(\text{MSE}_{val})$ does not change for a certain number of iterations. PIDL-CF parameters are stored at each iteration, and parameters with the minimum $\text{MSE}_{val}$, which are called the optimal parameters $\theta^*$, are recovered when the early-stopping is triggered. Then the recovered PIDL-CF model receives the final evaluation on the test data. 
The training process is detailed in Algorithm~\ref{alg:punn}.

\begin{algorithm}[H]
\label{alg:punn}
\SetAlgoLined
\LinesNumbered
\KwResult{Learned PUNN parameters $\theta^*$}
Initialization:
 {\begin{enumerate}[{(1)}]
     \item {optimal parameters of the \tcck{physics-based CFM} $\hat{\lambda}$ calibrated beforehand,} 
     \item {initialized PUNN parameters $\theta_{0}$ using Xavier,}
     \item {loss weight $\alpha \in [0,1]$, learning rate $lr$}. 
 \end{enumerate}
 }
 
 
 \While{not reaching early-stopping criteria} 
 {
  Calculate $\nabla_{\theta} \text{Loss}_{\theta}$ using the auto-differential technique;
  
  {
  Update the parameters using the gradient decent: $\theta_{k+1} = \theta_{k} + lr\nabla_{\theta} \text{Loss}_{\theta}$;

  }

  Calculate $\text{MSE}_{val}$ using PUNN with the updated parameters $\theta_{k+1}$. 
 }
 
 \caption{Training algorithm for PUNN}
 
\end{algorithm}

\noindent where,\\
$lr$: learning rate, which determines the step-size when updating parameters;\\
$\theta_k$: PUNN parameters at the $k$th iteration.

\subsection{PIDL for the joint-estimation problem}

In the joint-estimation problem, a PIDL model is trained to jointly predict acceleration and calibrate parameters used in a physics-based model. 

Because the model architecture and data structure of the joint-estimation problem is the same as the prediction-only problem, we primarily focus on the loss function and the training algorithm.

\subsubsection{Loss function}

\begin{align}\label{eqn:loss_join}
\begin{split}
\operatorname{Loss}_{\theta,\lambda} 
& = \alpha MSE_O + (1-\alpha) MSE_C \\
&=   \alpha \cdot \frac{1}{N_{O}} \sum_{i=1}^{N_{O}}\left|a^{(i)} - \hat{a}^{(i)} \right|^{2}
+(1-\alpha)\cdot \frac{1}{N_{C}} \sum_{j=1}^{N_{C}}\left| a^{(j)} - a_{phy}^{(j)}\right|^{2}, \\
& = \alpha \cdot \underbrace{ \frac{1}{N_O} \sum_{i=1}^{N_O} \left| f_{\theta}(\hat{\mathbf{s}}^{(i)}|\theta)-\hat{a}^{(i)} \right|^2 }_{\text {data discrepancy}} 
+
(1-\alpha) \underbrace{ \frac{1}{N_C} \sum_{j=1}^{N_C} \left|f_{\theta}(\mathbf{s}^{(j)}|\theta) - f_{\lambda}(\mathbf{s}^{(j)}|\lambda) \right|^2}_{\text {physics discrepancy}} 
\end{split}
\end{align}

where $f_{\lambda}(\cdot)$ is the \tcck{physics} parameterized by $\lambda$, which is trained along with the PUNN's parameters $\theta$.

\subsubsection{Training algorithm} \label{joint_train_alg}
Similarly, we use the Adam optimizer with an early-stopping condition to train the PUNN and estimate physics parameters simultaneously. While some modifications should be made. 

First, the parameter of the \tcck{physics-based CFM} usually has physical meaning, and its range varies depending on the specific physics knowledge this parameter stands for. 
By contrast, parameters of the PUNN do not carry physical meanings. 
So the magnitudes of both parameters and gradients in the PUNN and the \tcck{physics-based CFM} may differ, resulting in different training speeds and convergence. We solve this issue by using different learning rates for the PUNN and the \tcck{physics-based CFM}, denoted as $lr_\text{PUNN}$ and $lr_{phy}$, respectively. This provides flexibility of balancing the training on the PUNN part and on the \tcck{physics} part. 

A second issue during the training process is, some parameters of the \tcck{physics-based CFM} encounter significantly large gradients, which likely lead to values beyond their physical bounds. 
To cope with this problem, we set bounds over the gradients of the \tcck{physics-based CFM}, and then clip the calculated gradients if they are beyond the bounds. The lower and upper bounds of the gradients of the \tcck{physics-based CFM} are denoted as $minClip$ and $maxClip$, which are the minimum and maximum value the gradients of the \tcck{physics-based CFM} can achieve. A summary of the joint-training of the \tcck{physics-based CFM} and PUNN is shown in Algorithm~\ref{alg:punn_pinn}. We use bold texts to indicate the main differences from Algorithm~\ref{alg:punn}.

\begin{algorithm}[H]
\label{alg:punn_pinn}
\SetAlgoLined
\LinesNumbered
\KwResult{Learned PUNN parameters $\theta^*$ and parameters of the \tcck{physics-based CFM} $\lambda^*$}
Initialization:
 {\begin{enumerate}[{(1)}]
     \item {initialized parameters of the \tcck{physics-based CFM} $\lambda_0$,}
     \item {initialized PUNN parameters $\theta_{0}$ using Xavier,}
     \item {loss weight $\alpha \in [0,1]$, \textbf{learning rates for two networks} $lr_{phy}$ \textbf{and} $lr_{\text{PUNN}}$, and tolerance of the early-stop $tol$;}
     \item {\textbf{upper and lower bounds of the gradient of the \tcck{physics-based CFM}, $minClip$ and $maxClip$}}
 \end{enumerate}
 }
 
 
 \While{ not reaching early-stopping criteria }{
  {Calculate $\nabla_{\theta} \text{Loss}_{\theta,\lambda}$ and $\nabla_{\lambda} \text{Loss}_{\theta,\lambda}$ using the auto-differential technique, \textbf{and clip the gradient of the \tcck{physics parameter}} $\nabla_{\lambda} \text{Loss}_{\theta,\lambda} \longleftarrow clip(\nabla_{\lambda} \text{Loss}_{\theta,\lambda}, minClip, maxClip)$}

  {
  Update the parameters of \textbf{two networks} using the gradient decent: $\theta_{k+1} = \theta_{k} + lr_{\text{PUNN}}\nabla_{\theta} \text{Loss}_{\theta,\lambda}$, $\lambda_{k+1} = \lambda_{k} + lr_{\text{phy}}\nabla_{\lambda} \text{Loss}_{\theta,\lambda}$;
  }

  {Calculate $\text{MSE}_{val}$ using PUNN with the updated parameters $\theta_{k+1}$.} 
 }
 \caption{Joint-training algorithm for the \tcck{physics-based CFM} and PUNN}
\end{algorithm}

\subsection{Prediction and evaluation of PIDL-CF}
The output of the training process is a learned PUNN with parameters $\theta^*$ and a learned \tcck{physics-based CFM} with parameters $\hat{\lambda}$ or $\lambda^*$, depending on whether the \tcck{physics-based CFM} is trained by calibration or joint-training. Then the trained PUNN, $\cal{N}_{\theta^*}$, can be used for the prediction on a new dataset, which is the test data in this paper. 
By feeding the states at the current timestep into $\cal{N}_{\theta^*}$, it outputs the accelerations at the next timestep.  
A test MSE is used to evaluate the performance of our PIDL-CF models on the test data:

\begin{align}\label{eqn:test_mse}
\begin{split}
MSE_{\text{test}} 
& =  \frac{1}{N_{\text{test}}} \sum_{i=1}^{N_{\text{test}}} \left| a^{(i)}- \hat{a}^{(i)} \right|^2\\
& = \frac{1}{N_{\text{test}}} \sum_{i=1}^{N_{\text{test}}} \left| f_{\theta^*}(\hat{\mathbf{s}}^{(i)}|\theta^*) - \hat{a}^{(i)} \right|^2,
\quad i=1,...,N_{\text{test}}.
\end{split}
\end{align}\\
\noindent where,\\
$N_{test}$: the number of test data;\\
$f_{\theta^*}(\cdot)$: PUNN parameterized by $\theta^*$.

\subsection{Summary of PIDL-CF}

\begin{figure}[h!]
   \centering
   \subfloat[][]{\includegraphics[width=1\textwidth]{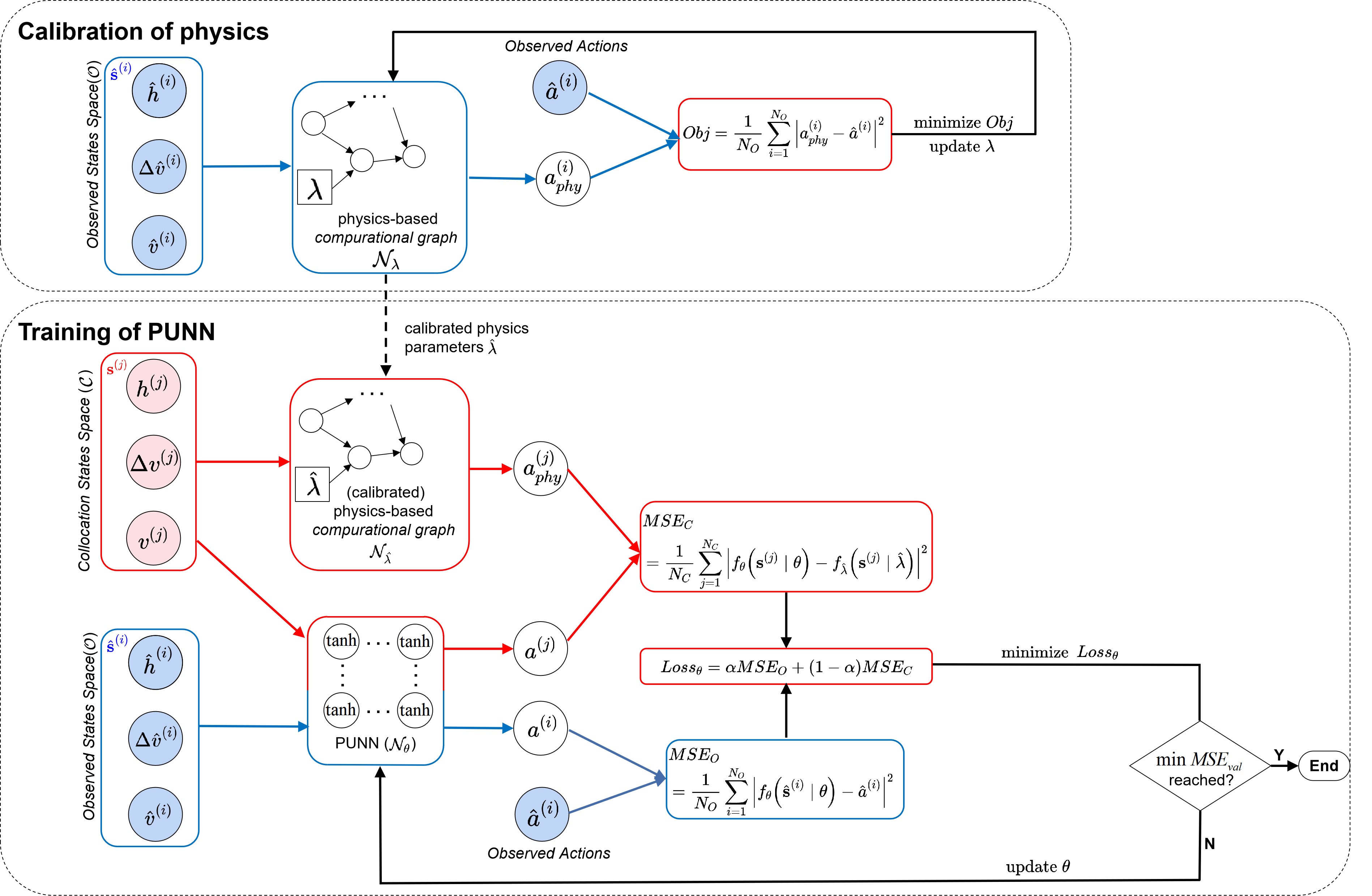}} 
   \hfill

   \subfloat[][]{\includegraphics[width=1\textwidth]{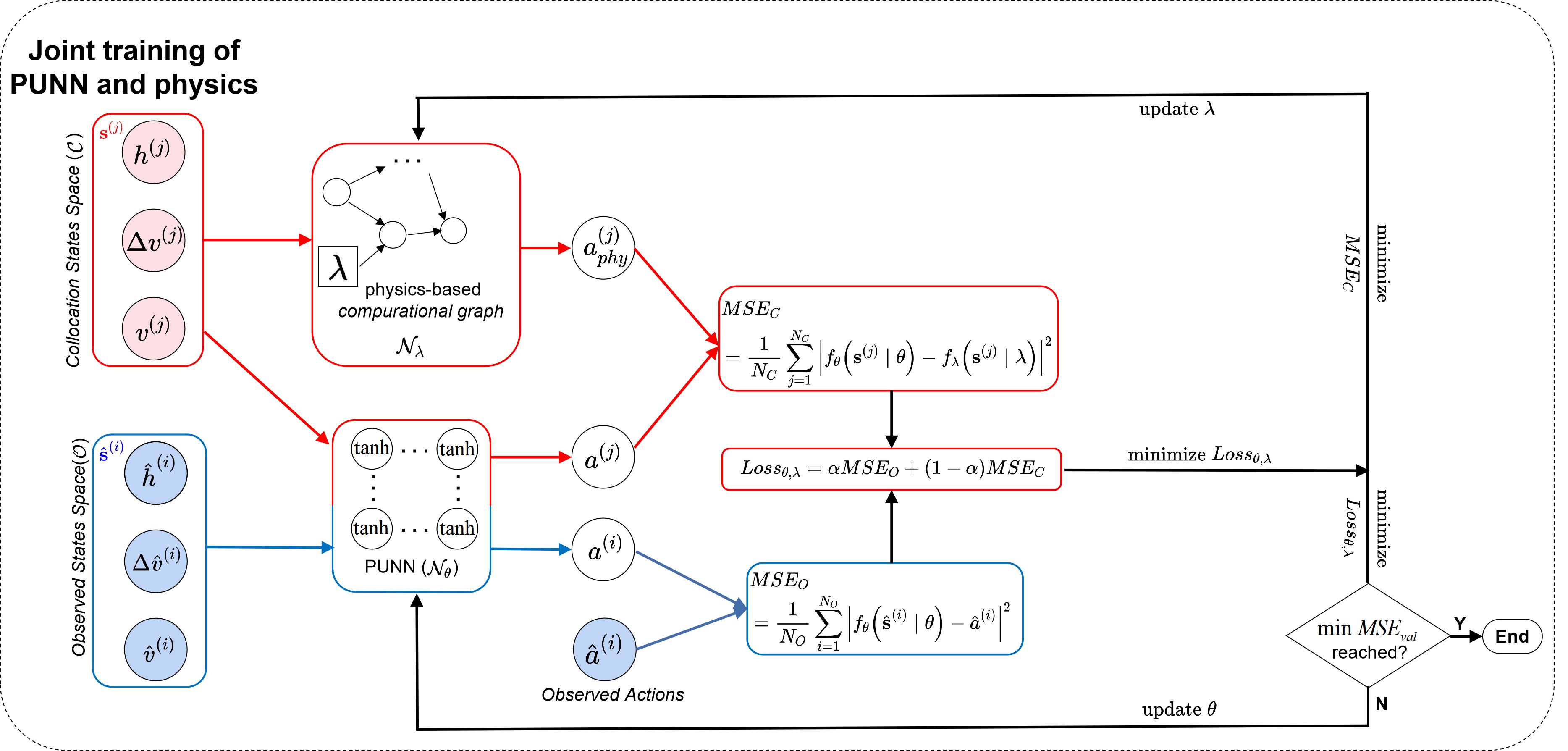}}
   \caption{Generic structure of PIDL-CF for (a) prediction-only problem and (b) joint-estimation problem.}
   \label{fig:generic_pidl}
\end{figure}

Now we sum up all the components that are detailed sequentially in the previous subsections. A complete picture of the generic structure of PIDL-CF is illustrated in Fig.~\ref{fig:generic_pidl}, \tcck{Fig.~\ref{fig:generic_pidl}(a) for the prediction-only problem and Fig.~\ref{fig:generic_pidl}(b) for the joint-estimation problem.}
Note that the time index is omitted as we consider the one-step prediction.

Fig.~\ref{fig:generic_pidl}(a), for the prediction-only problem, is composed of two parts: the upper part demonstrates the flow of the calibration of the physics, and the lower part aims to train the PUNN with observed data, collocation states, and the actions predicted by the calibrated physics on the collocation states. We will introduce each element of Fig.~\ref{fig:generic_pidl}(a) from top to bottom.  In the upper part, the colored circular nodes represent inputs and other nodes represent intermediate quantities. The leftmost box encloses observed states ($\hat{\mathbf{s}}^{(i)}$), which is a 3-dimensional vector consisting of spacing, velocity difference, and velocity. These nodes are colored to indicate that these states are observed. The observed states are fed into the physics-based computational graph, which outputs the acceleration $a_{phy}^{(j)}$ predicted by the encoded physics. This predicted acceleration is then compared to the actual observed acceleration $\hat{a}^{(i)}$ to compute the calibration objective function $Obj$ that is depicted in Eq.~\ref{equ:generic_CFM_optimization_PIDL_calibration} , which is used to update the physics parameters $\lambda$. The calibrated physics parameters $\hat{\lambda}$ is then used for the training of PUNN in the lower part. In the lower part, the leftmost two boxes enclose the observed states ($\hat{\mathbf{s}}^{(i)}$) and collocation states ($\mathbf{s}^{(j)}$) from top to bottom. 
Nodes of the collocation states are colored as red to indicate that they are given beforehand. The observed states are fed into PUNN, while the collocation states are fed into both PUNN and \tcck{physics-based CFM}. 
PUNN outputs two different acceleration quantities. 
The first one, $a^{(i)}$, is the approximate acceleration computed out of the observed states $\hat{\mathbf{s}}^{(i)}$;
The second one, $a^{(j)}$, is the approximate acceleration computed using collocation states $\mathbf{s}^{(j)}$. 
The \tcck{physics-based CFM} outputs one quantity, $a_{phy}^{(j)}$, by using the physics-based model with input as collocation states. 
The approximate acceleration $a^{(i)}$ based on the observed states is then compared to the actual observed acceleration $\hat{a}^{(i)}$ to compute $MSE_O$. 
The approximate acceleration $a^{(j)}$ based on the collocation states is then compared to the physics predicted acceleration $a_{phy}^{(j)}$ to compute $MSE_C$. 
Both $MSE_O$ and $MSE_C$ define the loss function $Loss_{\theta}$, which is used to train the parameters $\theta$. 
Based on $Loss_{\theta}$ computed from observed data, we train the PUNN.  
In training, we use $MSE_{val}$ to check early-stop and avoid overfitting. 
Once the optimal epoch number is found that minimizes the validation error, the training process stops.

Fig.~\ref{fig:generic_pidl}(b) illustrates the joint training problem of both the PUNN and the physics. Most of the elements share the same meaning as those in Fig.~\ref{fig:generic_pidl}(a) except for the training of physics parameters $\lambda$. The first difference is the box enclosing the physics-based computational graph (${\cal N_{\lambda}}$). As the physics is not calibrated prior to the training, the physics-based computational graph is parameterized by $\lambda$ instead of the calibrated parameters $\hat{\lambda}$. As $\lambda$ are trainable parameters in the joint-estimation problem, the corresponding physics discrepancy $MSE_C$ is also trainable with respect to $\lambda$. As a result, \tcck{the loss function $Loss_{\theta, \lambda}$ depends on the physics parameters $\lambda$ as well}. The second difference is that the physics parameters $\lambda$ are trained simultaneously along with the PUNN parameters $\theta$. In each training epoch, $\lambda$ is updated to minimize $Loss_{\theta, \lambda}$ (by minimizing $MSE_C$ as $MSE_O$ is unrelated to $\lambda$). In other words, the values of the physics parameters $\lambda$ is updated during the training process, so is $a_{phy}^{(j)}$ which is predicted by the physics.


The mathematical formulation of two PIDL-CF problems are summarized below:
\begin{enumerate}
    \item 
A general PIDL for the prediction-only problem is to solve the optimization problem below:
\begin{equation}
\label{equ-3-x2}
\begin{split}
\begin{gathered}
\mathop {\min }\limits_\theta  Loss_{\theta}\\
where \ \ \ \ \ \ \ \ \ \ \ \ \ \ \ \ \ \ \ \ \ \ \ \ \ \ \ \ \ \ \ \ \ \ \ \ \ \ \ \ \ \ \ \ \ \ \ \ \ \ \ \   \\ Loss_{\theta}   = \alpha \cdot \frac{1}{N_{O}} \sum_{i=1}^{N_{O}}\left|a^{(i)} - \hat{a}^{(i)} \right|^{2}
+(1-\alpha)\cdot \frac{1}{N_{C}} \sum_{j=1}^{N_{C}}\left| a^{(j)}- a_{phy}^{(j)} \right|^{2}\\
= \alpha \cdot \frac{1}{N_O} \sum_{i=1}^{N_O} \left| f_{\theta}(\hat{\mathbf{s}}^{(i)}|\theta)-\hat{a}^{(i)} \right|^2 
+ 
(1-\alpha)  \frac{1}{N_C} \sum_{j=1}^{N_C} \left|f_{\theta}(\mathbf{s}^{(j)}|\theta) - f_{\hat{\lambda}}(\mathbf{s}^{(j)}|\hat{\lambda}) \right|^2.
\end{gathered}
\end{split}
\end{equation}

\item 
A general PIDL for the joint-estimation problem is to solve the optimization problem below:

\begin{equation}
\label{equ-3-x3}
\begin{split}
\begin{gathered}
\mathop {\min }\limits_{\theta,\lambda} Loss_{\theta,\lambda}\\
\textrm{s.t. } \lambda \subseteq \Lambda, \\
where \ \ \ \ \ \ \ \ \ \ \ \ \ \ \ \ \ \ \ \ \ \ \ \ \ \ \ \ \ \ \ \ \ \ \ \ \ \ \ \ \ \ \ \ \ \ \ \ \ \ \ \  \\ Loss_{\theta,\lambda}  = \alpha \cdot \frac{1}{N_{O}} \sum_{i=1}^{N_{O}}\left|a^{(i)} - \hat{a}^{(i)} \right|^{2}
+(1-\alpha)\cdot \frac{1}{N_{C}} \sum_{j=1}^{N_{C}}\left| a^{(j)}- a_{phy}^{(j)} \right|^{2}\\
= \alpha \cdot \frac{1}{N_O} \sum_{i=1}^{N_O} \left| f_{\theta}(\hat{\mathbf{s}}^{(i)}|\theta)-\hat{a}^{(i)} \right|^2 
+ 
(1-\alpha)  \frac{1}{N_C} \sum_{j=1}^{N_C} \left|f_{\theta}(\mathbf{s}^{(j)}|\theta) - f_{\lambda}(\mathbf{s}^{(j)}|\lambda) \right|^2.
\end{gathered}
\end{split}
\end{equation}

\end{enumerate}

Table.~\ref{table:difference_pred_joint} summarizes the difference between the prediction-only and joint-estimation problems in updating parameters $\lambda$ and $\theta$ during the calibration and training processes. The first column of this table lists two problems for the PIDL, i.e., the prediction-only and joint-estimation problem. The remaining columns (the first 2 columns for calibration process and the remaining 4 columns for training process) address two questions as to the calibration or training of PIDL parameters $\lambda$ and $\theta$: 1) whether parameters are updated and 2) which \tcck{states and actions} are needed when updating the parameters. The parenthesis after each \tcck{quantity} indicates whether it is fixed or variant during the calibration or training process. The observed data ($\{ \hat{\mathbf{s}}^{(i)}, a^{(i)}\}$) and the collocation states ($\mathbf{s}^{(j)}$) are observed or given beforehand, so are fixed throughout the calibration and training processes. $a^{(i)}$ is the output of PUNN, whose value changes with the training process while PUNN parameters $\theta$ are updated. As for the output of physics-based computational graph, which is $a_{phy}^{(j)}$, its value is fixed during the training process in the prediction-only problem, because the physics parameters are already calibrated and fixed to be $\hat{\lambda}$. In the joint-estimation problem, the value of $a_{phy}^{(j)}$ varies with training, because the physics parameters $\lambda$ are updated during the training process.

\begin{table}[h!]
\begin{center}

\caption{\label{table:difference_pred_joint}Summary of differences between prediction-only and joint-estimation problems}
\begin{tabular}{ |c||c|c|c|c|c|c|  } 
\hline

\multirow{2}{*}{\textbf{Problem}} &  \multicolumn{2}{c|}{\textbf{Calibration}} & \multicolumn{4}{c|}{\textbf{Training}} \\
\cline{2-7}
& \makecell{ If $\lambda$ \\ is updated} & \makecell{States \\ and actions \\ needed \\ when \\ updating $\lambda$} & \makecell{ If $\lambda$ \\ is updated} & \makecell{ States \\ and actions \\ needed \\ when \\ updating $\lambda$} & \makecell{ If $\theta$ \\ is updated} & \makecell{ States \\ and actions \\ needed \\ when \\ updating $\theta$}\\
\hline\hline

\makecell{Prediction\\-only} & Yes
                            & \makecell{$\hat{\mathbf{s}}^{(i)}$(fixed)\\
                                            $\hat{a}^{(i)}$(fixed) 
                                            }  
                            & \makecell{ No. Value \\
                                        is fixed \\
                                        to be            $\hat{\lambda}$}  
                            &  - 
                            & Yes  
                            & \makecell{$\hat{\mathbf{s}}^{(i)}$ (fixed)\\
                                            $\hat{a}^{(i)}$(fixed) \\
                                            $\mathbf{s}^{(j)}$(fixed) \\
                                            $a_{phy}^{(j)}$(fixed)}  
                            \\
\hline
\makecell{Joint\\-estimation} & -
                                & -
                                & Yes
                                & \makecell{$\mathbf{s}^{(j)}$(fixed) \\ $a^{(j)}$(variant)}
                                & Yes 
                                & \makecell{$\hat{\mathbf{s}}^{(i)}$ (fixed)\\
                                            $\hat{a}^{(i)}$(fixed) \\
                                            $\mathbf{s}^{(j)}$(fixed) \\
                                            $a_{phy}^{(j)}$(variant)} 
                                \\
\hline

\end{tabular}

\end{center}
\end{table}

\section{PIDL for Existing Physics-based CFMs} \label{sec:numerical}

To demonstrate the performance of the PIDL-CF framework, in this and the next sections, we will first apply it to simulated data generated from widely used physics-based car-following models (in Section~\ref{sec:numerical}) and then move to the real-world dataset (in Section~\ref{sec:real_exp}). 
In other words, in the former case, we assume we know the data-generating mechanism, which is the known physics-based models. 
Then we use simulated data generated from these models (plus some artificial noise) as input to our PIDL-CF model, train the model, and then compare the performances to baselines. 
These simulated data from known models are homogeneous, free of errors, and are generated under controlled and ideal traffic environments, thus laying a stepping stone to the evaluation of PIDL-CF with real-world noisy data.  
\textcolor{black}{
More importantly, since we know the ground-truth physics (in other words, the car-following model type) used to generate these simulated data, we can directly encode these functional forms into our PIDL-CF framework, 
which would control test errors that may arise from using a wrong physics. 
Moreover, we can also compare directly estimated physics parameters output from our PIDL-CF to the ground-truth, without worrying if the selected physics-based model is inaccurate per se. 
When it comes to real-world datasets in Section~\ref{sec:real_exp}, we do not know the intrinsic physics, i.e., the car-following model type, that generates these trajectories and the only observation we have are the actual acceleration values. Therefore, we try out different physics-based models and compare the predicted accelerations output from the PIDL-CF encoded with various physics. 
}

In this section, we focus on two physics-based models, namely, IDM and OVM, which generate ground-truth trajectories of the following vehicle. These simulated trajectories, added with artificial noise, will be used as input data to our PIDL model. 
For both the IDM based PIDL-CF model (PIDL-IDM for short) and the OVM based PIDL-CF model (PIDL-OVM for short), the prediction-only problem and the joint-estimation problem will be solved. In the prediction-only problem, the predicted acceleration output from our PIDL model will be compared with the ground truth and physics-uninformed neural networks to justify the advantage of including physics. In the joint-estimation problem, the estimated parameters will be compared with the ground-truth parameters and the prediction error will also be shown, justifying that it can do accelerations prediction and parameters discovery simultaneously. 

We briefly introduce the experiment settings of the PIDL-IDM and PIDL-OVM models in these two problems, which are summarized in Table~\ref{table:PIDL_config_numerical}. The first two columns of this table list the names of the relevant parameters along with their notations, and the remaining columns are their values. Each row represents one parameter associated with either the model formula or the training process. 
These setting will be revisited in the subsequent subsections. 

\begin{table}[h!]
\begin{center}

\caption{\label{table:PIDL_config_numerical}Experiment settings of the PIDL-IDM model and the PIDL-OVM model for the simulated data}
\begin{tabular}{ |c|c||c|c|c|c|  } 
\hline

\multirow{2}{*}{\textbf{Name}} & \multirow{2}{*}{\textbf{Notation}}  & \multicolumn{2}{c|}{\textbf{PIDL-IDM}} & \multicolumn{2}{c|}{\textbf{PIDL-OVM}} \\
\cline{3-6}
& & \makecell{ prediction \\ -only} & \makecell{ joint \\ -estimation} & \makecell{ prediction \\ -only} & \makecell{ joint \\ -estimation}\\
\hline\hline

\makecell{number of observed data\\ used in training ($\times 20$)} & $n_O$ &  \makecell{1, 5, 9, \\ 13, 15, 20, \\ 30, 40, 50} & \makecell{1, 5, 9\\ 13, 15, 20} &  \makecell{1, 5, 9, \\ 13, 15, 20, \\ 30, 40, 50}  & \makecell{1, 5, 9\\ 13, 15, 20}  \\
\hline

\makecell{number of collocation data\\ used in training ($\times 20$)} & $n_C$ & 1 &  9 & 5 & 9 \\
\hline
weight of the loss function &$\alpha$ &\makecell{ 0.1, 0.4,\\0.7, 1.0} &  0.7 & \makecell{ 0.1, 0.4,\\0.7, 1.0} &  0.7 \\
\hline
\makecell{training-validation-test \\ split ratio} & -  & \multicolumn{4}{c|}{0.5:0.25:0.25}\\
\hline

\multirow{2}{*}{learning rate}  & $lr_{\text{PUNN}}$  &  \multicolumn{4}{c|}{0.001} \\
\cline{2-6}
                                 & $lr_{\text{phy}}$  & - & 0.1 & - & 0.1\\
\hline
PUNN hidden layers size &-&  \multicolumn{4}{c|}{(60,60,60)}\\
\hline
\multirow{5}{*}{\makecell{ ground-truth parameters \\ of the \tcck{physics}}}& \multirow{5}{*}{$\lambda$}  & \multicolumn{2}{c|}{$v_0$=30 $m \cdot s^{-1}$} & \multicolumn{2}{c|}{$v_{max}$=30 $m \cdot s^{-1}$ }  \\
                &  &\multicolumn{2}{c|}{$T_0$=1.5 $s$} & \multicolumn{2}{c|}{$h_c$=10 $m$}   \\
                &   &\multicolumn{2}{c|}{$s_0$=2 $m$} & \multicolumn{2}{c|}{$k$=0.03 $s^{-1}$} \\
                &    &\multicolumn{2}{c|}{$a_{max}$=0.73 $m \cdot s^{-2}$} & \multicolumn{2}{c|}{}  \\
                &   &\multicolumn{2}{c|}{$b$=1.63 $m \cdot s^{-2}$} & \multicolumn{2}{c|}{}  \\
\hline

traffic regimes & - & \makecell{four separate\\regimes} & \multicolumn{3}{c|}{\makecell{all regimes combined}}\\

\hline
\makecell{pre-calibration \\ method} & - & \makecell{ground-truth \\ is given} & - & \makecell{ground-truth \\ is given} & -   \\

\hline

\end{tabular}

\end{center}
\end{table}

\begin{figure}[h!]
   \centering
   \subfloat[][]{\includegraphics[width=1\textwidth]{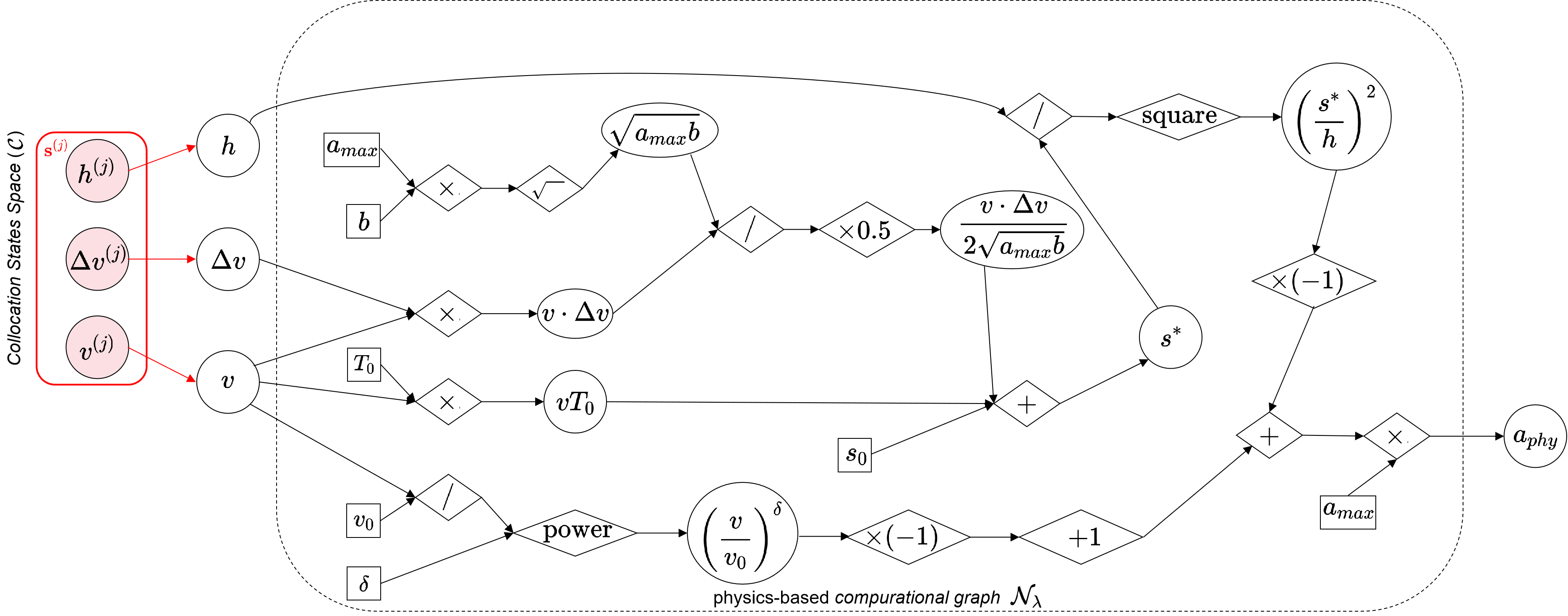}} 
   
   \subfloat[][]{\includegraphics[width=1\textwidth]{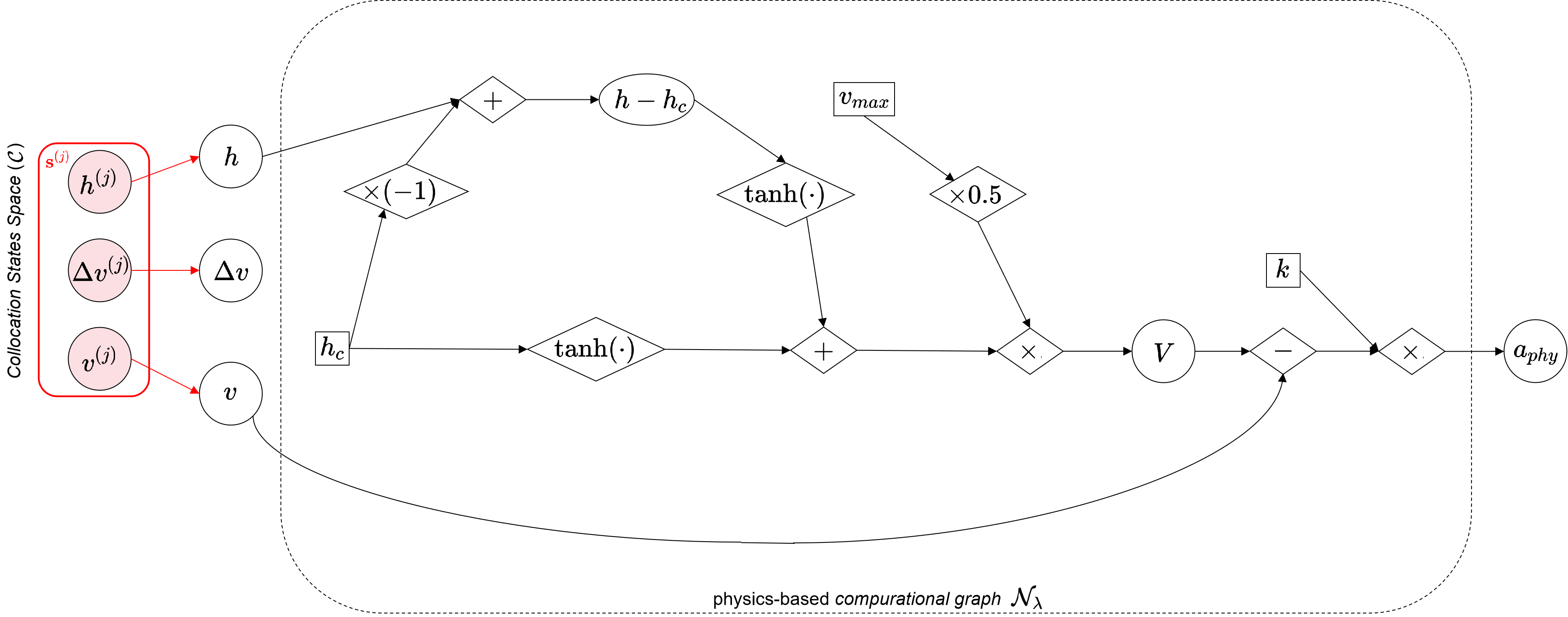}}
   \caption{Physics-based computational graph architecture of the PIDL-CF models. (a) is the physics-based computational graph of the PIDL-IDM model, which is encoded from an IDM model: $ a(t+\Delta t) = a_{max}[ 1-(\frac{v(t)}{v_0}) ^\delta - (\frac{s^*(v(t), \Delta v(t))}{h(t)})^2]$, $s^*(v(t),\Delta v(t)) =s_0 + v(t) T_0 + \frac{v(t) \cdot \Delta v(t)}{2\sqrt{a_{max}b}}$. (b) is the physics-based computational graph of the PIDL-OVM model, which is encoded from an OVM model: $a(t+\Delta t) = k\left[V\left(h\left(t\right)\right) - v(t)\right]$, $
    V(h(t)) = 0.5v_{max}[\text{tanh}(h(t) - h_c)+\text{tanh}(h_c)]$. }
   \label{fig:PIDL_IDM_OVM}
\end{figure}

\subsection{IDM-based physics}


We first apply the PIDL-IDM model on the numerical data generated from a known IDM model. We will introduce the architecture of the PIDL-IDM model, its experiment setting, and the results of both the prediction-only and the joint-estimation problems.

\subsubsection{IDM model specification} \label{sec:idm_eqn}

IDM model \citep{Treiber-2000} 
 is formulated as: 
\begin{equation}\label{eqn:idm}\left\{\begin{array}{l}
    a(t+\Delta t)   = a_{max}[ 1-(\frac{v(t)}{v_0}) ^\delta - (\frac{s^*(v(t), \Delta v(t))}{h(t)})^2] \\
    s^*(v(t),\Delta v(t)) =s_0 + v(t) T_0 + \frac{v(t) \cdot \Delta v(t)}{2\sqrt{a_{max}b}},
\end{array}\right.\end{equation}
where,\\
$v_0$: desired velocity; \\
$T_0$: desired time headway; \\
$s_0$:  minimum spacing in congested traffic;\\
$a_{max}$: maximum acceleration; \\
$b$: comfortable deceleration; \\ 
$\delta$: a constant that is usually set to 4 for the car-following problem.

The values of model parameters that are used to generate ground-truth trajectories are summarized in the row ``\tcck{ground-truth parameters of the physics}'' of Table~\ref{table:PIDL_config_numerical}. 



\subsubsection{PIDL-IDM architecture}
The PIDL-IDM model consists of a PUNN and a physics-based computational graph encoded from an IDM model. The physics in the PIDL-IDM model is encoded from the IDM model, whose architecture is illustrated in 
Fig.~\ref{fig:PIDL_IDM_OVM}(a). Symbols of the collocation states are kept the same as in Fig.~\ref{fig:data_stuff}.
In the physics-based computational graph, each circle represents a variable or a derived intermediate variable.  
Each rectangle represents a model parameter, including $v_0$, $T_0$, $s_0$, $a_{max}$, $b$, and $\delta$. 
These parameters can be trained either by pre-calibration (in the prediction-only problem) or by jointly training (in the joint estimation problem), except for $\delta$ that is usually fixed as 4. 
Each diamond indicates a known operator, including plus ($+$), minus ($-$), multiplication ($\times$), division (/), square root ($\sqrt{\quad}$), power, square, adding one ($+ 1$), multiplying one half ($\times0.5$), and multiplying negative one ($\times(-1)$).  

\subsubsection{Experiment settings} \label{sec:pidl-idm-exp-set}
Numerical experiments are designed to justify the performance of the PIDL-IDM models in the prediction-only problem and the joint-estimation problem. 
In this subsection we will introduce the experiment settings with a focus on the generation of the training data, the generation of the collocation data, calibration of the physics and sensitivity analysis, 
which will be detailed below.

\paragraph*{Generation of training data}

For the PIDL-IDM prediction-only problem, 
to demonstrate the influence of traffic regimes on the performance of PIDL-IDM, we divide vehicle trajectories into 4 driving regimes, including accelerating ($a(t)>$0 $m\cdot s^{-2}$), decelerating ($-$2 $m\cdot s^{-2}<a(t)<$0 $m\cdot s^{-2}$), cruising ($a(t)=$ 0 $m\cdot s^{-2}$), and emergency braking ($a(t)=-2$ $m\cdot s^{-2}$) for the simulation period $\forall t\in [0,T]$, where $T$ is the maximum time duration. 
For the cruising regime, $T$ is set to be 20 seconds. For other regimes, the simulation ends when the subject vehicle's acceleration becomes zero.
In each driving regime, we train one PIDL-IDM model using only one type of trajectories and also test it on the same driving regime. 
We want to guarantee that training and test data share similar distributions and avoid the complication of distributional shift between these two datasets.

To generate a sufficient number of data points for each driving regime, we vary the initial spacing and the initial velocity of the subject vehicle while preserving the sign of its acceleration over time. 
In other words, the subject vehicle always chooses its acceleration from the same regime throughout the simulation. 
We also vary the initial velocity of the leading vehicle in each run, but once the velocity is fixed, the leading vehicle drives at a constant speed. Real-world trajectories of leading vehicles will be used in the next section and our goal in this section is to investigate how our PIDL-IDM model performs in different regimes, thus the impact of leading vehicle behavior should be removed. 

\begin{figure}[H]
   \centering
   \subfloat[][Position, velocity, and acceleration of the accelerating regime]{\includegraphics[width=\textwidth]{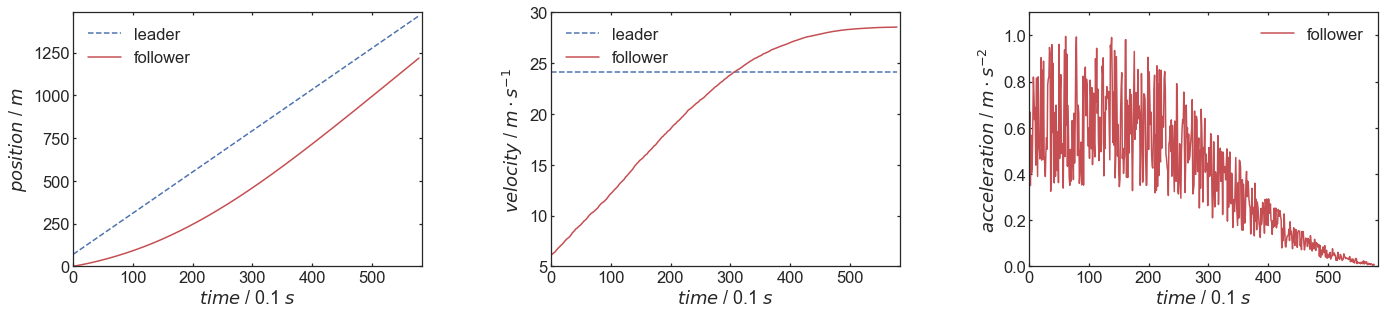}}\\
   \subfloat[][Position, velocity, and acceleration of the decelerating regime]{\includegraphics[width=\textwidth]{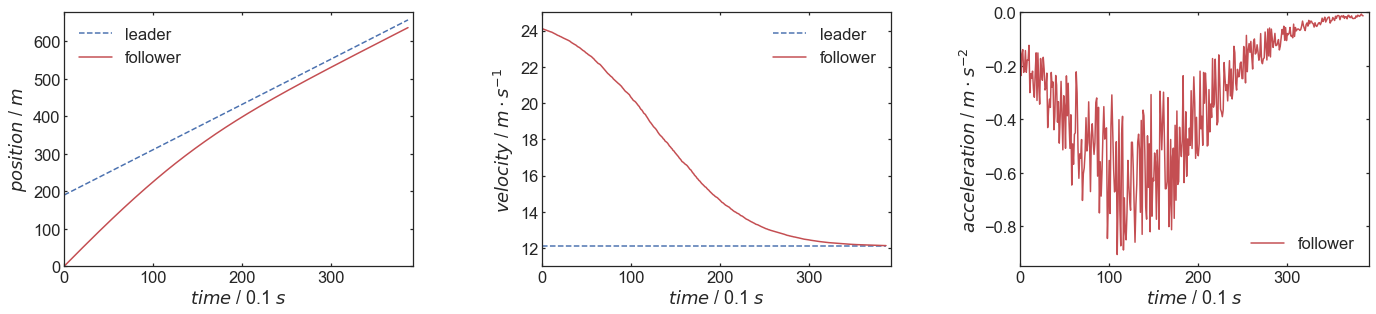}}\\
   \subfloat[][Position, velocity, and acceleration of the cruising regime]{\includegraphics[width=\textwidth]{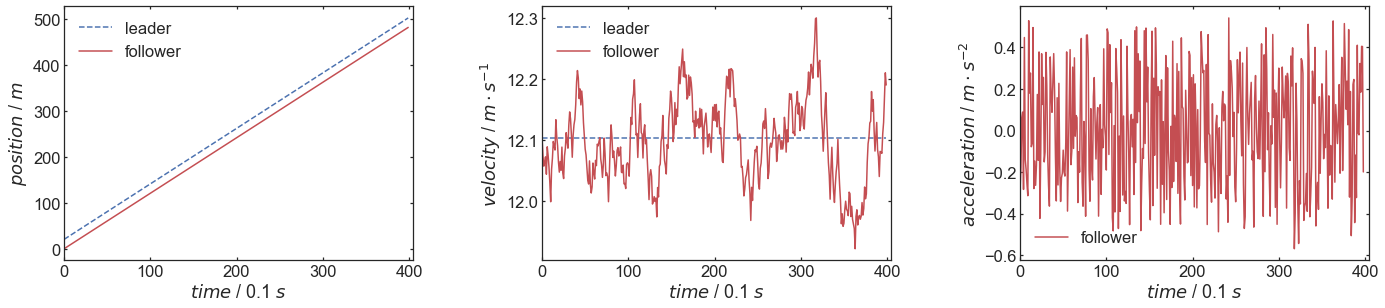}}\\
   \subfloat[][Position, velocity, and acceleration of the emergency braking regime. In this regime, we set a lower bound to the output of the IDM model, which is $-2$ $m\cdot s^{-2}$. ]{\includegraphics[width=\textwidth]{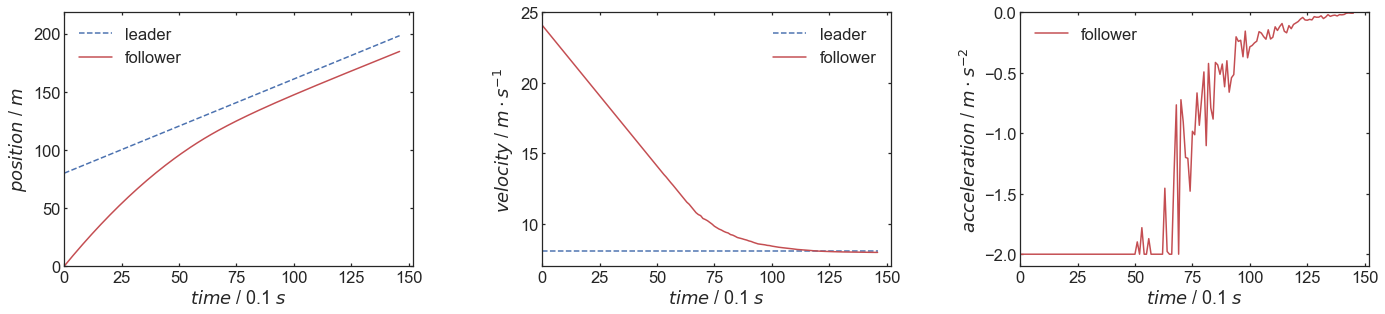}}\\
   \caption{Sample trajectories of 4 regimes. Each row represents accelerating, decelerating, cruising, and emergency braking respectively.
   Each column represents the position, velocity and the follower's acceleration, respectively.}
   \label{fig:numerical_traj}
\end{figure}

Fig.~\ref{fig:numerical_traj} illustrates sample trajectories generated from each regime. 
The figures on each row come from the same regime: 
the top row for the accelerating, 
the second row for decelerating, 
the third row for cruising,
and the bottom row for emergency braking. 
The figures on each column demonstrate one driving quantity of interest: from left to right are time-space trajectories, speed profile, and acceleration profile across time. 
We artificially add a Gaussian noise to the ground-truth trajectories computed from IDM 
to ensure that the PUNN can be trained to capture the noise. The time step $\Delta t$ 
is set to be 0.1 second. The total numbers of data in the accelerating, decelerating, cruising and emergency braking regimes are 468k, 145k, 200k, and 26k , respectively.

The training-validation-test split is also separately applied for each regime. Randomly selected 50\% of the observed data is for training, 25\% for validation, and the rest 25\% for testing. 
To show the PIDL-IDM model's performance for different sizes of training data, in both the prediction-only and joint-estimation problems, we vary the number of observed data  ($n_O$) that is used in the training process.

\paragraph*{Generation of collocation data }

 After the training data is generated, we can follow the data processing steps described in Fig.~\ref{fig:data_stuff} to generate the collocation data. 
 We fix the number of collocation data, $n_C$, in each experiment to be 20, within the range of the varying training data size. 
 Then we randomly select $n_C$ states from the state space $\cal S$, and compute their acceleration values using IDM. 
 The collocation states along with their accelerations constitute the collocation data.
\paragraph*{Sensitivity analysis}

For the prediction-only problem,  
 we perform sensitivity analysis by varying 
 the weights of the loss function ($\alpha$). Among these weights, the one that leads to the minimum test error is called the optimal weight of the loss function and is denoted as $\alpha^*$. Note that when $\alpha$ equals to 1, only the data discrepancy term exists in the loss function according to Eq.~\ref{eqn:loss}, and the PIDL-IDM model is reduced to a pure PUNN model. 

In the joint-estimation problem, $n_O$ is also varying to investigate the effect of the number of training data. The weight of loss function $\alpha$ is fixed to be 0.7. We use a fixed $\alpha$ because our purpose in the joint-estimation problem is to demonstrate whether the PIDL-IDM can do prediction and parameters discovery simultaneously, and the effect of $\alpha$, as is already investigated in the prediction-only problem, is trivial.

\paragraph*{Parameters tuning}
 \tcck{
 The parameters that need to be tuned include the number of collocation data used in training ($n_C$), learning rate for the PUNN ($lr_{\text{PUNN}}$) and the physics-based CFM ($lr_{phy}$), and the PUNN hidden layers size (including the number of hidden layers and the number of neurons in each layer), summarized in Table~\ref{table:PIDL_config_numerical}.
 }
 
 \tcck{Possible values for these parameters are as follows:
 \begin{itemize}
     \item {$n_c$: $\{20,60,100,140,180\}$}
     \item {$lr_{\text{PUNN}}$}: \{$10^{-1}, 10^{-2}, 10^{-3}, 10^{-4},10^{-5}$\}
     \item {$lr_{phy}$}: \{$10^{-1}, 10^{-2}, 10^{-3}, 10^{-4},10^{-5}$\}
     \item {number of hidden layers: } $\{1,2,3,4,5\}$
     \item {number of neurons in each layer: }$\{30,60,128,256,512\}$
 \end{itemize}}

 \tcck{By conducting experiments with different parameter values and comparing the corresponding performance, we select values of these parameters as in Table~\ref{table:PIDL_config_numerical}}. \tcck{Details of the selection of the neural network structure are introduced in Appendix. B}.




\paragraph*{Calibration of the Physcis}

In the prediction-only problem, the ground-truth IDM parameters are known and thus used without calibration. 
In the joint-estimation problem, parameters of the physics-based CFM are jointly trained with the PUNN. 

\subsubsection{PIDL-IDM for the prediction-only problem}

Results of the PIDL-IDM for the prediction-only problem will be presented in this subsection. 

\begin{figure}[h]
    \centering
    \includegraphics[width=0.45\columnwidth]{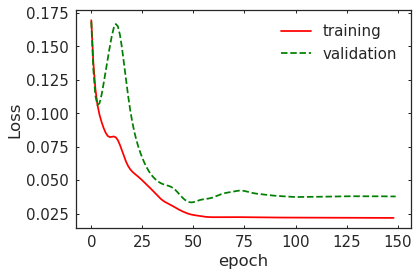}
    \caption{Training and validation error curves for the accelerating regime when $(n_O,n_C,\alpha) = (20, 20, 0.1)$.}
    \label{fig:error_curve}
\end{figure}

To illustrate the early-stop criteria in training, Fig.~\ref{fig:error_curve} presents how the training and validation errors evolve during the training process for the accelerating regime where $(n_O,n_C,\alpha) = (20,20,0.1)$. 
The solid red line and the dashed green line
represent the training error and the validation error, respectively. As the training process proceeds, the training error continuously decreases, while the validation error first decreases and then increases in general. In this specific case, the PIDL-IDM model reaches its minimum validation loss at the 53rd epoch, 
and the trained PUNN parameters at this epoch are used as the optimal parameters $\theta^*$.




\begin{figure}[H]
   
  \captionsetup[subfigure]{oneside,margin={0.5cm,0cm}}
   \centering
   \subfloat[$MSE_{\text{test}}$ of the accelerating regime]{\includegraphics[width=.4\textwidth]{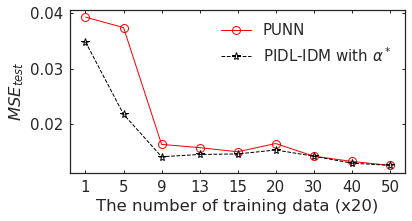}}\quad
   \subfloat[][$\alpha^*$ of the accelerating regime]{\includegraphics[width=.4\textwidth]{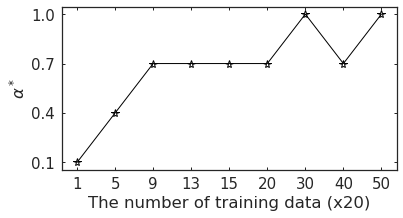}}\quad \\
    \subfloat[][$MSE_{\text{test}}$ of the decelerating regime]{\includegraphics[width=.4\textwidth]{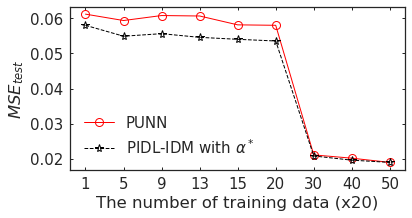}} \quad
    \subfloat[][$\alpha^*$ of the decelerating regime]{\includegraphics[width=.4\textwidth]{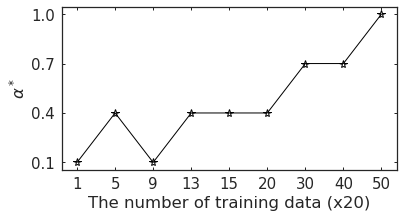}} \quad \\
   \subfloat[][$MSE_{\text{test}}$ of the cruising regime]{\includegraphics[width=.4\textwidth]{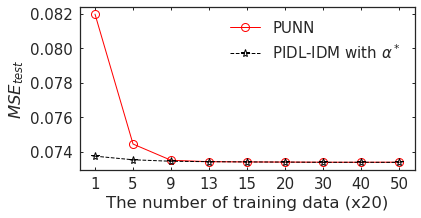}} \quad
   \subfloat[][$\alpha^*$ of the cruising regime]{\includegraphics[width=.4\textwidth]{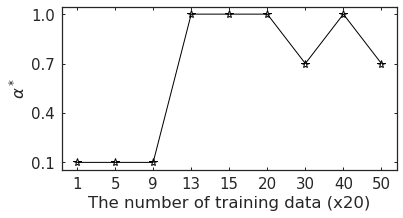}} \quad \\

    \subfloat[][$MSE_{\text{test}}$ of the emergency braking regime]{\includegraphics[width=.4\textwidth]{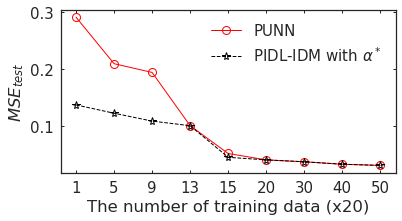}} \quad  
    \subfloat[][$\alpha^*$ of the emergency braking regime]{\includegraphics[width=.4\textwidth]{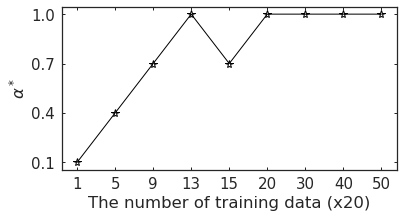}} \quad \\
   \caption{Performance of PIDL-IDM in 4 traffic regimes}
   \label{fig:results:numerical}
\end{figure} 

The prediction accuracy of the PIDL-IDM compared with the PUNN are presented in Fig.~\ref{fig:results:numerical} for 4 regimes, with the physics-uniformed PUNN as the baseline. Each row represents different regimes, the first row for the accelerating regime, the second for the decelerating regime, the third for the cruising regime, and the last for the emergency braking regime. Figures on the first column indicate the test errors of both PIDL-IDM and PUNN with varying training data sizes. The x-axis is the number of training data, and the y-axis is the test error. The red solid line and the dashed black line indicate the test error of the PUNN and the PIDL-IDM, respectively. 
The figures on the second column record the optimal weight of the loss function for each regime, with the x-axis standing for the number of training data and the y-axis for the optimal weight of the loss function. 

From the first-column test error figures, we observe that the PIDL-IDM outperforms the PUNN in all regimes, especially when the training data size is smaller. 
That being said, a neural network encoded with physics improves prediction compared to one without. 
The performance, however, differs in 4 regimes. 
To quantify such a difference, we calculate the average percentage reduction in MSE of the PIDL-IDM relative to the PUNN in each regime. 
PIDL-IDM outperforms PUNN the most in the emergency braking regime by $16.80\%$ and the least in the cruising regime by $1.26\%$. 
For the emergency braking regime, the sudden deceleration may be hard for the PUNN to learn when data is insufficient, while the PIDL-IDM is capable of capturing such changes from the encoded IDM imposed with the lower bound of deceleration. 
For the cruising regime, all the input features lead to the same acceleration rate, which is zero. 
Thus the PUNN can learn such a pattern even without being informed by the physics. 

From the second-column figures, which show the optimal weights of the loss function,
we observe that the value of $\alpha^*$ is low when the training data is small, and converges to one as the number of training data increases. A smaller $\alpha^*$ means a higher weight is given to the physics-discrepancy term in the loss function defined in Eq.~\ref{eqn:loss}. 
Thus the knowledge of the underlying car-following physics compensates for the data insufficiency. As the training data size increases, $\alpha^*$ increases, meaning higher weights are given to the data-discrepancy term in the loss function. This is because PUNN can better capture the underlying data-generating mechanism when sufficient training data is provided. 

\subsubsection{PIDL-IDM for the joint estimation problem}

In the joint-estimation problem, PIDL-IDM are trained on the combined regimes. We fix the weight of the loss function $\alpha$ to be 0.7 and the number of collocation data $n_C$ to be 180, and vary different numbers of observation data $n_O$. The results of the joint-estimation problem are shown in Table~\ref{table:results:numerical_joint_idm}, in which the first column is the number of observed data, the second column is the corresponding test error, and the remaining columns are the values of the estimated parameters along with their deviations from the ground-truth parameters. 

From the table we observe that the PIDL-IDM models with a training data size of 180 or more can achieve a satisfactory performance on both acceleration prediction and parameters estimation. Generally, more data helps increase the accuracy of both prediction and parameters estimation, while the improvement becomes less significant when the number of observed data is more than 260. 
The results demonstrate that the PIDL-IDM model is capable of jointly predicting accelerations and estimating IDM parameters. 

\begin{table}
\begin{center}

\caption{\label{table:results:numerical_joint_idm}Performance of the PIDL-IDM model in the joint-estimation problem}
\begin{tabular}{ c c c c c c c  }
\toprule
\multirow{2}{*}{$n_O (\times 20)$} & \multirow{2}{*}{$MSE_{\text{test}}$} & \multicolumn{5}{c}{estimated parameters (error\%)} \\
\cmidrule{3-7}
 & & $v_0(m \cdot s^{-1})$& $T_0(s)$ &$s_0(m)$&$a_{max}(m \cdot s^{-2})$&$b(m \cdot s^{-2})$ \\
\midrule
1 & 0.126 & 32.309 (7.70) & 1.706 (13.76) & 0.619 (69.04) & 0.872 (19.48) & 1.146 (31.39)\\
5 & 0.127 & 29.372 (2.09) & 1.578 (5.18) & 1.692 (15.42) & 0.813 (11.40) & 1.474 (11.75)\\
9 & 0.061 & 31.604 (5.35) & 1.589 (5.95) & 1.775 (11.26) & 0.784 (7.46) & 1.738 (4.06)\\
13 & 0.039 & 31.706 (5.69) & 1.585 (5.63) & 1.761 (11.97) & 0.737 (0.91) & 1.588 (4.91)\\
15 & 0.038 & 31.486 (4.95) & 1.571 (4.95) & 1.825 (8.73) & 0.746 (2.15) & 1.588 (4.90)\\
20 & 0.037 & 30.799 (2.66) & 1.585 (2.66) & 1.821 (8.93) & 0.737 (0.98) & 1.736 (5.58)\\
\bottomrule
\end{tabular}
\end{center}
\end{table}

\subsection{OVM-based physics}

Other than IDM, OVM is another widely used CFM. 
In this subsection, we will develop PIDL-OVM models. 
Below we will introduce the architecture of the PIDL-OVM model, its experiment setting, and the results of the prediction-only and the joint-estimation problems.

\subsubsection{OVM model specification}
OVM model \citep{lazar2016review} 
is formulated as:  
\begin{equation}\label{eqn:ovm}\left\{\begin{array}{l}
    a(t+\Delta t) = k\left[V\left(h\left(t\right)\right) - v(t)\right] \\
    V(h(t)) = 0.5v_{max}[\text{tanh}(h(t) - h_c)+\text{tanh}(h_c)],
\end{array}\right.\end{equation}
where,\\
$v_{max}$: maximum velocity in the free-flow traffic; \\
$h_c$: safe distance;\\ 
$k$: sensitivity constant; \\
$V(\cdot)$: the Optimal Velocity (OV) function.

The values of model parameters that are used to generate ground-truth trajectories are summarized in the row ``\tcck{ground-truth parameters of the physics}'' of Table~\ref{table:PIDL_config_numerical}.

\subsubsection{PIDL-OVM architecture}
The PIDL-OVM model consists of a PUNN and a physics-based computational graph encoded from an OVM model. Its PUNN shares the same structure as the PUNN component of the PIDL-IDM model. \tcck{Its physics is encoded} from an OVM model, whose architecture is illustrated in Fig.~\ref{fig:PIDL_IDM_OVM}(b). Each circle represents a variable or an intermediate variable.  Each rectangle represents a model parameter, including $v_{max}$, $hc$, and $k$. Each diamond indicates a known operator, where $tanh(\cdot)$ is the hyperbolic tangent function, and other operators have been introduced in the physics architecture of the PIDL-IDM model before.
\subsubsection{Experiment settings}
Numerical experiments are designed to justify the performance of the PIDL-OVM models in the
prediction-only problem and the joint-estimation problem. Experiment settings of these models are mostly the same as in the PIDL-IDM experiment. The major difference exists in the generation of training data, where we use a known OVM to generate the numerical data with all regimes combined. Besides, a minor modification is made on the number of the collocation data $n_C$ in the prediction-only problem, which is 100. Other settings, like generation of collocation data, calibration of the physics, and sensitivity analysis are the same and also summarized in Table~\ref{table:PIDL_config_numerical}.

\paragraph*{Generation of training data}
Vehicle trajectories are generated including all regimes for the simulated period $\forall t\in [0,T]$, where $T$ is the maximum simulation horizon and is set to be 20 seconds. 

Similar to the data generation of the PIDL-IDM experiment, we vary the subject vehicle's initial position and velocity together with its leader's initial velocity, 
Gaussian noise is added to the ground-truth trajectories. The time step $\Delta t$ is 0.1 second, and the total number of training data is 84k.

\subsubsection{PIDL-OVM for the prediction-only problem}
\begin{figure}[H]
   
  \captionsetup[subfigure]{oneside,margin={0.5cm,0cm}}
   \centering
   \subfloat[]{\includegraphics[width=.4\textwidth]{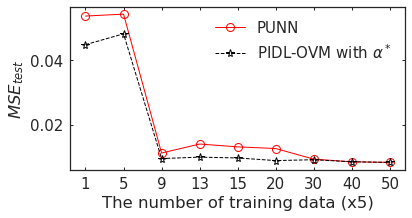}}\quad
   \subfloat[][]{\includegraphics[width=.4\textwidth]{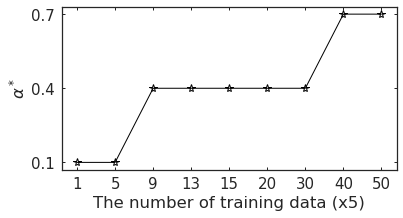}}\quad \\
   \caption{Performance of PIDL-OVM in combined traffic regimes}
   \label{fig:results:numerical_OVM}
\end{figure} 

The prediction accuracy of the PIDL-IDM compared with the PUNN are presented in Fig.~\ref{fig:results:numerical_OVM} for all regimes combined together, with the physics-uninformed PUNN as the baseline. 
From Fig.~\ref{fig:results:numerical_OVM}(a), we observe that the PIDL-OVM outperforms the PUNN, especially when data is insufficient. 
Likewise, a neural network encoded with physics improves prediction compared to one without.
This trend is the same as in the PIDL-IDM experiment. In Fig.~\ref{fig:results:numerical_OVM}(b), $\alpha^*$ is attained at a smaller value when data is insufficient. Its value gradually increases to one as the number of training data increases. This trend can be explained by the same reason as in the PIDL-IDM experiment: a small $\alpha^*$ means more weights are put on the physics-discrepancy term to compensate the data insufficiency, and a big $\alpha^*$ means more weights are put on the data-discrepancy term as the PUNN can better capture the underlying data-generating pattern with sufficient training data. 
The similar performance of PIDL-OVM also demonstrates that the PIDL model preserves its advantage when its physics component is changed from an IDM model to an OVM model.

\subsubsection{PIDL-OVM for the joint-estimation problem}
Results of the joint-estimation problem of PIDL-OVM are shown in Table~\ref{table:results:numerical_joint}, in which the estimated parameters are compared to the ground-truth parameters. 
From the table we observe that PIDL-OVM model with a training data size of 180 or more can achieve a satisfactory performance on both acceleration prediction and parameters estimation. These results demonstrate that the PIDL-OVM mode is capable of jointly predicting accelerations and estimating OVM parameters.

\begin{table}
\begin{center}

\caption{\label{table:results:numerical_joint}Performance of the PIDL-OVM model in the joint-estimation problem}
\begin{tabular}{ c c c c c  }
\toprule
\multirow{2}{*}{$n_O (\times 20)$} & \multirow{2}{*}{$MSE_{\text{test}}$} & \multicolumn{3}{c}{estimated parameters (error\%)} \\
\cmidrule{3-5}
 & & $v_{max}(m \cdot s^{-2})$& $h_c(m)$ &$k(s^{-1})$\\
\midrule
1 & 0.052 & 34.646 (15.49) & 10.300 (3.00) & 0.022 (25.25) \\
5 & 0.051 & 31.784 (5.95) &  10.217 (2.17) &  0.026 (12.86)  \\
9 & 0.015 & 30.572 (1.91) & 10.128 (1.28) &  0.029 (4.45)  \\
13 & 0.016 & 30.240 (0.80) & 10.065 (0.65) &  0.030 (0.23) \\
15 & 0.015 & 30.571 (1.90) & 10.083 (0.83) &  0.029 (2.47)  \\
20 & 0.013 & 30.565 (1.88) & 10.076 (0.76) &  0.029 (2.30)  \\
\bottomrule
\end{tabular}
\end{center}
\end{table}


\section{PIDL for the NGSIM Dataset} \label{sec:PIDL_for_ngsim}
\label{sec:real_exp}
In this section, we will validate the performance of PIDL-CF models on the real-world data, the Next Generation SIMulation (NGSIM) dataset. 
\tcck{As aforementioned at the beginning of Sec.~\ref{sec:numerical}, 
the intrinsic mechanism that generates NGSIM trajectories is unknown. Thus, we need to try different physics-based models. 
The performances of PIDL-CF models can only be judged through their predicted acceleration values, not calibrated physics. 
}Two PIDL-CF models (the PIDL-IDM model and PIDL-OVM model) are trained to solve the prediction-only and joint-estimation problems. 
The architectures of these two PIDL-CF models, which consists of a PUNN and a physics component, are the same as in the numerical experiment in Fig.~\ref{fig:PIDL_IDM_OVM}. 
Experiment settings of these two PIDL-CF models are summarized in Table~\ref{table:PIDL_config_ngsiml}.
The rows are the same as those in Table~\ref{table:PIDL_config_numerical}. 
The columns are organized in three hierarchies. 
The top level classifies two PIDL models based on physics models: PIDL-IDM and PIDL-OVM models. 
The middle level indicates that, for each model, we try both prediction-only and joint-estimation problems. 
The bottom level indicates that two calibration and evaluation methods are used. 
The parameter tuning process is the same as in Sec.~\ref{sec:pidl-idm-exp-set}. 





\begin{table}[h!]
\begin{center}

\caption{\label{table:PIDL_config_ngsiml}Experiment settings of the PIDL-IDM model and the PIDL-OVM model for the NGSIM data}
\begin{tabular}{ |c|c||c|c|c|c|c|c|  } 
\hline

\multirow{3}{*}{\textbf{Name}} & \multirow{3}{*}{\textbf{Notation}}  & \multicolumn{3}{c|}{\textbf{PIDL-IDM}} & \multicolumn{3}{c|}{\textbf{PIDL-OVM}} \\
\cline{3-8}
& & \multicolumn{2}{c|}{\makecell{ prediction \\ -only}} & \makecell{ joint \\ -estimation} & \multicolumn{2}{c|}{\makecell{ prediction \\ -only}} & \makecell{ joint \\ -estimation}\\
\cline{3-8}
& & \makecell{ one \\ -step} & \makecell{long \\ -term} & \makecell{one \\ -step} & \makecell{ one \\ -step} & \makecell{long \\ -term} & \makecell{one \\ -step} \\
\hline\hline

\makecell{number of observed data\\ used in training ($\times 5$)} & $n_O$ &  \multicolumn{6}{c|}{1, 2, 3, 4, 5, 10, 15, 20}   \\
\hline


weight of the loss function &$\alpha$ &\multicolumn{6}{c|}{0.1, 0.4, 0.7, 1.0} \\
\hline
\makecell{number of collocation data \\ used in training ($\times 5$)} & $n_C$ &\multicolumn{6}{c|}{10}  \\
\hline
\makecell{training-validation-test \\ split ratio} & -  & \multicolumn{6}{c|}{0.5:0.25:0.25}\\
\hline

\multirow{2}{*}{learning rate}  & $lr_{\text{PUNN}}$  &  \multicolumn{6}{c|}{0.001} \\
\cline{2-8}
                                 & $lr_{phy}$  & \multicolumn{2}{c|}{-} & 0.1 & \multicolumn{2}{c|}{-} & 0.1\\
\hline
PUNN hidden layers size &-&  \multicolumn{6}{c|}{(128, 256, 512, 256, 128)}\\
\hline

traffic regimes & - & \multicolumn{6}{c|}{\makecell{all regimes combined}}\\

\hline
\makecell{pre-calibration \\ method} & - & \makecell{LS } & GA & - & \makecell{LS} & GA & -   \\

\hline

\end{tabular}

\end{center}
\end{table}

\subsection{Experiment settings}

Below we will detail the extraction and pre-processing of the training data, PIDL models, baselines and calibration of physics-based models, and evaluation methods. 

\paragraph*{Extraction and pre-processing of training data}
The NGSIM data was collected from a segment of US Highway 101 by a camera placed on the top of a high building on June 15, 2005. This highway consists of five lanes in the mainline together with an auxiliary lane between an on-ramp and an off-ramp. High-resolution vehicular information is provided, including the position, velocity, acceleration, occupied lane, spacing and the vehicle class (automobile, truck and motorcycle) at every 0.1 second. We focus on trajectories of automobiles from all five lanes in the mainline. 

\tcck{
We adopt the same method of calculating the speed and acceleration as used in \cite{coifman2017critical}. 
The calculation of speed is as depicted in Eq.~\ref{equ:filter_v}. This process can be viewed as a non-linear low pass filter, which smooths out the high-frequency noise. 
A Savitzky-Golay filter \citep{coifman2017critical} is then applied on the calculated velocity to further reduce the remaining noise.}

\begin{equation}\label{equ:filter_v}
\begin{split}
    v(t) &= \underset{i \in \{1,2,...,7\}}{med}\frac{x(t+i\Delta T) - x(t-i\Delta T)}{2i\Delta T}
\end{split}
\end{equation}
where,\\
$med$: sample median,\\
$x(t+i\Delta T)$: position at time $t+i\Delta T$.\\
$x(t-i\Delta T)$: position at time $t-i\Delta T$.\\
\tcck{$\Delta T$: sampling interval, which is 0.1 second in the NGSIM dataset.}

\tcck{
With the processed speed, the calculation of acceleration is done via Eq.~\ref{equ:filter_a}. A Savitzky-Golay filter is further applied on the calculated acceleration to further smoothen out the remaining noise.}

\begin{equation}\label{equ:filter_a}
\begin{split}
    a(t) &= \frac{x(t+\Delta T) - x(t-\Delta T)}{2\Delta T}
\end{split}
\end{equation}

\tcck{
A CF case consists of trajectory segments of a subject vehicle and its leader, which is a time-series data of their positions. As we are more interested in the follower's state, i.e. a vector consisting of spacing, speed difference and speed, we represent a CF cases as $\{[h(t),\Delta v(t), v(t)]|t=1,...,T_i\}$, where $T_i$ is the duration of the $i$th CF case. A CF case is identified when the following criteria are satisfied:
\begin{enumerate}
    \item The subject vehicle and its leader are in the same lane;
    \item The spacing does not exceed 150 meters;
    \item The duration is longer than 10 seconds.
\end{enumerate}
}

\tcck{
To further reduce heterogeneity in data, we select CF cases with similar driving behaviors. The process of which is as follows:
}
\begin{enumerate}
    \item{ \tcck{Design a trajectory-level feature for each CF case. We characterize each CF case by 5 representative parameters estimated from IDM and 3 from OVM. First, we calibrate an IDM model and an OVM model for each CF case. The parameter sets of the $i$th CF case for each model are denoted as $\lambda^{IDM}_i$ and $\lambda^{OVM}_i$, respectively. For example, $\lambda^{IDM}_i=\{30, 1.5, 2, 0.73, 1.63\}$ and $\lambda^{OVM}_i=\{30, 10, 0.03\}$. Then we combine the parameters calibrated from two models, and use the eight calibrated parameters $\lambda_i = \lambda^{IDM}_i \cup \lambda^{OVM}_i  = \{30, 1.5, 2, 0.73, 1.63, 30, 10, 0.03\} $ as the feature to represent a CF case.}
    }
    \item{\tcck{Set similarity criteria. First features calculated in the previous step are normalized to range [0,1]. 
    The Euclidean distance is computed to measure the similarity between two CF cases in terms of features. Within the feature space, a centroid is defined as the feature that has the smallest total distance to all other features. Once the centroid is located, 50 CF cases are selected of which features have the smallest distances to the centroid. The total number of data points is 45K.}
    }
    
\end{enumerate}

\begin{note}
We add a one-second reaction time when pairing states with actions before training IDM and OVM models.
\end{note}

\paragraph*{PIDL models}

To test the performance of PIDL-CF on NGSIM, we try both physics-based models: IDM and OVM. 
Accordingly, we will train PIDL-IDM and PIDL-OVM models. 
For each model, we will also try both prediction-only and joint-estimation problems. 
The PUNNs in the PIDL-IDM and PIDL-OVM models share the same architecture. 

\paragraph*{Baselines and calibration of IDM/OVM}
Two types of baseline models are used to compare with the PIDL models for accelerations prediction accuracy:

\begin{enumerate}
    \item {A PUNN model. }
    \item {The physics-based models, which include an IDM model and an OVM  model. Two types of calibration methods are used for the IDM/OVM.}
    \begin{enumerate}
        \item {A one-step calibration method, which is the LS estimation.  In the LS estimation, physics parameters are optimized to minimize the error of the instant prediction of the current acceleration, thus it calibrates physics-based models from a ``one-step'' perspective.}
        \item {A \tcck{trajectorial} calibration method, which is GA. In GA, the accumulative error of predicting the whole trajectory can be considered and thus it calibrates physics-based models from a ``\tcck{trajectorial}'' perspective.}
    \end{enumerate}
   
\end{enumerate}

\paragraph*{Evaluation methods and metrics}
In accordance with the one-step and trajectorial calibration methods, two evaluation methods are proposed to compute the test errors of the trained PIDL-CF models: 
\begin{enumerate}[(a)]
    \item A one-step evaluation method, which only considers the instant prediction of the acceleration. For this evaluation method, we continue to use the test MSE to measure the discrepancy between the observed and predicted accelerations on the test data, as is formulated in Eq.~\ref{eqn:test_mse}. 
    \item A \tcck{trajectorial} evaluation method, which considers the accumulative error during the temporal evolution of states and aims to reproduce the entire trajectory with only the initial state. For this evaluation method, we are interested in whether the model can reproduce the whole trajectories, or to be more specific, the position and velocity profiles as shown in Fig.~\ref{fig:data_stuff}. Thus prediction errors are calculated on positions and velocities instead of accelerations. In this paper we use a Root Mean Square Percentage Error (RMSPE) to quantify the position error and the velocity error as follows: 

\begin{equation}
 \begin{split}
     RMSPE_x &= \sqrt{
     \frac{
            \sum_{i=1}^{N_T} \sum_{t=0}^{T_i}
            \left| 
                \frac{x(t)-\hat{x}(t)}{\hat{x}(t)}
            \right|^2 
        }
        {
            \sum_{i=1}^{N_T} \sum_{t=0}^{T_i} 1
        }
     },\\
     RMSPE_v &= \sqrt{
     \frac{
            \sum_{i=1}^{N_T} \sum_{t=0}^{T_i}
            \left| 
                \frac{v(t)-\hat{v}(t)}{\hat{v}(t)}
            \right|^2 
        }
        {
            \sum_{i=1}^{N_T} \sum_{t=0}^{T_i} 1
        }
     },
 \end{split}
 \end{equation}
where, \\
$x(t)$: the predicted position of the subject vehicle;\\
$v(t)$: the predicted velocity of the subject vehicle;\\
$\hat{x}(t)$: the observed position of the subject vehicle;\\
$\hat{v}(t)$: the observed velocity of the subject vehicle;\\
$T_i$: the time horizon of the $i$th CF case;\\
$N_T$: the total number of CF cases.

\end{enumerate}

\subsection{Performance comparisons}\label{sec:performance_compare}

To demonstrate the performance of PIDL-CF models on NGSIM, we will first present the models based on one-step comparison, including the prediction-only and joint-estimation problems of both PIDL-IDM and PIDL-OVM. 
We then present the models based on trajectorial comparison, including the prediction-only problem of both PIDL-IDM and PIDL-OVM. 

\subsubsection{One-step comparison}
\label{sec:one_step_comp}

\begin{figure}[H]
   \centering
   \subfloat[][]{\includegraphics[height=0.46\textwidth,]{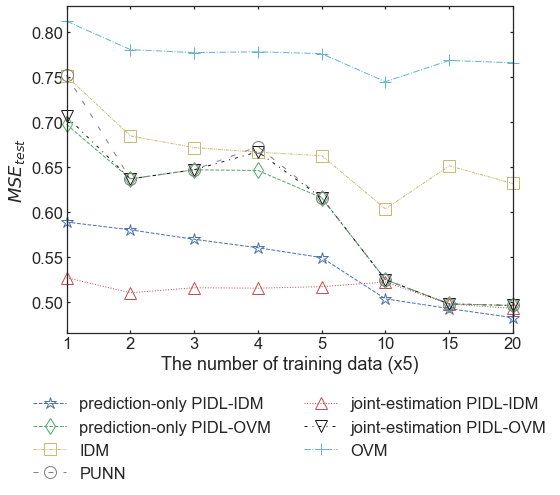}}  
   \subfloat[][]{\includegraphics[height=0.46\textwidth]{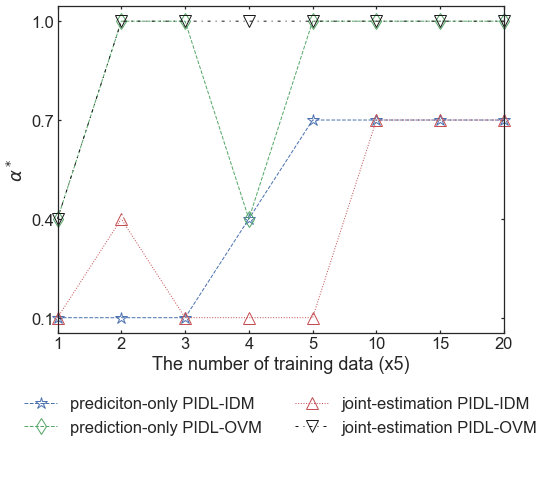}} 
   \caption{Prediction-only and joint-estimation for PIDL-IDM and PIDL-OVM using the one-step calibration and evaluation on the NGSIM (US-101) data.}
   \label{fig:results:one_step_all}
\end{figure} 

Fig.~\ref{fig:results:one_step_all} presents the prediction-only and joint-estimation problems of PIDL-IDM and PIDL-OVM models using the one-step calibration and evaluation, together with the baselines results. 
Fig.~\ref{fig:results:one_step_all}(a) is the test error of the PIDL-IDM model, PIDL-OVM model, and the baselines. The x-axis is the number of the training data, and the y-axis is the test error. 
The cyan line with crosses is for the OVM model, the yellow line with squares for the IDM model, the grey line with circles for the PUNN, the blue line with stars for the prediction-only PIDL-IDM model, the red line with triangles for the joint-estimation PIDL-IDM model, the green line with diamonds for the prediction-only PIDL-OVM model, and the black line with inverted triangles for the joint-estimation PIDL-OVM model.
Fig.~\ref{fig:results:one_step_all}(b) is the optimal weight $\alpha^*$ of the loss function for both the prediction-only and joint-estimation problems of PIDL-IDM and PIDL-OVM models. 
The x-axis stands for the number of training data and the y-axis for the optimal weight of the loss function. The line specification for each model is the same as in Fig.~\ref{fig:results:one_step_all}(a). 
To compare model performances, we provide three perspectives below:
\begin{enumerate}
    \item{\textbf{Comparison between PIDL models and baselines.} The pure physics-based IDM and OVM models perform the worst. 
    However, 
    by encoding the physics-based model into the PUNN, the resulting PIDL models outperform the PUNN. 
    For the PIDL-IDM model, both the prediction-only and joint-estimation models outperform the PUNN, especially when the training data is insufficient.
    For the PIDL-OVM model, the prediction-only and the joint-estimation models slightly outperform the PUNN for some training data sizes. } 
    
     \item{\textbf{Comparison between the PIDL-IDM model and the PIDL-OVM model.} The PIDL-IDM models outperform the PIDL-OVM models for both the prediction-only and joint-estimation problems. 
     It may likely be caused by their physics components.  
     The IDM model can better predict the acceleration than the OVM model, as indicated by their test error curves in Fig.~\ref{fig:results:one_step_all}(a). 
     In other words, IDM provides more accurate physics information than OVM, resulting in lower prediction errors in PIDL models. 
     From Fig.~\ref{fig:results:one_step_all}(b) we observe that $\alpha^*$ of the PIDL-IDM model is lower than that of the PIDL-OVM model for both the prediction-only and the joint-estimation problems, especially for the joint-estimation problem. This means a higher weight is given to the physics for the PIDL-IDM model, likely because IDM provides more accurate physics information than OVM.   }
    \item{\textbf{Comparison between the prediction-only problem and the joint-estimation problem.}} 
    For the PIDL-OVM model, the test errors of the prediction-only and joint-estimation are close, while for the PIDL-IDM model, the test error of the joint-estimation problem is prominently smaller than that of the prediction problem.
    This is because the joint-estimation problem jointly trains parameters of physics, which could lead to more accurate physics than the prediction model that calibrates physics prior to the training process. 
    This can be evidenced by $\alpha^*$ in Fig.~\ref{fig:results:one_step_all}(b): its value for the joint-estimation problem is smaller than that for prediction-only, indicating that a higher weight is assigned to the physics. In other words, more accurate physics information can be learned in the joint-estimation problem.

\end{enumerate}

The estimated parameters of the PIDL-IDM and PIDL-OVM models in the joint-estimation problem are presented in Table~\ref{table:results:ngsim_joint}. From this table, we observe that the physics parameters of both IDM and OVM begin to converge and fall in a realistic range. This result, together with the prediction errors in Fig.~\ref{fig:results:one_step_all}(a), demonstrates that the PIDL models are capable of jointly predicting accelerations and estimating physics parameters for the real-world data.

\begin{table}
\begin{center}

\caption{\label{table:results:ngsim_joint}Estimated physics parameters of the IDM and OVM models in the joint-estimated problem for the NGSIM data.}
\begin{tabular}{ c c c c c c:c c c  }
\toprule
\multirow{2}{*}{$n_O (\times 5)$} & \multicolumn{5}{c}{IDM} & \multicolumn{3}{c}{OVM}  \\
\cmidrule{2-6} \cmidrule{7-9} 
 & $v_0(m\cdot s^{-1})$ &  $T_0 (s)$ & $s_0 (m)$& $a_{max} (m \cdot s^{-2})$ & $b(m \cdot s^{-2})$ & $v_{max}(m\cdot s^{-1})$ & $h_c(m)$ & $k(s^{-1})$  \\
\midrule
1 & 15.109 & 0.744 & 2.150 & 0.209 & 3.931  & 6.963 & 0.049 & 0.019 \\
5  & 15.025 & 0.606 & 1.647 & 0.290 & 4.162 &  9.636  & 8.015 & 0.014  \\
20 & 16.775 & 0.504 & 1.667 & 0.430  & 3.216 &  10.219 & 6.409  & 0.066  \\
\bottomrule
\end{tabular}
\end{center}
\end{table}

\subsubsection{\tcck{Trajectorial} comparison} 
\begin{figure}[ht]
   
   \centering
   \subfloat[][]{\includegraphics[width=.5\textwidth]{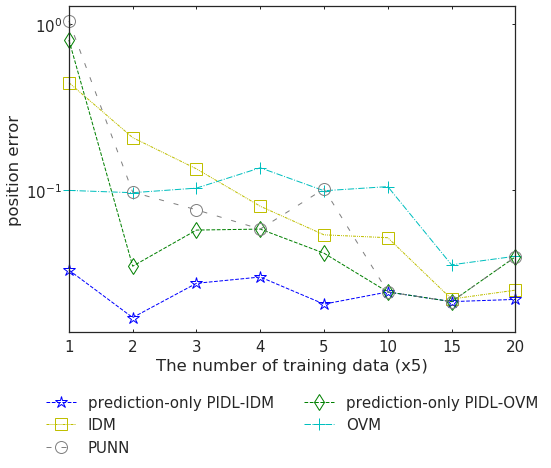}}
   \subfloat[][]{\includegraphics[width=.5\textwidth]{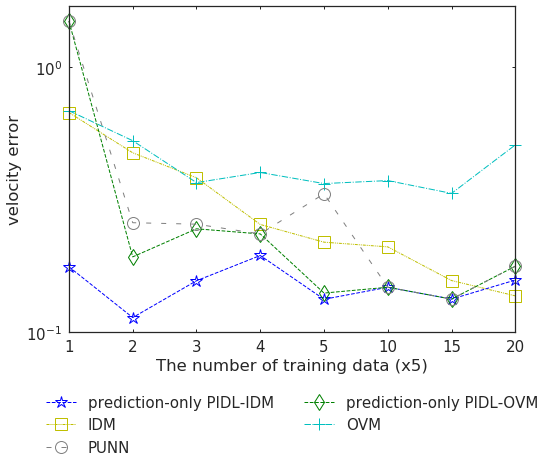}}
   \caption{Results of the prediction-only PIDL models using the trajectorial calibration and evaluation for the NGSIM data.}
    \label{fig:results:long-term-pre-ngsim}
\end{figure}

Fig.~\ref{fig:results:long-term-pre-ngsim} presents the position and velocity errors of the prediction-only PIDL models using the trajectorial calibration and evaluation, together with the baselines results. Fig.~\ref{fig:results:long-term-pre-ngsim}(a) is the position error and Fig.~\ref{fig:results:long-term-pre-ngsim}(b) is the velocity error, with each line indicating the error of each model. The x-axis is the number of training data and the y-axis is the error. From  Fig.~\ref{fig:results:long-term-pre-ngsim}(a) and (b), we can observe that the prediction-only PIDL models outperform the baselines in terms of both the position and velocity errors especially for insufficient training data. 

Comparing Fig.~\ref{fig:results:long-term-pre-ngsim} and Fig.~\ref{fig:results:one_step_all}(a), we observe that by using the trajectorial calibration and evaluation, the advantages of the PIDL models over the PUNN become more significant. 
A sharp contrast is the PIDL-OVM model. 
It shows a much more significant advantage over PUNN using the trajectorial calibration and evaluation than using the one-step calibration and evaluation. 
This is because OVM in Fig.~\ref{fig:results:long-term-pre-ngsim} is calibrated by GA, with which the trajectorial error is considered during the calibration process. 
As a result, OVM provides more accurate physics information to the PIDL models, leading to a better performance on the real-world data. 
This result demonstrates the superior performance of our PIDL models using the trajectorial calibration and evaluation over the one-step approach.

\begin{figure}[H]
   
  \captionsetup[subfigure]{oneside,margin={0.5cm,0cm}}
   \centering
   \subfloat[]{\includegraphics[width=.4\textwidth]{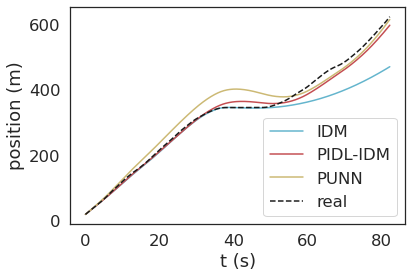}}\quad
   \subfloat[]{\includegraphics[width=.4\textwidth]{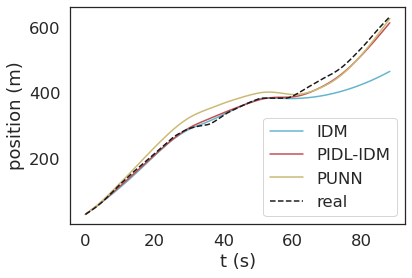}}\quad \\
    \subfloat[]{\includegraphics[width=.4\textwidth]{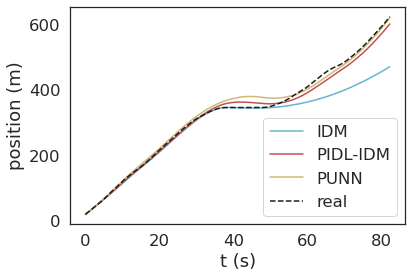}} \quad
    \subfloat[]{\includegraphics[width=.4\textwidth]{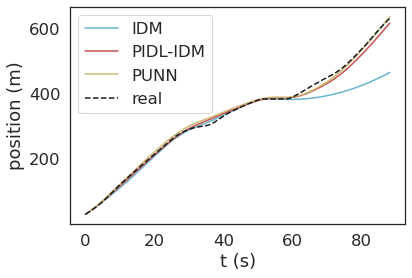}} \quad 
   \caption{Comparison of predicted and real positions. (a) and (b): $n_O$ = 10, $\alpha = 0.1$; (c) and (d): $n_O = 50$, $\alpha = 0.4$.}
   \label{fig:results:trajectory}
\end{figure} 

\textcolor{black}{The predicted position trajectories are compared with the ground truth in Fig.~\ref{fig:results:trajectory}. The first row is when $n_O=10$ and $\alpha = 0.1$, and the second row is when $n_O=50$ and $\alpha = 0.4$. Different columns are two selected representative trajectories where a stop-and-go is clearly observed.
 The x-axis is the time and the y-axis is position. The dashed line indicates the real-world trajectory, and the solid lines stand for the predicted ones by different models: the cyan for the IDM, the yellow for the PUNN, and the red for the PIDL. By comparison, we can see that the PIDL can better predict the vehicle position than the PUNN, especially in the stop-and-go regime}. 


\subsection{Sensitivity analysis for PIDL components}
\textcolor{black}{In this subsection, we will investigate the PIDL methods with different types of components. IDM and OVM were tested as the physical component in previous sections, and we will test another two widely used physics-based car-following models, the GHR and FVDM models. As for the deep learning component, we will try the LSTM, which is one of the state-of-the-art DL methods in modeling the microscopic driver behavior.}

\subsubsection{Physics-based car-following models}
\begin{figure}[h]
    \centering
    \includegraphics[width=0.6\columnwidth]{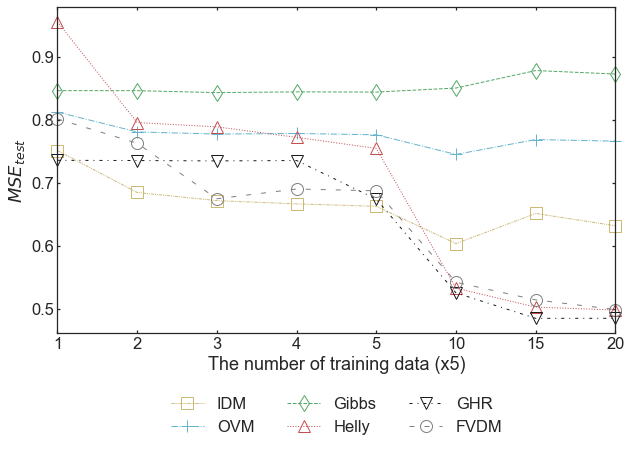}
    \caption{Performance of 6 physical microscopic car-following models.}
    \label{fig:6_cfm}
\end{figure}

\textcolor{black}{Apart from IDM and OVM models, we also consider a range of physics-based CFMs, like the GHR, FVDM, Helly, and Gipps models (defined in Table~\ref{tab:map}). 
To determine which physics can enhance the performance of PIDL-CFM on the NGSIM data, we first compare the test errors of calibrated physics on the same training set. 
Fig.~\ref{fig:6_cfm} shows the test MSE of those models, in which the x-axis is the number of training data and the y-axis is the test MSE. Different lines indicate the error of different physical models.  Different markers are used for different CFMs: squares for the IDM, diamonds for the Gipps, inverted triangles for the GHR, crosses for the OVM, triangles for the Helly, and circles for the FVDM. 
From this figure we can see that the IDM model performs the best when the number of training data is less than 25, and is worse than the GHR, Helly and FVDM models as training data increases. 
Since we are more interested in the data-insufficient region where the physics-based CFMs play a more important role in the PIDL performance, 
we select GHR and FVDM in addition to IDM as the alternative physics component for the PIDL.
} 

\begin{figure}[h]
   
   \centering
   \subfloat[][Prediction-only]{\includegraphics[height=.25\textwidth]{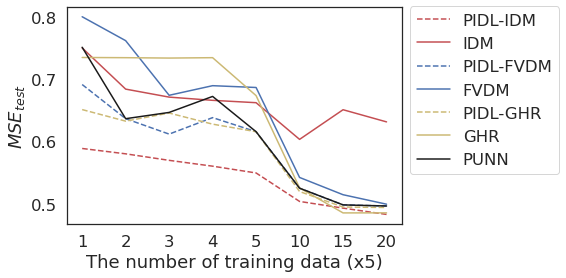}}
   \subfloat[][Joint-estimation]{\includegraphics[height=.25\textwidth]{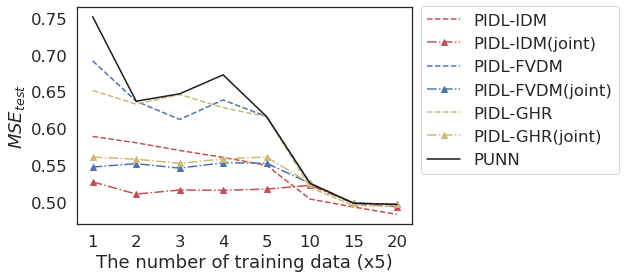}}
   \caption{Performance comparison among PIDL-IDM, PIDL-FVDM and PIDL-GHR on the NGSIM data.}
    \label{fig:results:FVDM_GHR}
\end{figure}

\textcolor{black}{The results of the PIDL-GHR, PIDL-FVDM, and the previously shown PIDL-IDM are presented in Fig.~\ref{fig:results:FVDM_GHR}. 
Fig.~\ref{fig:results:FVDM_GHR} (a) compares the prediction-only results with the baselines, and Fig.~\ref{fig:results:FVDM_GHR} (b) compares the joint-estimation results with the prediction-only ones.
The x-axis is the number of training data and the y-axis is the test MSE. Different line styles are used to indicate different types of results: solid lines for baselines, dashed lines for prediction-only results, and dashed lines with markers for the joint-estimation results. Different colors are used for different physical models: the red for the IDM, the blue for the FVDM, the yellow for the GHR, and the black for non-physical model (ANN). For sake of clarity, results are shown in two subfigures. 
}

\textcolor{black}{In Fig.\ref{fig:results:FVDM_GHR}(a),  
the test MSE of prediction-only PIDL models, including PIDL-IDM (dashed red), PIDL-FVDM (dashed blue), and PIDL-GHR (dashed yellow), are compared to the pure NN-based component (solid black) and their respective physics components (solid red lines for the IDM, solid blue for the FVDM, and solid yellow for the GHR).}
All the PIDL models outperform the ANN, which shows that the PIDL method is superior to its pure physics counterparts and thus agnostic of encoded physics. 
When the data size is smaller than 50, the PIDL-IDM outperforms the other two PIDL models. 
This is because the pure IDM outperforms the other two physics-based models, and can thus provide more accurate physics information to better train the PUNN. This advantage is less significant as the data increases, because the PIDL model weighs more on data than physics when the data size grows, rendering the physics less influential.

\textcolor{black}{In Fig.~\ref{fig:results:FVDM_GHR}(b),  the line specification of prediction-only models is the same as in Fig.~\ref{fig:results:FVDM_GHR}(a), and the dashed lines with markers indicate the results of joint-estimation PIDL models: the red for the PIDL-IDM, the blue for the PIDL-FVDM, and the yellow for the PIDL-GHR.
The joint-estimation PIDL models achieve smaller errors than the prediction-only PIDL models for all three physics-based car-following models, which validates the superiority of the joint-estimation models across different physics-based CFMs. The advantage of the PIDL-IDM model is also observed when the training data is less than 50, which becomes less significant as the training data increases. }

\subsubsection{Neural network architecture}

\begin{figure}[h]
   \centering
   \subfloat[][Structure of PIDL-LSTM]{\includegraphics[width=0.8\textwidth]{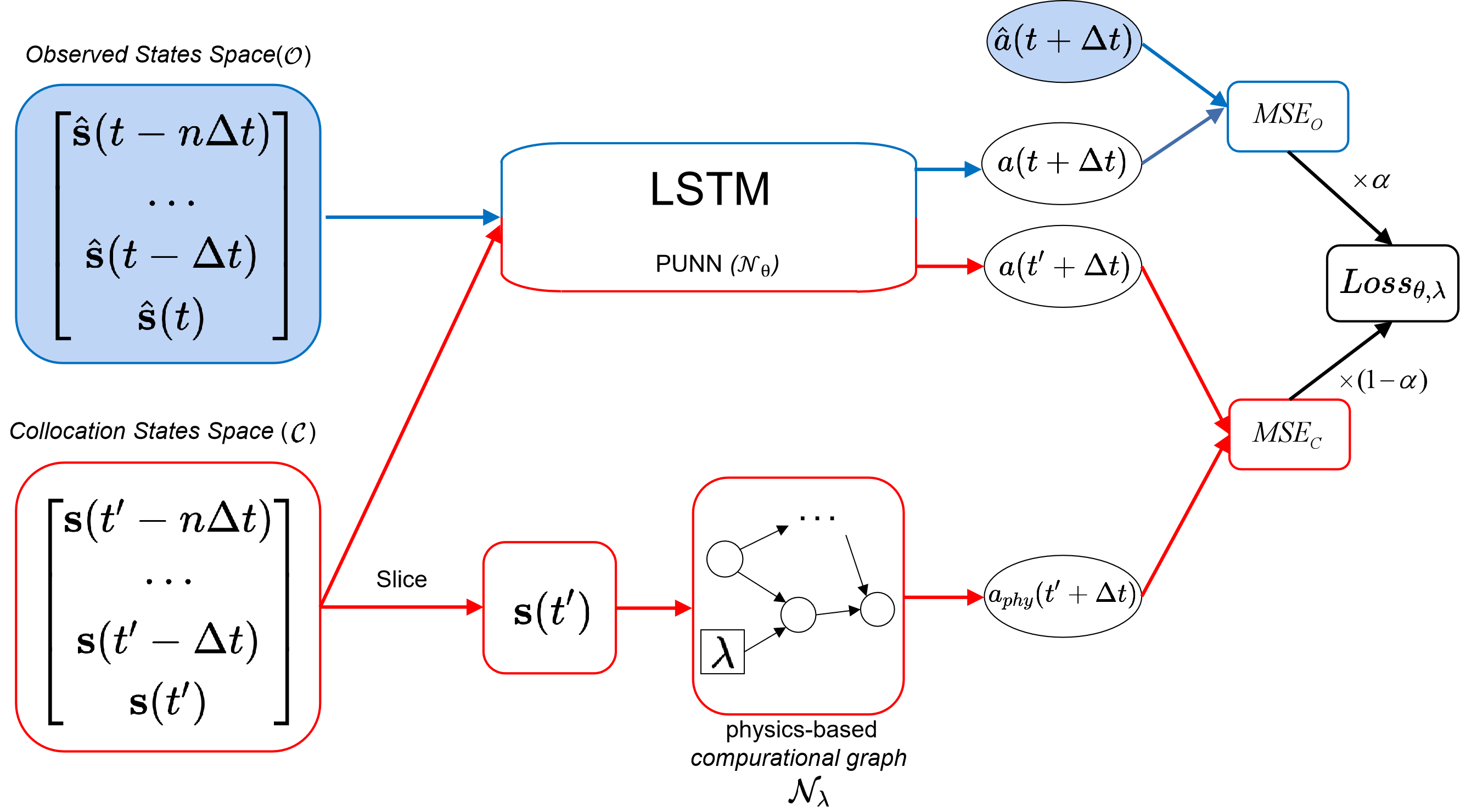}} \\
   \subfloat[][LSTM cells ($h$ and $c$ are hidden states and cell states. $\sigma(\cdot)$ and $tanh(\cdot)$ denote the sigmoid and hyperbolic tangent functions, respectively.)]{\includegraphics[width=\textwidth]{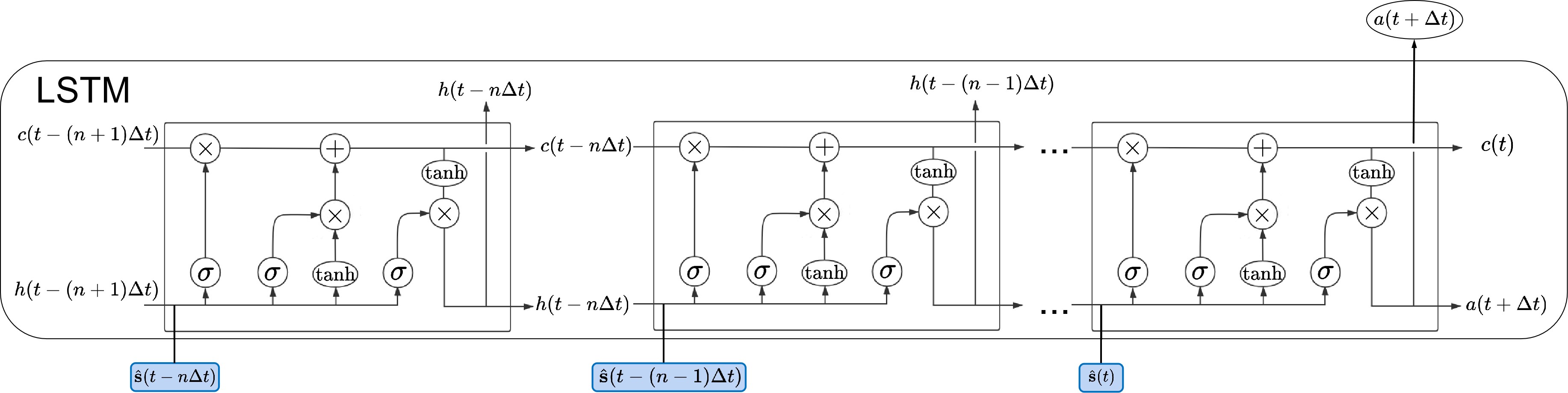}}
   \caption{PIDL-LSTM model}
    \label{fig:results:lstm_structure}
\end{figure}

\tcck{
To validate the functionality of the PIDL method for other deep learning model, we changed the DL component of the PIDL from an ANN to a long short-term memory (LSTM) model. The LSTM model is one of the states-of-the-art DL model used in the car-following behavior modeling. Different from ANN, LSTM considers historic data when making a prediction.  The PIDL-LSTM structure is shown in Fig.~\ref{fig:results:lstm_structure}.  Fig.~\ref{fig:results:lstm_structure} (a) introduces how to incorporate LSTM into the PIDL architecture, and Fig.~\ref{fig:results:lstm_structure} (b) zooms into the LSTM component with the details of LSTM cells. In Fig.~\ref{fig:results:lstm_structure}(a), $\mathbf{s}(t)$ is the feature vector $[h(t),\Delta v(t), v(t)]$, $n$ is the gap length. 
Note that there is no historic component in the physics-based CFM, only the latest feature is used for the physics regardless of the gap length. 
In Fig.\ref{fig:results:lstm_structure}(b), each rectangle encloses an LSTM cell. 
The row element of the input data, which is a feature vector, $\hat{\mathbf{s}}(t)$ as the observed feature and $\mathbf{s}(t^{\prime})$ as the collocation feature, is fed into the LSTM cell sequentially. For each input feature $\mathbf{s}(t)$, the LSTM cell outputs the cell state ($c(t)$) and hidden state ($h(t)$), which are also fed into the LSTM cell for the next time step. The hyperbolic tangent function, denoted as $tanh(\cdot)$, is a non-linear transform function, which adds non-linearity to the LSTM model. The sigmoid function, denoted as $\sigma(\cdot)$, outputs a number between zero and one, which controls how much information can pass through. One means letting everything through and zero means letting nothing through.
}

\begin{figure}[h]
   \centering
   \subfloat[][]{\includegraphics[width=0.48\textwidth]{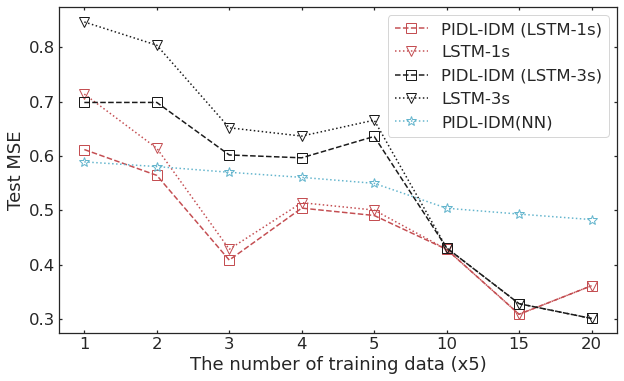}} \quad
   \subfloat[][]{\includegraphics[width=0.48\textwidth]{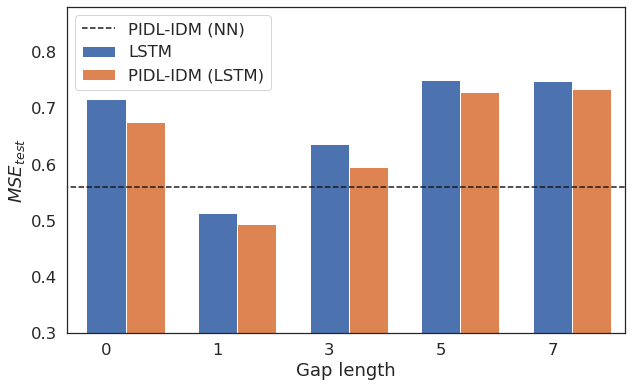}}
   \caption{Results of PIDL-IDM (LSTM) with varying numbers of training data (a) and varying gap lengths (b).}
    \label{fig:results:lstm_result}
\end{figure}

\tcck{
We apply the LSTM-based PIDL-IDM to NGSIM and present results in Fig.~\ref{fig:results:lstm_result}. 
Fig.~\ref{fig:results:lstm_result}(a) shows the test MSE on two LSTM-based PIDL-IDM models, one with a 1-second gap length (indicated by dashed red line with squares) and the other with a 3-second gap length (indicated by dashed black line with squares), against the training data size. 
Baseline models include pure data-driven LSTM with gap lengths to be 1 second (dashed red line with inverted triangles) and 3 seconds (dashed black line with inverted triangles), together with an ANN based PIDL-IDM model (dashed blue line with stars).
The x-axis the training data size and the y-axis is the test MSE. When the length gap is 3 seconds, both the LSTM and the LSTM-based PIDL model are worse than the ANN-based PIDL model. 
The LSTM-based PIDL model achieves the best MSE when the gap length is 1 second, but it can only outperform the ANN-based PIDL model when the training data size is more than 10. 
The LSTM-based PIDL model shows its advantage when the training data size is more than 25. It is because the LSTM's complex structure requires more training data but at the same time is more capable of learning driver reaction.
Figure.~\ref{fig:results:lstm_result} (b) shows the PIDL-LSTM results across different gap lengths when the number of training data is fixed as 20. Note that when the gap length equals to 0, it means that LSTM just uses data of time step $t$ as how the ANN does.  We can see that the LSTM-based PIDL can only outperform the ANN-based PIDL when the gap length equals to 1 second, 
this is partially because the LSTM-based PIDL model of longer gap lengths demand more training data to achieve a similar performance level as that of shorter gap lengths, and thus may not be amenable when we have sparse observations. 
In conclusion, the PIDL with more complex NN architecture requires more data to train and may not outperform those with simpler NN architecture, especially when data is insufficient. 
When the training data grows, LSTM-PIDL is superior to the ANN-based PIDL.
}

\subsection{Sensitivity analysis on different datasets}

\tcck{
In this section, we will further validate the PIDL method by varying data. We first present how the PIDL deals with variability of training data on US-101. Then we will apply the PIDL on I-80 data, another highway in the NGSIM dataset. 
}

\subsubsection{Variability of training data}

\begin{figure}[h]
   \centering
   \includegraphics[width=0.5\textwidth]{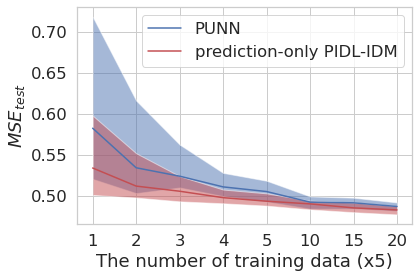}
   \caption{The performance comparison of the prediction-only PIDL-IDM and PUNN over 50 experiments.}
    \label{fig:results:box_plot}
\end{figure}

\tcck{
 To show that the advantage of the PIDL model persists across different training data, we \tcck{trained} the prediction-only PIDL-IDM model and the baselines for 50 times with randomly sampled data each time, and evaluate the overall performance. The comparison is shown in Fig.~\ref{fig:results:box_plot}. The x-axis is the number of training data and the y-axis is the test MSE. 
 The solid lines are the test MSE median among all 50 experiments: the red for the prediction-only PIDL-IDM and the blue for the PUNN. 
 The bands stand for the first and the third quartiles of the test MSE across 50 experiments. 
 The median test errors of both models decrease as the training data size increases, so are their bandwidths. 
 The overall performance of the PIDL model beats the pure data-driven model even with training data variability, but the difference shrinks with more training data.
 }

\subsubsection{I-80 dataset}

\begin{figure}[H]
   \centering
   \subfloat[][]{\includegraphics[height=0.41\textwidth,]{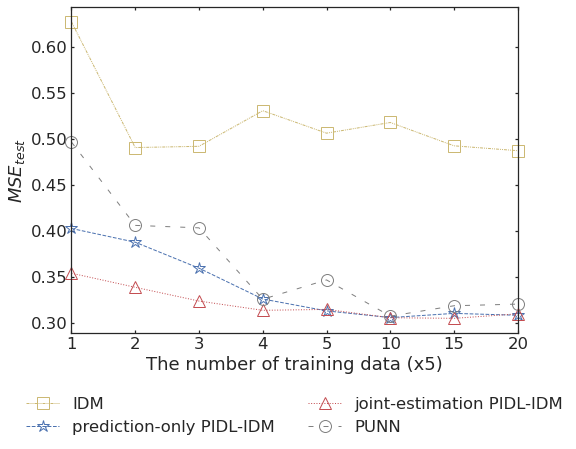}}  
   \subfloat[][]{\includegraphics[height=0.41\textwidth]{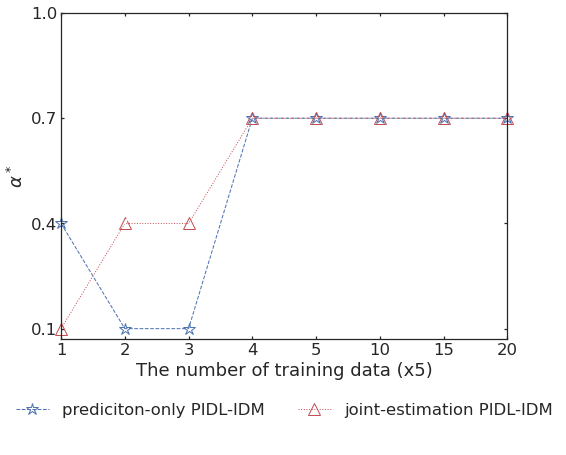}} 
   \caption{Prediction-only and joint-estimation for PIDL-IDM one-step calibration and evaluation on the NGSIM (I-80) data.}
   \label{fig:results:one_step_I80}
\end{figure}

\tcck{
We then apply our PIDL method to the I-80 dataset. It consists of a total of 45 minutes multi-lane vehicle data collected every 0.1 second. The data is processed the same way as the aforementioned US-101 dataset. 
The results of the prediction-only PIDL-IDM model together with the baselines are shown in Fig.~\ref{fig:results:one_step_I80}. The x-axis is the number of training data, and the y-axis is the test MSE. The line specification for each model is the same as in Fig.~\ref{fig:results:one_step_all}(a). 
We can see that both the prediction-only and joint-estimation PIDL-IDM models outperform their individual components, i.e., the baseline IDM and PUNN, respectively. Furthermore, the joint-estimation PIDL-IDM model achieves a better performance than the prediction-only PIDL-IDM model. 
The advantage of the PIDL-IDM models over PUNN becomes less significant as the training data size increases, 
which is the same as observed in the NGSIM results in Fig.~\ref{fig:results:one_step_all}. From this figure, we conclude that the superiority of the PIDL method persists for the I-80 dataset for both the prediction-only and joint-estimation problems.
}

\section{Conclusions}\label{sec:conclusion}
This paper lays a methodological framework of physics-informed deep learning (PIDL) for car-following modeling. It leverages the merits of both physics-based models and deep learning models. 
Accordingly, this modeling framework not only results in a higher estimation accuracy than either physics-based or neural network based models, their training is also data-efficient and generalizable.
We first propose the architectures of PIDL for two popular CFMs: IDM and OVM. 
Then we formulate the physics modeling framework with the designed loss function containing one term describing the deviation from data and one indicating the distance from physics prediction. 
To validate this approach, we first generate simulated data from an IDM or OVM model that is then used as observational data to estimate one's acceleration. We compare its estimation errors with the simulated values and find that the PIDL-CF outperforms the physics-uninformed neural networks. 
Then we test the approach on the NGSIM data. 
Building on a comprehensive set of numerical experiments, we find that our PIDL model is capable of not only predicting accelerations and but also estimating physics parameters on the real-world data. 
The developed PIDL-CF framework is beneficial for system identification of driving models and for the development of driving-based controls for automated vehicles.

\tcck{
To the best of our knowledge, this paper is the first-of-its-kind that employs a hybrid PIDL paradigm where a physics-based CFM is encoded into a neural network for car-following behavior modeling. 
}
Thus there are still a lot of unresolved research questions. 
This work will be extended in several directions. 
\tcck{
(1) More advanced CFMs such as \citep{li2012prediction} will be encoded into the PUNN, which may hold the potential to capturing more complex car-following behaviors.
(2) A systematic procedure of identification of which physics-based models may better fit real-world data should be developed. 
(3) Driver heterogeneity will be accounted for in the PIDL framework. 
One possible direction worth of exploration is to integrate the framework developed in this paper and that in \cite{yuan2020modeling}.
}

\section*{Acknowledgements}
This work is sponsored by the Region 2 University Transportation Research Center (UTRC) (subcontract RUTGER PO 966112/PID824227) 
and Amazon AWS Machine Learning Research Award Gift (\#US3085926).

\appendix
\tcck{\section{Metrics for model accuracy}}
\label{appen:a}
MSE is used in the experiments in Sec.~\ref{sec:numerical} and Sec.~\ref{sec:PIDL_for_ngsim} to measure prediction accuracy. We also tried other 3 metrics for the calibration of physics - Mean Absolute Error (MAE), Mean Square Error (RMSE), and Root Mean Square Percentage Error (RMSPE).  These metrics, together with the aforementioned MSE, are depicted as follows: 
\begin{itemize}
    \item MSE: \\
    \begin{equation}
     \begin{split}
     \frac{1}{N_O}\sum_{i=1}^{N_O} \left|f_{\lambda}(\hat{\mathbf{s}}^{(i)} | \lambda) - \hat{a}^{(i)}\right|^2
        \end{split}
     \end{equation}
 
    \item MAE: \\
    \begin{equation}
     \begin{split}
     \frac{1}{N_O}\sum_{i=1}^{N_O} \left|f_{\lambda}(\hat{\mathbf{s}}^{(i)} | \lambda) - \hat{a}^{(i)}\right|
        \end{split}
     \end{equation}
     
     \item RMSE: \\
     \begin{equation}
     \begin{split}
     \sqrt{
     \frac{1}{N_O}\sum_{i=1}^{N_O} \left|f_{\lambda}(\hat{\mathbf{s}}^{(i)} | \lambda) - \hat{a}^{(i)}\right|^2
     }
        \end{split}
     \end{equation}
     
     \item RMSPE: \\
     \begin{equation}
     \begin{split}
     \sqrt{
     \frac{1}{N_O}\sum_{i=1}^{N_O} \left|\frac{f_{\lambda}(\hat{\mathbf{s}}^{(i)} | \lambda) - \hat{a}^{(i)}}{\hat{a}^{(i)}}\right|^2 
     }\times 100\%
        \end{split}
     \end{equation}
     
     where, \\
     $\lambda$: physics parameters;\\
     $f_{\lambda}(\cdot)$: physics parameterized by ${\lambda}$;\\
     $\hat{\mathbf{s}}^{(i)}$: the $i$th observed feature; \\
     $a_{phy}^{(j)}$: the $j$th acceleration predicted by the physics.

\end{itemize}

To investigate how different metrics used in training affect the model performance, we use the same training data but with different metrics as the objective function. For comparison, we use the MSE during the evaluation on test dataset to provide a uniform evaluation metric. NGSIM US-101 data is used, and the size of training data is varied as in Sec.~\ref{sec:numerical} and Sec.~\ref{sec:PIDL_for_ngsim}. Results are shown in Table.~\ref{table:phy_metric}.

\begin{table}[h]
\begin{center}

\caption{\label{table:phy_metric}Test MSEs of models with different metrics in the training process.}
\begin{tabular}{ c c c c c }
\toprule
\multirow{2}{*}{$n_O (\times 5)$} & \multicolumn{4}{c}{Metrics used in training}  \\
\cmidrule{2-5}  
 & MSE &  MAE & RMSPE & RMSE  \\
\midrule
1 & 0.632 & 0.825 & 0.680 & 1.472 \\
2 & 0.626 & 0.653 & 0.679 & 0.623 \\
3 & 0.619 & 0.637 & 0.645 & 0.619 \\
4 & 0.611 & 0.612 & 0.633 & 0.595 \\
5 & 0.609 & 0.625 & 0.643 & 0.668 \\
10 & 0.601 & 0.612 & 0.647 & 0.641 \\
15 & 0.647 & 0.621 & 0.630 & 0.613 \\
20 & 0.592 & 0.581 & 0.619 & 0.600 \\
\bottomrule
\end{tabular}
\end{center}
\end{table}

We can see that models with different training metrics achieve similar performance on the same test dataset. Thus, we believe the PIDL-CFM framework is agnostic of the training metrics.

\tcck{\section{Selection of neural network structure in the numerical experiment}}
\label{appen:b}
\tcck{
We present the details of determining the structure of the neural networks here, which contains two parameters, the number of hidden layers and the number of neurons in each layer. Candidates of these two parameters are introduced in Sec.~\ref{sec:pidl-idm-exp-set}. We use numerical data of the accelerating regime to evaluate the performance of different model structure. The number of training data ($n_O$) equals to 100, and the weights of the loss function ($\alpha$) equals to 1, which means physics is not involved. The learning rate of the PUNN ($lr_{\text{PUNN}}$) is set to be $10^{-3}$. The early-stopping criterion is adopted during the training process, and the test MSE for the epoch with the minimal validation error is collected for comparison.
}

\begin{table}[h!]
\begin{center}
\caption{\label{table:nn_structure_testmse}Test MSEs of neural networks with different structure in the numerical experiment.}
\begin{tabular}{ c c c c c c }
\toprule
\multirow{2}{*}{Number of hidden layers} & \multicolumn{5}{c}{Number of neutrons in each layer}  \\
\cmidrule{2-6}  
 & 30 &  60 & 128 & 256 & 512  \\
\midrule
1 & 2.26 $\times 10^{-2}$ & 2.39 $\times 10^{-2}$ & 2.47 $\times 10^{-2}$ & 2.21 $\times 10^{-2}$ & 2.00 $\times 10^{-2}$ \\
2 & 2.20 $\times 10^{-2}$ & 2.11 $\times 10^{-2}$ & 2.12 $\times 10^{-2}$ & 2.18 $\times 10^{-2}$ & 2.16 $\times 10^{-2}$ \\
3 & 2.34 $\times 10^{-2}$ & 2.04 $\times 10^{-2}$ & 2.11 $\times 10^{-2}$ & 2.05 $\times 10^{-2}$ & 1.97 $\times 10^{-2}$ \\
4 & 2.19 $\times 10^{-2}$ & 2.16 $\times 10^{-2}$ & 2.36 $\times 10^{-2}$ & 1.94 $\times 10^{-2}$ & 1.73 $\times 10^{-2}$\\
5 & 2.10 $\times 10^{-2}$ & 2.22 $\times 10^{-2}$ & 1.63 $\times 10^{-2}$ & 1.97 $\times 10^{-2}$ & 2.03 $\times 10^{-2}$ \\
\bottomrule
\end{tabular}
\end{center}
\end{table}
\tcck{
The results are shown in Table.~\ref{table:nn_structure_testmse}. The first column is the number of layers, and the first row is the number of neurons in each layer. Other entries are the test error of the corresponding model structure adopting the early-stopping during training. Note that the test data is held out beforehand, so the test error reflects the generalization ability of the model. The table shows that a complex model, if early-stopping is adopted, tends to have a lower test error. However, we did not select a model structure that is too complex, such as the one containing 4 layers with 512 neurons in each layer, because a complex model needs more time to train. After balancing the test error and training time, we select the structure of 3 layers with 60 neutrons in each layer.
}

\bibliographystyle{elsarticle-harv}
\bibliography{lit_dementia,literatureCV,repeated_ref,survey,survey_DCL,survey_Kuang,survey_mdp,survey_MFG,survey_rongye,survey_Zhaobin, Car-Following,survey_Di}

\begin{thebibliography}{67}
\expandafter\ifx\csname natexlab\endcsname\relax\def\natexlab#1{#1}\fi
\providecommand{\url}[1]{\texttt{#1}}
\providecommand{\href}[2]{#2}
\providecommand{\path}[1]{#1}
\providecommand{\DOIprefix}{doi:}
\providecommand{\ArXivprefix}{arXiv:}
\providecommand{\URLprefix}{URL: }
\providecommand{\Pubmedprefix}{pmid:}
\providecommand{\doi}[1]{\href{http://dx.doi.org/#1}{\path{#1}}}
\providecommand{\Pubmed}[1]{\href{pmid:#1}{\path{#1}}}
\providecommand{\bibinfo}[2]{#2}
\ifx\xfnm\relax \def\xfnm[#1]{\unskip,\space#1}\fi
\bibitem[{Abadi(2016)}]{abadi2016tensorflow}
\bibinfo{author}{Abadi, M.}, \bibinfo{year}{2016}.
\newblock \bibinfo{title}{Tensorflow: learning functions at scale}, in:
  \bibinfo{booktitle}{Proceedings of the 21st ACM SIGPLAN International
  Conference on Functional Programming}, pp. \bibinfo{pages}{1--1}.
\bibitem[{Alber et~al.(2019)Alber, Tepole, Cannon, De, Dura-Bernal, Garikipati,
  Karniadakis, Lytton, Perdikaris, Petzold and Kuhl}]{alber2019integrating}
\bibinfo{author}{Alber, M.}, \bibinfo{author}{Tepole, A.B.},
  \bibinfo{author}{Cannon, W.R.}, \bibinfo{author}{De, S.},
  \bibinfo{author}{Dura-Bernal, S.}, \bibinfo{author}{Garikipati, K.},
  \bibinfo{author}{Karniadakis, G.}, \bibinfo{author}{Lytton, W.W.},
  \bibinfo{author}{Perdikaris, P.}, \bibinfo{author}{Petzold, L.},
  \bibinfo{author}{Kuhl, E.}, \bibinfo{year}{2019}.
\newblock \bibinfo{title}{Integrating machine learning and multiscale
  modeling—perspectives, challenges, and opportunities in the biological,
  biomedical, and behavioral sciences}.
\newblock \bibinfo{journal}{npj Digital Medicine} \bibinfo{volume}{2},
  \bibinfo{pages}{1--11}.
\bibitem[{Bando et~al.(1995)Bando, Hasebe and et~al.}]{Bando-1995}
\bibinfo{author}{Bando, M.}, \bibinfo{author}{Hasebe, K.},
  \bibinfo{author}{et~al.}, \bibinfo{year}{1995}.
\newblock \bibinfo{title}{Dynamical model of traffic congestion and numerical
  simulation}.
\newblock \bibinfo{journal}{Physical Review E} \bibinfo{volume}{51},
  \bibinfo{pages}{1035}.
\bibitem[{Batista and Twrdy(2010)}]{batista2010optimal}
\bibinfo{author}{Batista, M.}, \bibinfo{author}{Twrdy, E.},
  \bibinfo{year}{2010}.
\newblock \bibinfo{title}{Optimal velocity functions for car-following models}.
\newblock \bibinfo{journal}{Journal of Zhejiang University-SCIENCE A}
  \bibinfo{volume}{11}, \bibinfo{pages}{520--529}.
\bibitem[{Coifman and Li(2017)}]{coifman2017critical}
\bibinfo{author}{Coifman, B.}, \bibinfo{author}{Li, L.}, \bibinfo{year}{2017}.
\newblock \bibinfo{title}{A critical evaluation of the next generation
  simulation (ngsim) vehicle trajectory dataset}.
\newblock \bibinfo{journal}{Transportation Research Part B: Methodological}
  \bibinfo{volume}{105}, \bibinfo{pages}{362--377}.
\bibitem[{Di and Shi(2021)}]{di2021survey}
\bibinfo{author}{Di, X.}, \bibinfo{author}{Shi, R.}, \bibinfo{year}{2021}.
\newblock \bibinfo{title}{A survey on autonomous vehicle control in the era of
  mixed-autonomy: From physics-based to {AI}-guided driving policy learning}.
\newblock \bibinfo{journal}{Transportation Research Part C: Emerging
  Technologies} \bibinfo{volume}{125}, \bibinfo{pages}{103008}.
\bibitem[{Di~Giovanni et~al.(2020)Di~Giovanni, Sondak, Protopapas and
  Brambilla}]{di2020finding}
\bibinfo{author}{Di~Giovanni, M.}, \bibinfo{author}{Sondak, D.},
  \bibinfo{author}{Protopapas, P.}, \bibinfo{author}{Brambilla, M.},
  \bibinfo{year}{2020}.
\newblock \bibinfo{title}{Finding multiple solutions of odes with neural
  networks}, in: \bibinfo{booktitle}{AAAI-MLPS 2020},
  \bibinfo{organization}{CEUR-WS}. pp. \bibinfo{pages}{1--7}.
\bibitem[{Durrani et~al.(2016)Durrani, Lee and Maoh}]{durrani2016calibrating}
\bibinfo{author}{Durrani, U.}, \bibinfo{author}{Lee, C.},
  \bibinfo{author}{Maoh, H.}, \bibinfo{year}{2016}.
\newblock \bibinfo{title}{Calibrating the wiedemann’s vehicle-following model
  using mixed vehicle-pair interactions}.
\newblock \bibinfo{journal}{Transportation research part C: emerging
  technologies} \bibinfo{volume}{67}, \bibinfo{pages}{227--242}.
\bibitem[{Fang and Zhan(2020)}]{Fang-2020}
\bibinfo{author}{Fang, Z.}, \bibinfo{author}{Zhan, J.}, \bibinfo{year}{2020}.
\newblock \bibinfo{title}{Physics-informed neural network framework for partial
  differential equations on 3{D} surfaces: Time independent problems}.
\newblock \bibinfo{journal}{IEEE Access} .
\bibitem[{Fritzsche(1994)}]{fritzsche1994model}
\bibinfo{author}{Fritzsche, H.T.}, \bibinfo{year}{1994}.
\newblock \bibinfo{title}{A model for traffic simulation}.
\newblock \bibinfo{journal}{Traffic Engineering+ Control} \bibinfo{volume}{35},
  \bibinfo{pages}{317--21}.
\bibitem[{Gazis et~al.(1961)Gazis, Herman and Rothery}]{gazis1961nonlinear}
\bibinfo{author}{Gazis, D.C.}, \bibinfo{author}{Herman, R.},
  \bibinfo{author}{Rothery, R.W.}, \bibinfo{year}{1961}.
\newblock \bibinfo{title}{Nonlinear follow-the-leader models of traffic flow}.
\newblock \bibinfo{journal}{Operations research} \bibinfo{volume}{9},
  \bibinfo{pages}{545--567}.
\bibitem[{Gipps(1981)}]{gipps1981behavioural}
\bibinfo{author}{Gipps, P.G.}, \bibinfo{year}{1981}.
\newblock \bibinfo{title}{A behavioural car-following model for computer
  simulation}.
\newblock \bibinfo{journal}{Transportation Research Part B: Methodological}
  \bibinfo{volume}{15}, \bibinfo{pages}{105--111}.
\bibitem[{Greenshields et~al.(1935)Greenshields, Bibbins, Channing and
  Miller}]{greenshields1935study}
\bibinfo{author}{Greenshields, B.}, \bibinfo{author}{Bibbins, J.},
  \bibinfo{author}{Channing, W.}, \bibinfo{author}{Miller, H.},
  \bibinfo{year}{1935}.
\newblock \bibinfo{title}{A study of traffic capacity}, in:
  \bibinfo{booktitle}{Highway research board proceedings},
  \bibinfo{organization}{National Research Council (USA), Highway Research
  Board}.
\bibitem[{Gu et~al.(2020)Gu, Li, Di and Shi}]{gu2020lstm}
\bibinfo{author}{Gu, Z.}, \bibinfo{author}{Li, Z.}, \bibinfo{author}{Di, X.},
  \bibinfo{author}{Shi, R.}, \bibinfo{year}{2020}.
\newblock \bibinfo{title}{An lstm-based autonomous driving model using a waymo
  open dataset}.
\newblock \bibinfo{journal}{Applied Sciences} \bibinfo{volume}{10},
  \bibinfo{pages}{2046}.
\bibitem[{He et~al.(2019)He, Shi and Song}]{he2019weight}
\bibinfo{author}{He, D.}, \bibinfo{author}{Shi, Y.}, \bibinfo{author}{Song,
  X.}, \bibinfo{year}{2019}.
\newblock \bibinfo{title}{Weight-free multi-objective predictive cruise control
  of autonomous vehicles in integrated perturbation analysis and sequential
  quadratic programming optimization framework}.
\newblock \bibinfo{journal}{Journal of Dynamic Systems, Measurement, and
  Control} \bibinfo{volume}{141}.
\bibitem[{He et~al.(2015)He, Zheng and Guan}]{he2015simple}
\bibinfo{author}{He, Z.}, \bibinfo{author}{Zheng, L.}, \bibinfo{author}{Guan,
  W.}, \bibinfo{year}{2015}.
\newblock \bibinfo{title}{A simple nonparametric car-following model driven by
  field data}.
\newblock \bibinfo{journal}{Transportation Research Part B: Methodological}
  \bibinfo{volume}{80}, \bibinfo{pages}{185--201}.
\bibitem[{Helly(1959)}]{helly1959simulation}
\bibinfo{author}{Helly, W.}, \bibinfo{year}{1959}.
\newblock \bibinfo{title}{Simulation of bottlenecks in single-lane traffic
  flow} .
\bibitem[{Huang et~al.(2019)Huang, Di, Du and Chen}]{huang2019stable}
\bibinfo{author}{Huang, K.}, \bibinfo{author}{Di, X.}, \bibinfo{author}{Du,
  Q.}, \bibinfo{author}{Chen, X.}, \bibinfo{year}{2019}.
\newblock \bibinfo{title}{Stabilizing traffic via autonomous vehicles: A
  continuum mean field game approach}.
\newblock \bibinfo{journal}{the 22nd IEEE International Conference on
  Intelligent Transportation Systems (ITSC) (DOI: 10.1109/ITSC.2019.8917021)} .
\bibitem[{Huang et~al.(2020)Huang, Di, Du and Chen}]{huang2020stable}
\bibinfo{author}{Huang, K.}, \bibinfo{author}{Di, X.}, \bibinfo{author}{Du,
  Q.}, \bibinfo{author}{Chen, X.}, \bibinfo{year}{2020}.
\newblock \bibinfo{title}{Scalable traffic stability analysis in mixed-autonomy
  using continuum models}.
\newblock \bibinfo{journal}{Transportation Research Part C: Emerging
  Technologies} \bibinfo{volume}{111}, \bibinfo{pages}{616--630}.
\bibitem[{Huang et~al.(2018a)Huang, Sun and Sun}]{huang2018car}
\bibinfo{author}{Huang, X.}, \bibinfo{author}{Sun, J.}, \bibinfo{author}{Sun,
  J.}, \bibinfo{year}{2018}a.
\newblock \bibinfo{title}{A car-following model considering asymmetric driving
  behavior based on long short-term memory neural networks}.
\newblock \bibinfo{journal}{Transportation research part C: emerging
  technologies} \bibinfo{volume}{95}, \bibinfo{pages}{346--362}.
\bibitem[{Huang et~al.(2018b)Huang, Jiang, Zhang, Hu, Tian, Jia and
  Gao}]{huang2018experimental}
\bibinfo{author}{Huang, Y.X.}, \bibinfo{author}{Jiang, R.},
  \bibinfo{author}{Zhang, H.}, \bibinfo{author}{Hu, M.B.},
  \bibinfo{author}{Tian, J.F.}, \bibinfo{author}{Jia, B.},
  \bibinfo{author}{Gao, Z.Y.}, \bibinfo{year}{2018}b.
\newblock \bibinfo{title}{Experimental study and modeling of car-following
  behavior under high speed situation}.
\newblock \bibinfo{journal}{Transportation research part C: emerging
  technologies} \bibinfo{volume}{97}, \bibinfo{pages}{194--215}.
\bibitem[{Jiang et~al.(2001)Jiang, Wu and Zhu}]{jiang2001full}
\bibinfo{author}{Jiang, R.}, \bibinfo{author}{Wu, Q.}, \bibinfo{author}{Zhu,
  Z.}, \bibinfo{year}{2001}.
\newblock \bibinfo{title}{Full velocity difference model for a car-following
  theory}.
\newblock \bibinfo{journal}{Physical Review E} \bibinfo{volume}{64},
  \bibinfo{pages}{017101}.
\bibitem[{Jin and Orosz(2014)}]{jin2014dynamics}
\bibinfo{author}{Jin, I.G.}, \bibinfo{author}{Orosz, G.}, \bibinfo{year}{2014}.
\newblock \bibinfo{title}{Dynamics of connected vehicle systems with delayed
  acceleration feedback}.
\newblock \bibinfo{journal}{Transportation Research Part C: Emerging
  Technologies} \bibinfo{volume}{46}, \bibinfo{pages}{46--64}.
\bibitem[{Jin et~al.(2010)Jin, Wang, Tao and Li}]{jin2010non}
\bibinfo{author}{Jin, S.}, \bibinfo{author}{Wang, D.}, \bibinfo{author}{Tao,
  P.}, \bibinfo{author}{Li, P.}, \bibinfo{year}{2010}.
\newblock \bibinfo{title}{Non-lane-based full velocity difference car following
  model}.
\newblock \bibinfo{journal}{Physica A: Statistical Mechanics and Its
  Applications} \bibinfo{volume}{389}, \bibinfo{pages}{4654--4662}.
\bibitem[{Karlik and Olgac(2011)}]{karlik2011performance}
\bibinfo{author}{Karlik, B.}, \bibinfo{author}{Olgac, A.V.},
  \bibinfo{year}{2011}.
\newblock \bibinfo{title}{Performance analysis of various activation functions
  in generalized mlp architectures of neural networks}.
\newblock \bibinfo{journal}{International Journal of Artificial Intelligence
  and Expert Systems} \bibinfo{volume}{1}, \bibinfo{pages}{111--122}.
\bibitem[{Karpatne et~al.(2017)Karpatne, Atluri and et~al}]{Karpatne-2017}
\bibinfo{author}{Karpatne, A.}, \bibinfo{author}{Atluri, G.},
  \bibinfo{author}{et~al}, \bibinfo{year}{2017}.
\newblock \bibinfo{title}{Theory-guided data science: A new paradigm for
  scientific discovery from data}.
\newblock \bibinfo{journal}{IEEE Transactions on Knowledge and Data
  Engineering} \bibinfo{volume}{29}, \bibinfo{pages}{2318--2331}.
\bibitem[{Kuefler et~al.(2017)Kuefler, Morton, Wheeler and
  Kochenderfer}]{kuefler2017imitating}
\bibinfo{author}{Kuefler, A.}, \bibinfo{author}{Morton, J.},
  \bibinfo{author}{Wheeler, T.}, \bibinfo{author}{Kochenderfer, M.},
  \bibinfo{year}{2017}.
\newblock \bibinfo{title}{Imitating driver behavior with generative adversarial
  networks}, in: \bibinfo{booktitle}{2017 IEEE Intelligent Vehicles Symposium
  (IV)}, \bibinfo{organization}{IEEE}. pp. \bibinfo{pages}{204--211}.
\bibitem[{Kumar et~al.(2010)Kumar, Husain, Upreti and Gupta}]{kumar2010genetic}
\bibinfo{author}{Kumar, M.}, \bibinfo{author}{Husain, M.},
  \bibinfo{author}{Upreti, N.}, \bibinfo{author}{Gupta, D.},
  \bibinfo{year}{2010}.
\newblock \bibinfo{title}{Genetic algorithm: Review and application}.
\newblock \bibinfo{journal}{Available at SSRN 3529843} .
\bibitem[{Lazar et~al.(2016)Lazar, Rhoulami and Rahmani}]{lazar2016review}
\bibinfo{author}{Lazar, H.}, \bibinfo{author}{Rhoulami, K.},
  \bibinfo{author}{Rahmani, D.}, \bibinfo{year}{2016}.
\newblock \bibinfo{title}{A review analysis of optimal velocity models}.
\newblock \bibinfo{journal}{Periodica Polytechnica Transportation Engineering}
  \bibinfo{volume}{44}, \bibinfo{pages}{123--131}.
\bibitem[{Li et~al.(2012)Li, Wang and Ouyang}]{li2012prediction}
\bibinfo{author}{Li, X.}, \bibinfo{author}{Wang, X.}, \bibinfo{author}{Ouyang,
  Y.}, \bibinfo{year}{2012}.
\newblock \bibinfo{title}{Prediction and field validation of traffic
  oscillation propagation under nonlinear car-following laws}.
\newblock \bibinfo{journal}{Transportation research part B: methodological}
  \bibinfo{volume}{46}, \bibinfo{pages}{409--423}.
\bibitem[{Nageshrao et~al.(2019)Nageshrao, Tseng and
  Filev}]{NN_adam_nageshrao2019autonomous}
\bibinfo{author}{Nageshrao, S.}, \bibinfo{author}{Tseng, E.},
  \bibinfo{author}{Filev, D.}, \bibinfo{year}{2019}.
\newblock \bibinfo{title}{Autonomous highway driving using deep reinforcement
  learning}.
\newblock \bibinfo{journal}{arXiv preprint arXiv:1904.00035} .
\bibitem[{Ossen and Hoogendoorn(2005)}]{ossen2005car}
\bibinfo{author}{Ossen, S.}, \bibinfo{author}{Hoogendoorn, S.P.},
  \bibinfo{year}{2005}.
\newblock \bibinfo{title}{Car-following behavior analysis from microscopic
  trajectory data}.
\newblock \bibinfo{journal}{Transportation Research Record}
  \bibinfo{volume}{1934}, \bibinfo{pages}{13--21}.
\bibitem[{Ossen and Hoogendoorn(2011)}]{ossen2011heterogeneity}
\bibinfo{author}{Ossen, S.}, \bibinfo{author}{Hoogendoorn, S.P.},
  \bibinfo{year}{2011}.
\newblock \bibinfo{title}{Heterogeneity in car-following behavior: Theory and
  empirics}.
\newblock \bibinfo{journal}{Transportation research part C: emerging
  technologies} \bibinfo{volume}{19}, \bibinfo{pages}{182--195}.
\bibitem[{Panwai and Dia(2007)}]{Panwai2007}
\bibinfo{author}{Panwai, S.}, \bibinfo{author}{Dia, H.}, \bibinfo{year}{2007}.
\newblock \bibinfo{title}{{Neural agent car-following models}}.
\newblock \bibinfo{journal}{IEEE Transactions on Intelligent Transportation
  Systems} \bibinfo{volume}{8}, \bibinfo{pages}{60--70}.
\newblock \DOIprefix\doi{10.1109/TITS.2006.884616}.
\bibitem[{Rahman et~al.(2017)Rahman, Chowdhury, Dey, Islam and
  Khan}]{rahman2017evaluation}
\bibinfo{author}{Rahman, M.}, \bibinfo{author}{Chowdhury, M.},
  \bibinfo{author}{Dey, K.}, \bibinfo{author}{Islam, M.R.},
  \bibinfo{author}{Khan, T.}, \bibinfo{year}{2017}.
\newblock \bibinfo{title}{Evaluation of driver car-following behavior models
  for cooperative adaptive cruise control systems}.
\newblock \bibinfo{journal}{Transportation Research Record}
  \bibinfo{volume}{2622}, \bibinfo{pages}{84--95}.
\bibitem[{Rai and Sahu(2020)}]{rai2020driven}
\bibinfo{author}{Rai, R.}, \bibinfo{author}{Sahu, C.K.}, \bibinfo{year}{2020}.
\newblock \bibinfo{title}{Driven by data or derived through physics? a review
  of hybrid physics guided machine learning techniques with cyber-physical
  system (cps) focus}.
\newblock \bibinfo{journal}{IEEE Access} \bibinfo{volume}{8},
  \bibinfo{pages}{71050--71073}.
\bibitem[{Raissi(2018)}]{Raissi-2018a}
\bibinfo{author}{Raissi, M.}, \bibinfo{year}{2018}.
\newblock \bibinfo{title}{Deep hidden physics models: Deep learning of
  nonlinear partial differential equations}.
\newblock \bibinfo{journal}{Journal of Machine Learning Research}
  \bibinfo{volume}{19}, \bibinfo{pages}{932--955}.
\bibitem[{Raissi and Karniadakis(2018)}]{Raissi-2018b}
\bibinfo{author}{Raissi, M.}, \bibinfo{author}{Karniadakis, G.E.},
  \bibinfo{year}{2018}.
\newblock \bibinfo{title}{Hidden physics models: Machine learning of nonlinear
  partial differential equations}.
\newblock \bibinfo{journal}{Journal of Computational Physics}
  \bibinfo{volume}{357}, \bibinfo{pages}{125--141}.
\bibitem[{Raissi et~al.(2019a)Raissi, Perdikaris and
  Karniadakis}]{raissi2019physics}
\bibinfo{author}{Raissi, M.}, \bibinfo{author}{Perdikaris, P.},
  \bibinfo{author}{Karniadakis, G.E.}, \bibinfo{year}{2019}a.
\newblock \bibinfo{title}{Physics-informed neural networks: A deep learning
  framework for solving forward and inverse problems involving nonlinear
  partial differential equations}.
\newblock \bibinfo{journal}{Journal of Computational Physics}
  \bibinfo{volume}{378}, \bibinfo{pages}{686--707}.
\bibitem[{Raissi et~al.(2019b)Raissi, Wang, Triantafyllou and
  Karniadakis}]{Maziar-2019}
\bibinfo{author}{Raissi, M.}, \bibinfo{author}{Wang, Z.},
  \bibinfo{author}{Triantafyllou, M.S.}, \bibinfo{author}{Karniadakis, G.E.},
  \bibinfo{year}{2019}b.
\newblock \bibinfo{title}{Deep learning of vortex-induced vibrations}.
\newblock \bibinfo{journal}{Journal of Fluid Mechanics} \bibinfo{volume}{861},
  \bibinfo{pages}{119--137}.
\bibitem[{Raissi et~al.(2020)Raissi, Yazdani and Karniadakis}]{Maziar-2020}
\bibinfo{author}{Raissi, M.}, \bibinfo{author}{Yazdani, A.},
  \bibinfo{author}{Karniadakis, G.E.}, \bibinfo{year}{2020}.
\newblock \bibinfo{title}{Hidden fluid mechanics: Learning velocity and
  pressure fields from flow visualizations}.
\newblock \bibinfo{journal}{Science} \bibinfo{volume}{367},
  \bibinfo{pages}{1026--1030}.
\bibitem[{Roehrl et~al.(2020)Roehrl, Runkler, Brandtstetter, Tokic and
  Obermayer}]{roehrl2020modeling}
\bibinfo{author}{Roehrl, M.A.}, \bibinfo{author}{Runkler, T.A.},
  \bibinfo{author}{Brandtstetter, V.}, \bibinfo{author}{Tokic, M.},
  \bibinfo{author}{Obermayer, S.}, \bibinfo{year}{2020}.
\newblock \bibinfo{title}{Modeling system dynamics with physics-informed neural
  networks based on lagrangian mechanics}.
\newblock \bibinfo{journal}{arXiv preprint arXiv:2005.14617} .
\bibitem[{Sharma et~al.(2019)Sharma, Zheng and Bhaskar}]{sharma2019more}
\bibinfo{author}{Sharma, A.}, \bibinfo{author}{Zheng, Z.},
  \bibinfo{author}{Bhaskar, A.}, \bibinfo{year}{2019}.
\newblock \bibinfo{title}{Is more always better? the impact of vehicular
  trajectory completeness on car-following model calibration and validation}.
\newblock \bibinfo{journal}{Transportation research part B: methodological}
  \bibinfo{volume}{120}, \bibinfo{pages}{49--75}.
\bibitem[{Shi et~al.(2021a)Shi, Mo and Di}]{shi2020aaai}
\bibinfo{author}{Shi, R.}, \bibinfo{author}{Mo, Z.}, \bibinfo{author}{Di, X.},
  \bibinfo{year}{2021}a.
\newblock \bibinfo{title}{Physics-informed deep learning for traffic state
  estimation: A hybrid paradigm informed by second-order traffic models}.
\newblock \bibinfo{journal}{35th AAAI Conference on Artificial Intelligence
  (AAAI 2021),}
  \bibinfo{volume}{\url{https://www.aaai.org/AAAI21Papers/AAAI-3617.ShiR.pdf}}.
\bibitem[{Shi et~al.(2021b)Shi, Mo, Huang, Di and Du}]{shi2020pidl}
\bibinfo{author}{Shi, R.}, \bibinfo{author}{Mo, Z.}, \bibinfo{author}{Huang,
  K.}, \bibinfo{author}{Di, X.}, \bibinfo{author}{Du, Q.},
  \bibinfo{year}{2021}b.
\newblock \bibinfo{title}{Physics-informed deep learning for traffic state
  estimation}.
\newblock \bibinfo{journal}{(under review)
  \url{https://arxiv.org/abs/2101.06580}} .
\bibitem[{Shou et~al.(2020)Shou, Wang, Han, Liu, Tiwari and Di}]{shou2020long}
\bibinfo{author}{Shou, Z.}, \bibinfo{author}{Wang, Z.}, \bibinfo{author}{Han,
  K.}, \bibinfo{author}{Liu, Y.}, \bibinfo{author}{Tiwari, P.},
  \bibinfo{author}{Di, X.}, \bibinfo{year}{2020}.
\newblock \bibinfo{title}{Long-term prediction of lane change maneuver through
  a multilayer perceptron}.
\newblock \bibinfo{journal}{2020 IEEE Intelligent Vehicles Symposium (IEEE IV
  2020)} .
\bibitem[{Sirignano and Spiliopoulos(2019)}]{sirignano2019scaling_Xavier}
\bibinfo{author}{Sirignano, J.}, \bibinfo{author}{Spiliopoulos, K.},
  \bibinfo{year}{2019}.
\newblock \bibinfo{title}{Scaling limit of neural networks with the xavier
  initialization and convergence to a global minimum}.
\newblock \bibinfo{journal}{arXiv preprint arXiv:1907.04108} .
\bibitem[{Treiber et~al.(2000a)Treiber, Hennecke and Helbing}]{Treiber-2000}
\bibinfo{author}{Treiber, M.}, \bibinfo{author}{Hennecke, A.},
  \bibinfo{author}{Helbing, D.}, \bibinfo{year}{2000}a.
\newblock \bibinfo{title}{Congested traffic states in empirical observations
  and microscopic simulations}.
\newblock \bibinfo{journal}{Physical review E} \bibinfo{volume}{62},
  \bibinfo{pages}{1805}.
\bibitem[{Treiber et~al.(2000b)Treiber, Hennecke and
  Helbing}]{treiber2000congested}
\bibinfo{author}{Treiber, M.}, \bibinfo{author}{Hennecke, A.},
  \bibinfo{author}{Helbing, D.}, \bibinfo{year}{2000}b.
\newblock \bibinfo{title}{Congested traffic states in empirical observations
  and microscopic simulations}.
\newblock \bibinfo{journal}{Physical review E} \bibinfo{volume}{62},
  \bibinfo{pages}{1805}.
\bibitem[{Treiber and Kesting(2013)}]{treiber2013microscopic}
\bibinfo{author}{Treiber, M.}, \bibinfo{author}{Kesting, A.},
  \bibinfo{year}{2013}.
\newblock \bibinfo{title}{Microscopic calibration and validation of
  car-following models--a systematic approach}.
\newblock \bibinfo{journal}{Procedia-Social and Behavioral Sciences}
  \bibinfo{volume}{80}, \bibinfo{pages}{922--939}.
\bibitem[{Wang et~al.(2017)Wang, Rakha and Fadhloun}]{wang2017comparison}
\bibinfo{author}{Wang, J.}, \bibinfo{author}{Rakha, H.A.},
  \bibinfo{author}{Fadhloun, K.}, \bibinfo{year}{2017}.
\newblock \bibinfo{title}{Comparison of car-following models: A vehicle fuel
  consumption and emissions estimation perspective}.
\newblock \bibinfo{type}{Technical Report}.
\bibitem[{Wei and Liu(2013)}]{wei2013analysis}
\bibinfo{author}{Wei, D.}, \bibinfo{author}{Liu, H.}, \bibinfo{year}{2013}.
\newblock \bibinfo{title}{Analysis of asymmetric driving behavior using a
  self-learning approach}.
\newblock \bibinfo{journal}{Transportation Research Part B: Methodological}
  \bibinfo{volume}{47}, \bibinfo{pages}{1--14}.
\bibitem[{Wiedemann and Reiter(1992)}]{wiedemann1992microscopic}
\bibinfo{author}{Wiedemann, R.}, \bibinfo{author}{Reiter, U.},
  \bibinfo{year}{1992}.
\newblock \bibinfo{title}{Microscopic traffic simulation: the simulation system
  mission, background and actual state, cec project icarus (v1052), final
  report, vol. 2, appendix a}.
\newblock \bibinfo{journal}{Brussels: CEC} .
\bibitem[{Wilson(2001)}]{wilson2001analysis}
\bibinfo{author}{Wilson, R.E.}, \bibinfo{year}{2001}.
\newblock \bibinfo{title}{An analysis of gipps's car-following model of highway
  traffic}.
\newblock \bibinfo{journal}{IMA journal of applied mathematics}
  \bibinfo{volume}{66}, \bibinfo{pages}{509--537}.
\bibitem[{Wu and Work(2018)}]{Wu-2018}
\bibinfo{author}{Wu, F.}, \bibinfo{author}{Work, D.B.}, \bibinfo{year}{2018}.
\newblock \bibinfo{title}{Connections between classical car following models
  and artificial neural networks}, in: \bibinfo{booktitle}{Proceedings of the
  21st International Conference on Intelligent Transportation Systems (ITSC))},
  \bibinfo{address}{Maui, HI, USA}. pp. \bibinfo{pages}{3191--3198}.
\bibitem[{Yang et~al.(2018)Yang, Zhu, Liu, Wu and Ran}]{yang2018novel}
\bibinfo{author}{Yang, D.}, \bibinfo{author}{Zhu, L.}, \bibinfo{author}{Liu,
  Y.}, \bibinfo{author}{Wu, D.}, \bibinfo{author}{Ran, B.},
  \bibinfo{year}{2018}.
\newblock \bibinfo{title}{A novel car-following control model combining machine
  learning and kinematics models for automated vehicles}.
\newblock \bibinfo{journal}{IEEE Transactions on Intelligent Transportation
  Systems} \bibinfo{volume}{20}, \bibinfo{pages}{1991--2000}.
\bibitem[{Yang and Perdikaris(2019)}]{Yang-2019}
\bibinfo{author}{Yang, Y.}, \bibinfo{author}{Perdikaris, P.},
  \bibinfo{year}{2019}.
\newblock \bibinfo{title}{Adversarial uncertainty quantification in
  physics-informed neural networks}.
\newblock \bibinfo{journal}{Journal of Computational Physics}
  \bibinfo{volume}{394}, \bibinfo{pages}{136--152}.
\bibitem[{Yu et~al.(2021)Yu, Canales-Rodr{\'\i}guez, Pizzolato, Piredda,
  Hilbert, Fischi-Gomez, Weigel, Barakovic, Cuadra, Granziera
  et~al.}]{yu2021model}
\bibinfo{author}{Yu, T.}, \bibinfo{author}{Canales-Rodr{\'\i}guez, E.J.},
  \bibinfo{author}{Pizzolato, M.}, \bibinfo{author}{Piredda, G.F.},
  \bibinfo{author}{Hilbert, T.}, \bibinfo{author}{Fischi-Gomez, E.},
  \bibinfo{author}{Weigel, M.}, \bibinfo{author}{Barakovic, M.},
  \bibinfo{author}{Cuadra, M.B.}, \bibinfo{author}{Granziera, C.}, et~al.,
  \bibinfo{year}{2021}.
\newblock \bibinfo{title}{Model-informed machine learning for multi-component
  t2 relaxometry}.
\newblock \bibinfo{journal}{Medical Image Analysis} \bibinfo{volume}{69},
  \bibinfo{pages}{101940}.
\bibitem[{Yuan et~al.(2020)Yuan, Wang and Yang}]{yuan2020modeling}
\bibinfo{author}{Yuan, Y.}, \bibinfo{author}{Wang, Q.}, \bibinfo{author}{Yang,
  X.T.}, \bibinfo{year}{2020}.
\newblock \bibinfo{title}{Modeling stochastic microscopic traffic behaviors: a
  physics regularized gaussian process approach}.
\newblock \bibinfo{journal}{arXiv preprint arXiv:2007.10109} .
\bibitem[{Zhou and Laval(2019)}]{zhou2019longitudinal}
\bibinfo{author}{Zhou, H.}, \bibinfo{author}{Laval, J.}, \bibinfo{year}{2019}.
\newblock \bibinfo{title}{Longitudinal motion planning for autonomous vehicles
  and its impact on congestion: A survey}.
\newblock \bibinfo{journal}{arXiv preprint arXiv:1910.06070} .
\bibitem[{Zhou et~al.(2017a)Zhou, Qu and Li}]{Zhou2017g}
\bibinfo{author}{Zhou, M.}, \bibinfo{author}{Qu, X.}, \bibinfo{author}{Li, X.},
  \bibinfo{year}{2017}a.
\newblock \bibinfo{title}{{A recurrent neural network based microscopic car
  following model to predict traffic oscillation}}.
\newblock \bibinfo{journal}{Transportation Research Part C: Emerging
  Technologies} \bibinfo{volume}{84}, \bibinfo{pages}{245--264}.
\newblock \URLprefix \url{http://dx.doi.org/10.1016/j.trc.2017.08.027},
  \DOIprefix\doi{10.1016/j.trc.2017.08.027}.
\bibitem[{Zhou and Ahn(2019)}]{zhou2019robust}
\bibinfo{author}{Zhou, Y.}, \bibinfo{author}{Ahn, S.}, \bibinfo{year}{2019}.
\newblock \bibinfo{title}{Robust local and string stability for a decentralized
  car following control strategy for connected automated vehicles}.
\newblock \bibinfo{journal}{Transportation Research Part B: Methodological}
  \bibinfo{volume}{125}, \bibinfo{pages}{175--196}.
\bibitem[{Zhou et~al.(2017b)Zhou, Ahn, Chitturi and Noyce}]{zhou-2017}
\bibinfo{author}{Zhou, Y.}, \bibinfo{author}{Ahn, S.},
  \bibinfo{author}{Chitturi, M.}, \bibinfo{author}{Noyce, D.A.},
  \bibinfo{year}{2017}b.
\newblock \bibinfo{title}{Rolling horizon stochastic optimal control strategy
  for acc and cacc under uncertainty}.
\newblock \bibinfo{journal}{Transportation Research Part C: Emerging
  Technologies} \bibinfo{volume}{83}, \bibinfo{pages}{61--76}.
\bibitem[{Zhou et~al.(2020)Zhou, Fu, Wang and Zhang}]{zhou2020modeling}
\bibinfo{author}{Zhou, Y.}, \bibinfo{author}{Fu, R.}, \bibinfo{author}{Wang,
  C.}, \bibinfo{author}{Zhang, R.}, \bibinfo{year}{2020}.
\newblock \bibinfo{title}{Modeling car-following behaviors and driving styles
  with generative adversarial imitation learning}.
\newblock \bibinfo{journal}{Sensors} \bibinfo{volume}{20},
  \bibinfo{pages}{5034}.
\bibitem[{Zhou et~al.(2019)Zhou, Wang and Ahn}]{zhou2019distributed}
\bibinfo{author}{Zhou, Y.}, \bibinfo{author}{Wang, M.}, \bibinfo{author}{Ahn,
  S.}, \bibinfo{year}{2019}.
\newblock \bibinfo{title}{Distributed model predictive control approach for
  cooperative car-following with guaranteed local and string stability}.
\newblock \bibinfo{journal}{Transportation research part B: methodological}
  \bibinfo{volume}{128}, \bibinfo{pages}{69--86}.
\bibitem[{Zhu et~al.(2020)Zhu, Wang and Hu}]{zhu2020impact}
\bibinfo{author}{Zhu, M.}, \bibinfo{author}{Wang, X.}, \bibinfo{author}{Hu,
  J.}, \bibinfo{year}{2020}.
\newblock \bibinfo{title}{Impact on car following behavior of a forward
  collision warning system with headway monitoring}.
\newblock \bibinfo{journal}{Transportation research part C: emerging
  technologies} \bibinfo{volume}{111}, \bibinfo{pages}{226--244}.
\bibitem[{Zhu et~al.(2018)Zhu, Wang and Wang}]{NN_DRL_zhu2018human}
\bibinfo{author}{Zhu, M.}, \bibinfo{author}{Wang, X.}, \bibinfo{author}{Wang,
  Y.}, \bibinfo{year}{2018}.
\newblock \bibinfo{title}{Human-like autonomous car-following model with deep
  reinforcement learning}.
\newblock \bibinfo{journal}{Transportation research part C: emerging
  technologies} \bibinfo{volume}{97}, \bibinfo{pages}{348--368}.

\end{thebibliography}


\end{document}